\documentclass{article}

\usepackage{arxiv}

\usepackage[utf8]{inputenc} 
\usepackage[T1]{fontenc}    
\usepackage{hyperref}       
\usepackage{url}            
\usepackage{booktabs}       
\usepackage{amsfonts}       
\usepackage{nicefrac}       
\usepackage{microtype}      
\usepackage{lipsum}		
\usepackage{graphicx}
\usepackage{doi}
\usepackage{caption}
\usepackage{listings}
\usepackage{xcolor}
\usepackage{fancyhdr}       
\usepackage{graphicx}       
\usepackage{fancyvrb}        

\usepackage{setspace} 

\usepackage{soul}
\usepackage{amsmath}
\usepackage{listings}
\usepackage{caption}

\usepackage{multirow}
\usepackage{helvet} 
\usepackage{tikz}
\usetikzlibrary{shapes.geometric, arrows, arrows.meta, positioning,fit,backgrounds, shadows}
\usepackage{enumitem}

\usetikzlibrary{decorations.pathreplacing}

\newcounter{textbox}
\renewcommand{\thetextbox}{\arabic{textbox}} 

\newcommand{\textboxcaption}[1]{
  \refstepcounter{textbox}
  \noindent\small\textbf{Text Box \thetextbox:} #1\par\medskip
}

\usepackage{sansmath} 

\tikzstyle{process} = [rectangle, rounded corners, minimum width=3cm, minimum height=1cm, text centered, draw=black, fill=orange!30]
\tikzstyle{start} = [ellipse, minimum width=3cm, minimum height=1cm, text centered, draw=black, fill=yellow!30]
\tikzstyle{decision} = [diamond, minimum width=3cm, minimum height=1cm, text centered, draw=black, fill=green!30]
\tikzstyle{arrow} = [thick,->,>=stealth]

\tikzstyle{phase} = [rectangle, rounded corners, minimum width=3cm, minimum height=1cm, text centered, draw=black, fill=orange!30]
\tikzstyle{result} = [ellipse, minimum width=3cm, minimum height=1cm, text centered, draw=black, fill=green!30]

\tikzstyle{dataset} = [rectangle, rounded corners, minimum width=2cm, minimum height=1cm, text centered, draw=black, fill=blue!30]
\tikzstyle{masking} = [rectangle, rounded corners, minimum width=2cm, minimum height=1cm, text centered, draw=black, fill=red!30]

\usepackage[skins,breakable]{tcolorbox}

\definecolor{aigold}{RGB}{244,210, 1} 
\definecolor{aigreen}{RGB}{210,244,211} 

\sethlcolor{aigreen}

\definecolor{aired}{RGB}{255,180,181}

\definecolor{aigold}{RGB}{255,180,181}

\definecolor{aiblue}{RGB}{173,216,230} 
 
\definecolor{lightred}{rgb}{1,0.9,0.9} 

\newcommand{\hlred}[1]{{%
  \sethlcolor{lightred}
  \hl{#1}
}}

\newcommand{\hllightgreen}[1]{{%
  \sethlcolor{aigreen}
  \hl{#1}
}}

\tcbset{
  customboxsmalll/.style={
    width=0.95\textwidth,  
    fonttitle=\ttfamily\scriptsize, 
    colback=white,  
    colframe=black,  
    colbacktitle=blue!75!black, 
    coltitle=white,  
    boxrule=0.5pt,  
    rounded corners,  
    boxed title style={
        colback=blue!75!black, 
        colframe=black, 
        boxrule=0.5pt  
            before={\setlength{\parskip}{0pt} \setlength{\parindent}{-6pt}},  
    after={\setlength{\parskip}{0pt} \setlength{\parindent}{-6pt}},   
    halign=center,  
    before skip=-5pt,  
    after skip=-5pt,   
    innertop=-3pt,     
    innerbottom=-3pt,  
    valign=center,    
    }
    
  }
}

\tcbset{
  custombox/.style={
    width=512.18663pt,
    top=3pt,
    fonttitle=\bfseries\scriptsize\sffamily, 
    colback=white,
    colframe=black,
    colbacktitle=black,
    boxrule=0.5pt, 
    colback=white, 
    enhanced,
    rounded corners, 
    center,
    boxed title style={
            sharp corners,
            size=small,
            colframe=blue!75!black, 
        },
    attach boxed title to top left={yshift=-0.1in,xshift=0.15in},
    boxed title style={boxrule=0pt,colframe=white,},
    before=\setlength{\baselineskip}{2pt}, 
    before upper=\setlength{\baselineskip}{2pt}, 
    after=\setlength{\baselineskip}{2pt}, 
  }
} 

\newtcolorbox{LLMbox}[2][]{custombox, title=#2,#1}
\newtcolorbox{LLMboxSmall}[2][]{customboxsmalll, title=#2,#1}

\DefineVerbatimEnvironment{thinkingcolor}{Verbatim}{
  formatcom=\color{blue}, 
  fontsize=\scriptsize  
}
\DefineVerbatimEnvironment{reflectcolor}{Verbatim}{
  formatcom=\color{orange}, 
  fontsize=\scriptsize  
}

\title{ 
\textit{In-situ} Graph Reasoning and Knowledge Expansion using Graph-PReFLexOR  }

\author{ \href{https://orcid.org/0000-0002-4173-9659}{\includegraphics[scale=0.06]{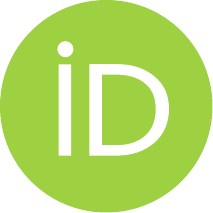}\hspace{1mm}Markus J. Buehler}\thanks{Corresponding author.} \\
	Laboratory for Atomistic and Molecular Mechanics\\
	MIT\\
	Cambridge, MA 02139, USA \\
	\texttt{mbuehler@MIT.EDU} \\
}

\newtcbox{\mybox}[1][green]{on line,
arc=0pt,outer arc=0pt,colback=#1!10!white,colframe=#1!50!black,
boxsep=0pt,left=0pt,right=0pt,top=0pt,bottom=0pt,
boxrule=0pt,bottomrule=0pt,toprule=0pt}

\tcbset{
  customboxmultipage/.style={
    width=512.18663pt,
    top=3pt,
    fonttitle=\ttfamily\scriptsize, 
    colback=white,
    breakable,
    colframe=black,
    colbacktitle=black,
    boxrule=0.5pt,
    rounded corners,
    center,
    boxed title style={
        sharp corners,
        size=small,
        colframe=blue!75!black,
    },
    attach boxed title to top left={yshift=-0.1in,xshift=0.15in},
    boxed title style={boxrule=0pt,colframe=white,},
    before={\par\noindent},
    after={\par},
    before upper={\setstretch{0.3}}, 
    after lower={\nointerlineskip},
    every page on break/.style={
      interior style={top color=white, bottom color=white},
    },
  }
}

\newtcolorbox{LLMboxmultipage}[2][]{customboxmultipage,title=#2,#1}

\renewcommand{\headeright}{}
\renewcommand{\shorttitle}{\textit{In situ} graph reasoning and knowledge expansion using Graph-PReFLexOR }

\hypersetup{
pdftitle={
\textit{In-situ} Graph Reasoning and Knowledge Expansion using Graph-PReFLexOR},
pdfsubject={q-bio.NC, q-bio.QM},
pdfauthor={Markus J. Buehler, MIT, mbuehler@MIT.EDU},
pdfkeywords={Language Modeling, Graph Theory, Category Theory, Materials Science, Materiomics, Symbolic Reasoning},
}

\begin{document}
\maketitle

\begin{abstract}
The pursuit of automated scientific discovery has fueled progress from symbolic logic to modern AI, forging new frontiers in reasoning and pattern recognition. Transformers function as potential systems, where every possible relationship remains latent potentiality until tasks impose constraints, akin to measurement. Yet, refining their sampling requires more than probabilistic selection: solutions must conform to specific structures or rules, ensuring consistency and the invocation of general principles. We present Graph-PReFLexOR (Graph-based Preference-based Recursive Language Modeling for Exploratory Optimization of Reasoning), a framework that combines graph reasoning with symbolic abstraction to dynamically expand domain knowledge. Inspired by reinforcement learning, Graph-PReFLexOR defines reasoning as a structured mapping \(\mathcal{M}: \mathcal{T} \rightarrow (\mathcal{G}, \mathcal{P}, \mathcal{A})\), where tasks yield knowledge graphs \(\mathcal{G}\), abstract patterns \(\mathcal{P}\), and final answers \(\mathcal{A}\). Inspired by category theory, it encodes concepts as nodes and their relationships as edges, supporting hierarchical inference and adaptive learning through isomorphic representations. Demonstrations include hypothesis generation, materials design, and creative reasoning, such as discovering relationships between mythological concepts like ``thin places'' with materials science. We propose a ``knowledge garden growth'' strategy that integrates insights across domains, promoting interdisciplinary connections. Results with a 3-billion-parameter Graph-PReFLexOR model show superior reasoning depth and adaptability, underscoring the potential for transparent, multidisciplinary AI-driven discovery. It lays the groundwork for general autonomous reasoning solutions.
\end{abstract}


\section{Introduction}

Discovery is driven by the ability to think deeply, reflect, and iteratively refine ideas before arriving at conclusions. Emergent artificial intelligence (AI) methods have achieved success in domains such as natural language processing~\cite{Vaswani2017AttentionNeed,AlecRadfordImprovingPre-Training,Xue2021ByT5:Models,Jiang2023Mistral7Bb,Phi-2:Research,dubey2024llama3herdmodels,Brown2020LanguageLearners}, materials science, and molecular biology, including protein folding~\cite{Jumper2021HighlyAlphaFold}, and others. However, much of these advances rely on models that produce outputs directly, without an explicit intermediate reasoning process akin to scientific thinking, and hence, generally lack self-awareness and capacity to reflect. For AI to ultimately accelerate discovery, we must develop systems capable of reasoning in relational and symbolical manners~\cite{Spivak2011CategoryNetworks,Spivak2011ReoccurringAnalogies, Giesa2012CategoryDesign,abbott2024flashattentionnapkindiagrammaticapproach,Buehler2025GraphAwareGPT}, reflecting critically, and explaining their answers in an iterative process that mirrors scientific methods of exploration.

\begin{figure}
\centering

\includegraphics[width=0.9\linewidth]{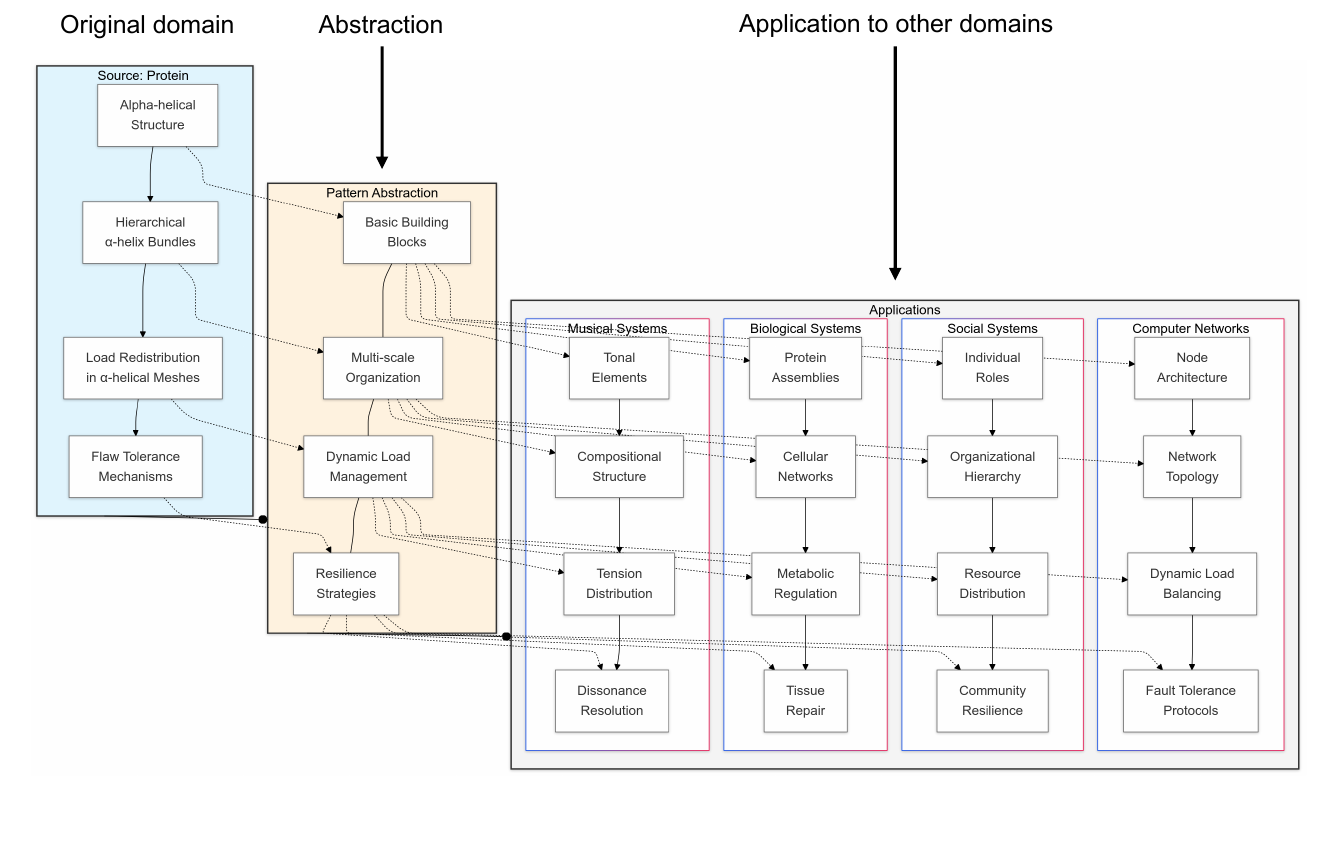}

\sffamily 
\scriptsize
\sansmath 
\begin{tikzpicture}[node distance=2.5cm and .8cm, auto,
    task/.style={rectangle, draw, fill=blue!20, text width=2.cm, text centered, rounded corners, minimum height=1.2cm},
    graph/.style={rectangle, draw, fill=violet!40, text width=2.5cm, text centered, rounded corners, minimum height=1.2cm},
    symbolic/.style={rectangle, draw, fill=violet!20, text width=2.5cm, text centered, rounded corners, minimum height=1.2cm},
    hypothesis/.style={rectangle, draw, fill=violet!10, text width=2.5cm, text centered, rounded corners, minimum height=1.2cm},
    answer/.style={rectangle, draw, fill=green!20, text width=2.5cm, text centered, rounded corners, minimum height=1.2cm},
    math_symbols/.style={rectangle, draw, fill=yellow!30, text width=2.cm, text centered, rounded corners, minimum height=1.2cm},
    other_tasks/.style={rectangle, draw, fill=orange!30, text width=4cm, text centered, rounded corners, minimum height=1.2cm},
    line/.style={draw, -stealth', shorten >=1pt},
    dashed line/.style={draw, dashed, shorten >=1pt}]

    \node [task] (task) {Task};
    \node [graph, right=of task] (graph_rep) {Graph Representation};
    \node [symbolic, right=of graph_rep] (symbolic_rep) {Symbolic Representation };
    \node [hypothesis, right=of symbolic_rep] (hypothesis) {Relevant Materials \& Concepts, ..., Hypothesis};
    \node [answer, right=of hypothesis] (final_answer) {Final Answer};

    \path [line] (task) -- (graph_rep);
    \path [line] (graph_rep) -- (symbolic_rep);
    \path [line] (symbolic_rep) -- (hypothesis);
    \path [line] (hypothesis) -- (final_answer);

    \node [math_symbols, above=of symbolic_rep, yshift=-1cm] (math_symbols) {$\alpha$, $\beta$, $\delta$, $\rightarrow$, $\propto$, ...};
    \path [dashed line] (math_symbols) -- (symbolic_rep);

    \node [above=of graph_rep, yshift=-2cm, align=center] (sample_graph) {
        \begin{tikzpicture}[node distance=0.8cm]
            \node[circle, draw, fill=blue!30, minimum size=0.2cm] (a) {A};
            \node[circle, draw, fill=blue!30, minimum size=0.2cm, right=of a] (b) {B};
            \node[circle, draw, fill=blue!30, minimum size=0.2cm, below=of b] (c) {C};
            
            \path [line] (a) -- node[midway, above] {IS-A} (b);
            \path [line] (b) -- node[midway, right] {RELATES-TO} (c);
            \path [line] (c) -- node[midway, left] {INFLUENCES} (a);
        \end{tikzpicture}
    };

    \begin{pgfonlayer}{background}
        \node [draw=orange, thick, dashed, fit=(graph_rep) (symbolic_rep) (hypothesis) (math_symbols) (sample_graph), inner sep=0.5cm, 
               label={[text=orange]above: Mapping to Generalized Representation: Abstraction}] (abstract_box) {};
    \end{pgfonlayer}

    \node [other_tasks, below=of abstract_box, yshift=1.5cm] (other_tasks) {Other tasks \\ (map to similar abstract representations)};
    \path [dashed line] (other_tasks.north) -- (abstract_box.south);

\end{tikzpicture}

\sffamily 
        \caption{Visualization of generalization via abstraction. Top: Example, where a phenomenon in an original domain (here: protein materials fracture, specifically flaw-tolerance in alpha-helical protein meshes~\cite{Ackbarow2009Alpha-helicalFlaw-tolerant}) is modeled as relational abstract patterns, and then used to describe distinct phenomena in other domains. The diagram shows how structural patterns in protein materials can be abstracted and applied across domains through categorical mappings and graph-based relationships. The three-level hierarchy demonstrates functorial relationships between source domain concepts (protein materials), abstract pattern recognition, and diverse applications in networks, social systems, biological systems, and musical composition. Bottom: Flowchart for visualizing the process from a task to a graph representation (with shared relational descriptions such as \texttt{IS-A}, \texttt{RELATES-TO}, and \texttt{INFLUENCES}), symbolic abstraction, hypothesis generation, and the final answer. The vertical dashed line with mathematical symbols ($\alpha$, $\beta$, $\delta$, $\rightarrow$) represents the shared representation of all problems in tokenized form, where the model learns to generalize representations across domains.}
	\label{fig:fig_01}
\end{figure}
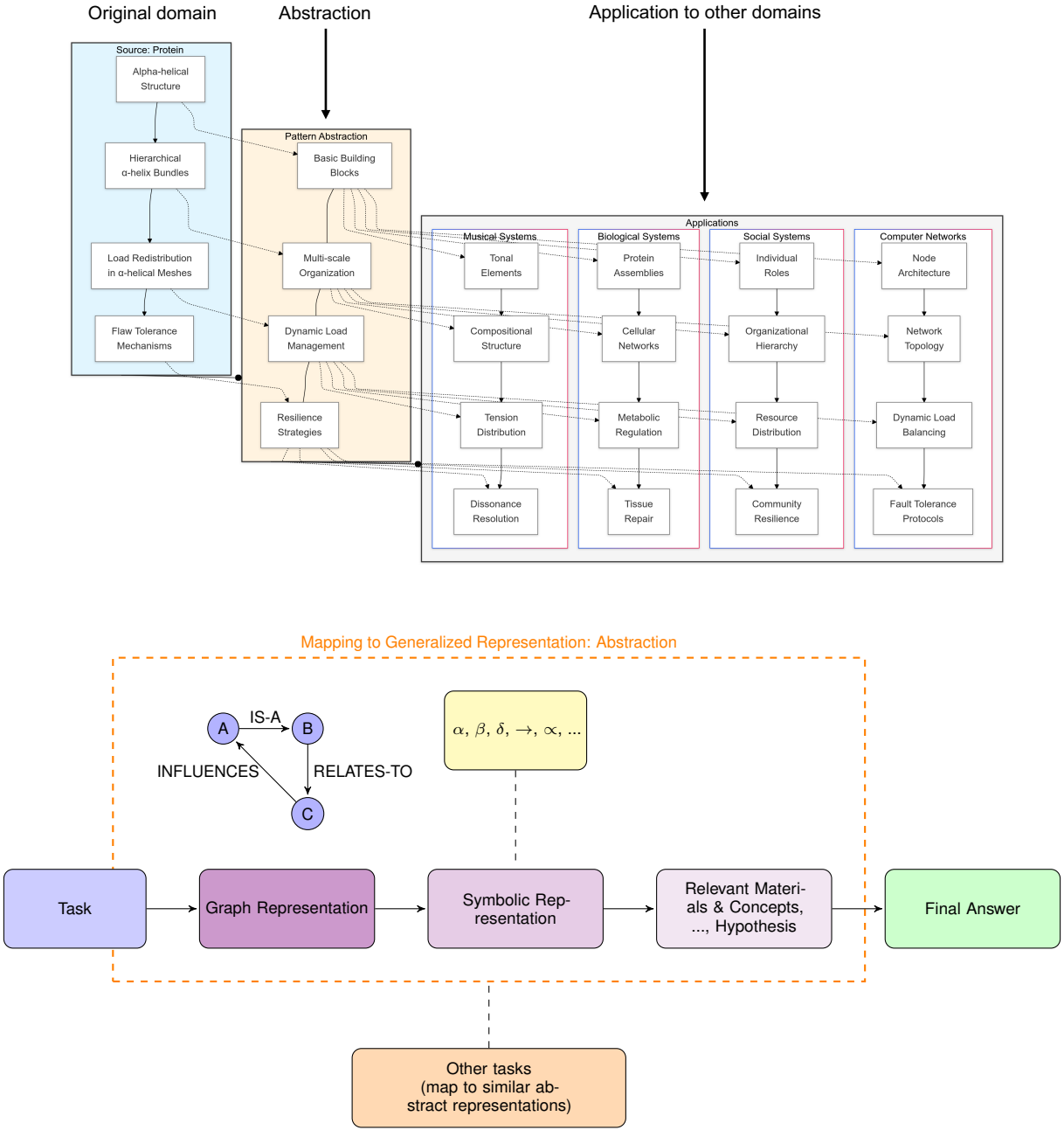

When we solve a problem by identifying how things are connected and flow together, we often discover a template that works far beyond your original situation, serving as a model that can be understood in a different context or applied to a new scenario to extrapolate. For example, the hierarchical structure of bone, combining stiffness and toughness across scales, has inspired the design of lightweight, durable materials for aerospace and architecture~\cite{Nepal2022HierarchicallyNanocomposites,Ritchie2009PlasticityBone,Launey2010OnBone}. Category theory~\cite{Eilenberg1942GroupHomology,Eilenberg1945GeneralEquivalences,Spivak2011ReoccurringAnalogies, Giesa2012CategoryDesign,Spivak2011CategoryNetworks}, a branch of mathematics that emphasizes how objects relate rather than their internal detail, can be a powerful tool to construct models of complex phenomena through a lens on relational aspects. At its core, category theory emphasizes morphisms—arrows that represent interactions or mappings between objects. Such an approach can reveal hidden patterns - where the same structures keep appearing in nature, human organizations, technology, and science. By mapping out these connections and flows, we can unlock powerful solutions that have worked across many fields. We can view this like discovering a universal language that helps translate successful ideas from one domain to another, letting us recognize and apply useful, and transferrable, patterns wherever they might be useful (Figure~\ref{fig:fig_01}). In the example depicted in the figure, we show how flaw-tolerance in alpha-helical protein meshes can be modeled as relational abstract patterns, and then used to describe phenomena in other domains~\cite{Ackbarow2009Alpha-helicalFlaw-tolerant,Giesa2012CategoryDesign,Giesa2015Matriarch:Architecture,Spivak2011CategoryNetworks}. 

A central concept is that of an isomorphism, a special type of morphism that establishes a structural equivalence between objects, enabling insights from one domain to be applied to another. For example, consider Newton's second law of motion, \( F = ma \), which describes the proportionality between force \( F \), mass \( m \), and acceleration \( a \). This relationship is isomorphic to Ohm’s law in electrical circuits, \( V = IR \), where voltage \( V \) plays the role of force, current \( I \) corresponds to acceleration, and resistance \( R \) serves as the proportionality factor akin to mass. Both laws can be abstracted into a general form, \( y = kx \), where \( y \) represents the driving factor, \( x \) the response, and \( k \) the proportionality constant. This abstraction reveals that the structural relationships governing these systems are fundamentally equivalent, even though the physical quantities differ. By identifying such isomorphisms, category theory allows scientists to transfer insights across domains, uncovering deep connections between seemingly unrelated phenomena. This approach not only simplifies complex systems but also provides a systematic framework for discovering universal laws, as it emphasizes relational properties that transcend specific instances. Conversely, if we can discover isomorphisms directly from data, it opens the possibility of uncovering new theories and equations by identifying deep structural parallels across disparate domains, revealing universal principles that might otherwise remain hidden.

\begin{figure}
	\centering
	\includegraphics[width=0.9\linewidth]{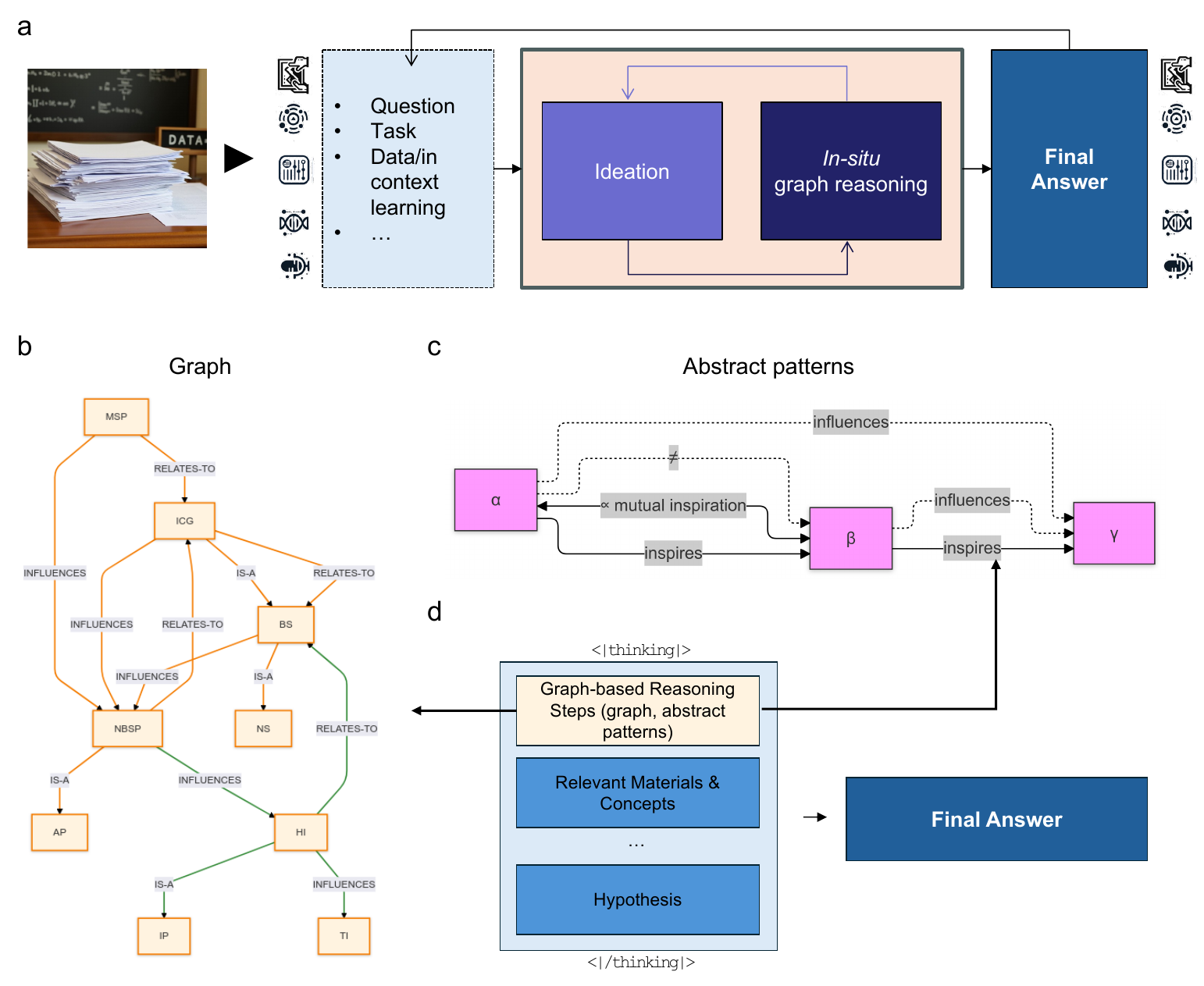}
        \caption{Overview of the approach used in this paper, presenting the concept of multi-step reflection (panel a), graph-based modeling of context and tasks (panel b), abstract pattern formulation (panel c), and finally, integrated in the multi-stage reasoning mechanisms (panel d).}
	\label{fig:fig_5}
\end{figure}

\subsection{Modeling isomorphisms for generalization} 

To model isomorphisms computationally, we require a neural network architecture capable of capturing the structural equivalences between relational systems. Graph Isomorphism Networks (GINs) are particularly well-suited for this purpose. GINs are designed to operate on graph-structured data, where nodes represent objects and edges capture relationships or interactions. Unlike traditional graph neural networks (GNNs), which may struggle to distinguish between non-isomorphic graphs, GINs achieve maximum expressiveness for distinguishing graph structures by leveraging a theoretically grounded update rule that closely resembles the Weisfeiler-Lehman graph isomorphism test~\cite{weisfeiler1968reduction}. In a GIN, the node update function aggregates features from neighboring nodes using a weighted sum and applies a learned multi-layer perceptron (MLP) to the result, ensuring the network can capture higher-order structural properties. For example, to model the isomorphism between Newton's second law, \( F = m\times a \), and Ohm’s law, \( V = I\times R \), a GIN can represent the relationships between variables and operators (e.g., force, mass, acceleration, =, $\times$) as nodes and their interactions as edges. By learning the graph embeddings, the GIN can identify the shared relational structure underlying these systems, demonstrating their equivalence. After processing, the embeddings for both equations would be nearly identical in the latent space, reflecting their structural similarity and isomorphic relationship as they can be mapped to the same representation (see, Section~\ref{GIN_mapping_ex} and Figure~\ref{fig:fig_7000}, for an example). This capability to recognize and model isomorphisms makes GINs a powerful tool for tasks that involve reasoning over relational data, such as predicting properties of molecules, designing robust materials, or uncovering universal scientific laws.

How does this relate to Transformer-based LLMs? Recent work has shown that Transformers implicitly function as graph isomorphism neural networks~\cite{Buehler2025GraphAwareGPT}, which provides a powerful lens for explicitly integrating graph-based reasoning into these architectures. Whereas~\cite{Buehler2025GraphAwareGPT} focused on theoretically grounded advancements in the Transformer architecture itself to strengthen its expressive capacity, in this paper we focus on utilizing a standard Transformer interpreted with the expressive power of a GIN, and endow it with explicit capabilities to conduct \textit{in-situ} graph reasoning and symbolic deduction. We hypothesize that by explicitly leveraging graph theory within Transformers can unlock new levels of performance, generalization, and interpretability, accelerating progress in various scientific domains. Moreover, modeling complex principles can be inspired from biology, specifically the use of hierarchically organized systems that utilize some of the core ideas of category theory in materials and scientific exploration, where natural systems are found to often reuse relational patterns and  mechanisms such as amino acids, DNA, and others. Building on these ideas, it has been suggested that inspiration for such advances may come from biology, such as emergent hierarchical intelligence based on agentic systems and other strategies that mimic biological materials and design principles~\cite{Cranford2012Biomateriomics,Groen2011IntroducingMateriomics,Arevalo2023LearningMaterials,Buehler2023MeLMProblemsc,ghafarollahi2024sciagentsautomatingscientificdiscovery}. Related principles include concerted problem solving as illustrated in a recent study of ant dynamics~\cite{doi:10.1073/pnas.2414274121}, providing ample examples for latent opportunities to discover mechanisms that can be translated across domains.

Large Language Models (LLMs) have demonstrated capabilities in generating human-like text and extracting insights from scientific literature. A grand challenge in the application of AI models in science in particular remains to develop algorithms that yield more sophisticated reasoning mechanisms~\cite{Buehler2023MeLMProblemsc, Singhal2023, Ghafarollahi2024, Luu2023BioinspiredLLM:Materials, Yu2024, Hu2023DeepScience, Buehler2024X-LoRA:Design,Buehler2023MechGPTModalities_updated, lu2024finetuninglargelanguagemodels}. These models often excel in fluency and pattern recognition, yet typically lack the ability to engage in structured reasoning. Addressing this limitation, earlier work like PReFLexOR (abbreviation of: Preference-based Recursive Language Modeling for Exploratory Optimization of Reasoning)~\cite{buehler2024preflexorpreferencebasedrecursivelanguage} introduced a framework where models were developed that innately develop a thinking and reflection strategy before answering, leveraging reinforcement learning (RL) to refine reasoning through iterative reflection and dynamic feedback loops, inspired by earlier research~\cite{zelikman2022starbootstrappingreasoningreasoning,zelikman2024quietstarlanguagemodelsteach,wei2023chainofthoughtpromptingelicitsreasoning,kojima2023largelanguagemodelszeroshot}. A key goal of the work reported in this paper is to develop a strategy that allows AI models to not simply retrieve information but to learn underlying abstractions and relational motives, and then use these insights to generalize across new domains and ultimately to expand its understanding and capabilities to generalize. 

Building on this foundation, we expand the concept of ``thinking before answering'' to incorporate \textit{in situ} graph-based reasoning. Graphs provide a powerful framework for representing relationships, causality, and structures inherent in scientific problems (Figure~\ref{fig:fig_5}). From molecular interactions, dynamics, to material properties, many scientific phenomena are naturally represented as graphs. By enabling models to autonomously construct and manipulate these symbolic graph representations, we create a model that mimics the reflective, iterative reasoning processes integral to scientific inquiry. In Transformers, knowledge is represented as a superposition of potentialities that collapse into specific outputs when conditioned on tasks. This interpretation offers a framework for enhancing both reasoning and creativity. We show how, by balancing structured coherence with divergent exploration, models can traverse new ideas, generating insights while maintaining rigorous, task-aligned solutions.

\subsection{Foundations of \textit{in-situ} graph reasoning} 

In this work, we introduce a novel approach that unifies the linguistic fluency of LLMs with the relational reasoning capabilities that have been quite successful in architectures such as Graph Neural Networks (GNNs). Our model extends the reflective capabilities introduced in the original PReFLexOR model, enabling the construction of graph-based intermediate reasoning mechanisms, by taking advantage of the flexible capabilities of the transformer architecture. Through these symbolic representations via special tokens, the model engages in a ``thinking phase,'' reasoning over the graph to refine its understanding before generating an answer. This approach not only enhances interpretability but also improves the model's ability to solve complex scientific problems requiring relational and structural reasoning. We achieve this by formalizing reasoning as a structured mapping:
\begin{equation} \label{eq:M_T_G_P}
    \mathcal{M}: \mathcal{T} \rightarrow (\mathcal{G}, \mathcal{P}, \mathcal{A}), 
\end{equation}
where a task \( \mathcal{T} \) produces a knowledge graph \( \mathcal{G} \), abstract patterns \( \mathcal{P} \), and final answers \( \mathcal{A} \). The knowledge graph 
\begin{equation}\label{eq:G_V_E}
\mathcal{G} = (V, E) 
\end{equation}
encodes concepts as nodes \( V \) and relationships as directed edges \( E \). 

By explicitly constructing and abstracting relational graphs, the model can encode structural information that standard next‐token prediction often overlooks or treats only implicitly. In this setup, each entity and its relations become first‐class objects in the learned representation, enabling the network to detect and exploit common subgraph motifs and isomorphisms that recur across different inputs. Unlike pure sequence‐based transformers, which must infer latent structure solely from token order and distributional cues, a graph‐driven approach anchors learning in explicitly linked entities, preserving both local and global connectivity. As a result, it is not only easier for the model to discover universal features (such as repeated subgraphs or underlying algebraic forms) but also more straightforward to apply symbolic abstractions, since the graph representation makes these recurring structures and higher‐level patterns more salient and amenable to consistent transformations.

Our model thereby offers opportunities for diverse use cases in AI for science and beyond. Tasks such as multi-step reasoning, hypothesis generation, and causal inference become more robust and explainable. This work bridges connectionist and symbolic paradigms, pushing the boundaries of what AI can achieve in scientific domains~\cite{rumelhart1986learning,newell1972human,mccarthy1960programs}. By creating models that have a more explicit process of relational deliberation, both symbolically and structurally, before answering, we take an important step toward AI systems capable of true scientific reasoning and discovery.

\subsection{Outline of this paper}

The plan of the paper is as follows. First, we briefly review the PReFLexOR architecture published in earlier work~\cite{buehler2024preflexorpreferencebasedrecursivelanguage}, then describe the training process of developing Graph-PReFLexOR, and then review several case studies of how the method can be applied. We conclude with final thoughts and a discussion on future opportunities.

\section{Results and Discussion}

We report the results of a series of experiments and analyses conducted based on the model. We start with a review of the PReFLexOR framework~\cite{buehler2024preflexorpreferencebasedrecursivelanguage}, how we adapted it for graph reasoning capability, and then move on to the results of the experiments conducted. We conclude with a discussion of the results, interpretations, and an outlook to future opportunities. 

\subsection{Review of PReFLexOR}

\begin{figure}
	\centering
	\includegraphics[width=0.7\linewidth]{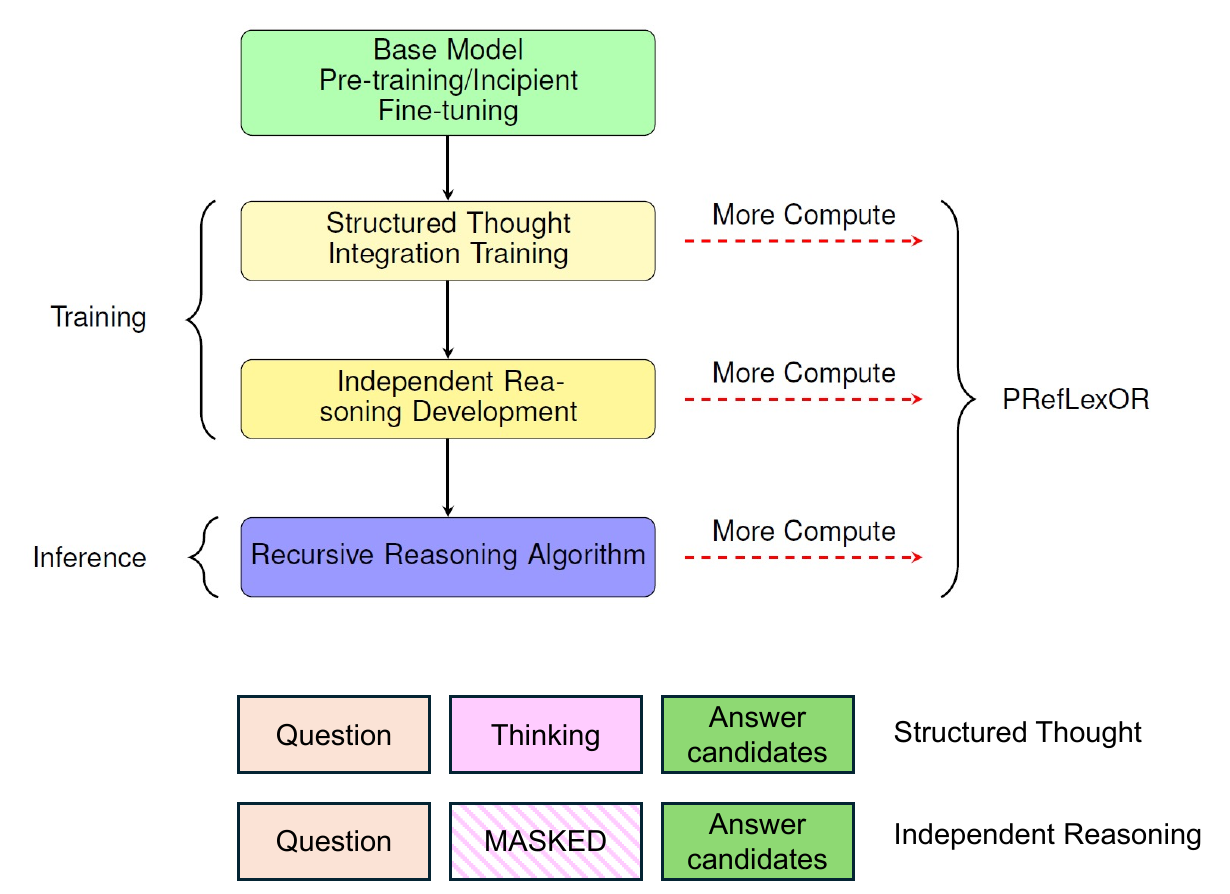}
        \caption{Overview of the PRefLexOR framework as reported in~\cite{buehler2024preflexorpreferencebasedrecursivelanguage}, presented here for completeness. The training process involves two stages: (1) Structured Thought Integration Training, focusing on incorporating structured reasoning components, and (2) Independent Reasoning Development, aimed at fostering model autonomy in reasoning. During inference, the Recursive Reasoning Algorithm is employed to iteratively refine responses. Below, the role of reasoning components is depicted in the two training phases, showing transitions from unmasked to masked reasoning.}
	\label{fig:fig_5000}
\end{figure}

PRefLexOR, Preference-based Recursive Language Modeling for Exploratory Optimization of Reasoning~\cite{buehler2024preflexorpreferencebasedrecursivelanguage}, is a framework that enhances the reasoning capabilities of language models by integrating preference optimization and recursive reasoning (Figure~\ref{fig:fig_5000}). The approach relies on the introduction of special tokens such as \texttt{<|thinking|>}..\texttt{<|/thinking|>} and \texttt{<|reflect|>}..\texttt{<|/reflect|>}, which explicitly mark distinct phases of reasoning, enabling structured generation and iterative refinement. During training, the model undergoes two stages: Structured Thought Integration, where reasoning processes are guided using special tokens and optimized through Odds Ratio Preference Optimization (ORPO), and Independent Reasoning Development, which leverages masking of thinking tokens and applies Efficient Exact Optimization (EXO)~\cite{ji2024efficientexactoptimizationlanguage_EXO} to align final outputs with preferred reasoning patterns without prescribing the specific reasoning steps. Thinking tokens allow the model to simulate intermediate reasoning steps, while masking ensures that the model infers the best reasoning pathways without direct supervision. We note that EXO's reverse Kullback-Leibler divergence~\cite{kullback1951information} objective promotes mode-seeking behavior, emphasizing dominant reasoning strategies over diluted alternatives. By dynamically generating datasets and employing recursive feedback loops, PRefLexOR enables models to self-teach, iteratively refining their reasoning through on-the-fly task generation and reflection. This combination of token-driven structure, recursive refinement, and preference alignment makes PRefLexOR a suitable foundation for reasoning-intensive applications.

As was shown in the original paper~\cite{buehler2024preflexorpreferencebasedrecursivelanguage}, with a trained model the reasoning process can be iteratively refined at inference time, where the model generates intermediate reasoning (\textit{thinking phase}), the critic evaluates it (\textit{reflection phase}), and then produces improved responses. Formally, the reasoning at step $i$, $\mathbf{R}_i$, is represented as:
\begin{equation} \label{eq:recursive}
\mathbf{R}_{i+1} = f_\text{critic}(\mathbf{R}_i, \mathbf{F}_i),
\end{equation}
where $f_\text{critic}$ applies feedback $\mathbf{F}_i$ to refine the intermediate reasoning. At each step, reflection is guided by feedback on gaps or inconsistencies in $\mathbf{R}_i$, enabling the next iteration to improve alignment with the desired output. The final response $\mathbf{A}$ is derived after $N$ refinement steps as:
\begin{equation}
\mathbf{A} = g(\mathbf{R}_N),
\end{equation}
where $g$ extracts the synthesized final answer. During training, masking is applied to tokens embedded within \texttt{<|thinking|>}..\texttt{<|/thinking|>}, requiring the model to infer reasoning paths indirectly while optimizing the final output. This iterative combination of thinking and reflection, structured via tokens and recursive refinement, ensures that the model autonomously enhances its reasoning capabilities, providing accurate and well-aligned outputs.

Here is an example:

\begin{LLMbox}{Basic structure of the reasoning strategy using a thinking phase before answering.}
\begin{lstlisting}
(*@\bf\hlred{User:}@*) [User question or task]

(*@\bf\hllightgreen{Assistant:}@*) 
(*@\begin{thinkingcolor}
<|thinking|>

...

[GRAPH]

...

[ABSTRACT PATTERNS]

...

[ADDITIONAL THINKING STEPS]

...

<|/thinking|>
(*@\end{thinkingcolor}

[Answer]
\end{lstlisting}
\end{LLMbox}

\subsection{Graph-PReFLexOR algorithm design and training}

Here we use the original PReFLexOR framework but create a different function to develop the structured training data by teaching the model to explicitly construct graphs and symbolic representations in its thinking phase. Details are included in the Materials and Methods section, but we review highlights here. All training is done on top of the \texttt{meta-llama/Llama-3.2-3B-Instruct} model, a small but performant base LLM that serves as the foundational platform for development. 

Structured reasoning is generated on-the-fly during training as described in the Materials and Methods section. At its core, the method constructs dynamic knowledge graphs \textit{in-situ} \( G = (V, E) \), where nodes \( V \) represent key concepts extracted from enriched context, and edges \( E \) encode relationships such as \texttt{IS-A}, \texttt{RELATES-TO}, and \texttt{INFLUENCES}, common and shared relational descriptions. These relationships are quantified using a semantic scoring function \( f(c_i, c_j; r_k) \), which evaluates the significance of relationships \( r_k \) between concepts \( c_i \) and \( c_j \). Retrieval-Augmented Generation (RAG) is employed to enrich the input context and identify abstract patterns and dependencies within the graph, enabling higher-order reasoning.

The thinking section organizes reasoning into structured headings, including `Core Concepts and Relationships', `Reasoning Steps', and `Abstract Patterns', providing a framework for logical progression and shared token representation to encourage generation (see, Figure~\ref{fig:fig_01} and the discussion in Section~\ref{GIN_mapping_ex}). For example, the `Core Concepts and Relationships' heading defines the key nodes and their connections in the knowledge graph, while `Reasoning Steps' outlines step-by-step deductions using the generalized formalisms. The `Abstract Patterns' section summarizes these relationships into symbolic transformations, such as \( \alpha \to \beta \to \gamma \to \delta \to \epsilon \), with proportional dependencies \( \alpha \propto \epsilon \) indicating critical links. These structured headings not only guide the model’s reasoning but also ensure interpretability and coherence in outputs and aid to boost extrapolation capacity in recursive reasoning steps.

Special tokens delineate the reasoning phases, facilitating recursive refinement of the knowledge graph and symbolic abstractions during training and inference. By uniting graph-based reasoning, symbolic methods, and structured thinking sections, Graph-PReFLexOR achieves robust, interpretable performance in generating high-quality responses for reasoning-intensive tasks.

We train the model using the same multi-stage training process as done in the original model~\cite{buehler2024preflexorpreferencebasedrecursivelanguage}, featuring first ORPO and then DPO-EXO. The training dataset is the same as in the original paper, consisting of raw text of around 1,000 papers in the realm of biological and bio-inspired materials~\cite{Luu2023BioinspiredLLM:Materials}. 

\subsection{Detailed analysis of example responses: Single task-response pair}

In the first example, we task the model as follows, challenging the model to answer a question that partially involves a materials-specific task but also invokes a domain that was not included in the training data, to test how well the model can extrapolate its structured reasoning to other domains that were not included in the training set. The task is:

\begin{figure}
	\centering
	\includegraphics[width=1.\linewidth]{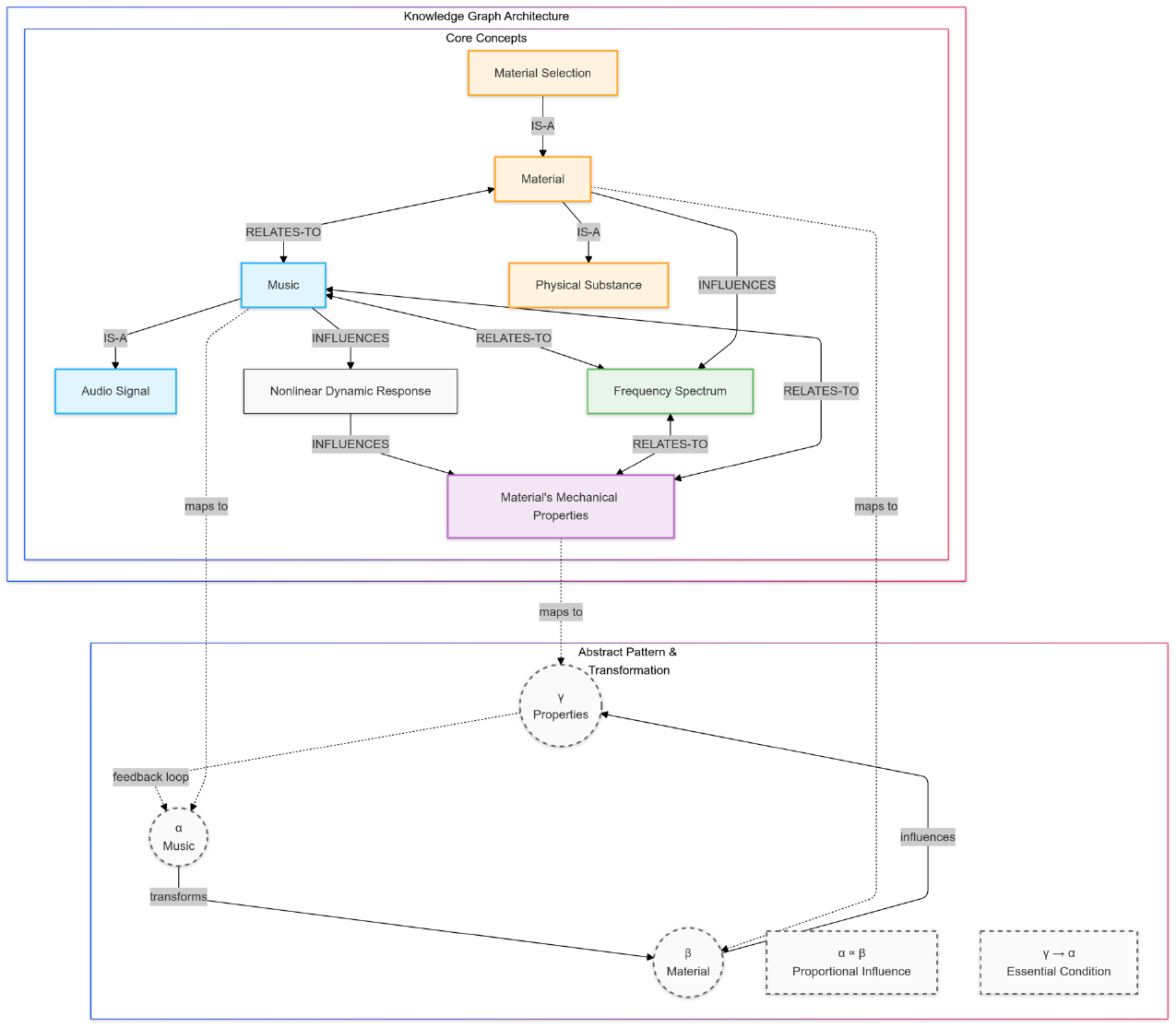}
        \caption{Overview of the data generated in response to the question: \texttt{Propose a new idea to relate music and materials.} The knowledge graph (top) illustrates the relationships between core concepts: music (blue), material properties (purple), and frequency spectrum (green). Key relationships include IS-A hierarchies (e.g., Music is an Audio Signal) and influence paths through nonlinear dynamic responses. The abstract pattern (bottom) formalizes these interactions through a triple system ($\alpha, \beta, \gamma$) with proportional influence ($\alpha \propto \beta$) and feedback loop ($\gamma \rightarrow \alpha$). The integration between graph and pattern manifests in multiple ways: the music-to-material influence path in the graph maps to $\alpha \rightarrow \beta$ in the pattern; the material's mechanical properties feedback in the graph corresponds to the essential condition $\gamma \rightarrow \alpha$; and the frequency spectrum mediation shown in the graph provides the physical mechanism for the proportional influence ($\alpha \propto \beta$) in the pattern. Dotted lines explicitly map concrete elements to their abstract counterparts, demonstrating how the theoretical framework emerges from and guides the practical implementation. This dual representation captures both the detailed mechanisms of music-material interaction and its fundamental mathematical structure.}
	\label{fig:fig_10}
\end{figure}

\begin{LLMbox}{User task}
\begin{lstlisting}
(*@\bf\hlred{User:}@*) Propose a new idea to relate music and materials.
\end{lstlisting}
\end{LLMbox}

The model responds first by thinking about the problem by outputting a series of ``thoughts'' enclosed within the \texttt{<|thinking|>} and \texttt{<|/thinking|>} tokens. The response closely follows the structured approach developed during training.  First, the model provides a knowledge graph:

\begin{LLMbox}{Knowledge graph}
\begin{lstlisting}
**Knowledge Graph:**

1. **Music** -[IS-A]-> **Audio Signal**
2. **Material** -[IS-A]-> **Physical Substance**
3. **Nonlinear Dynamic Response** -[RELATES-TO]-> **Material**
4. **Frequency** -[RELATES-TO]-> **Music**
5. **Nonlinear Dynamic Response** -[INFLUENCES]-> **Material's Mechanical Properties**
6. **Music** -[INFLUENCES]-> **Nonlinear Dynamic Response** -[INFLUENCES]-> **Material's Mechanical Properties**
7. **Frequency Spectrum** -[RELATES-TO]-> **Music**
8. **Frequency Spectrum** -[RELATES-TO]-> **Material's Mechanical Properties**
9. **Material** -[INFLUENCES]-> **Frequency Spectrum** -[INFLUENCES]-> **Music**
10. **Material Selection** -[IS-A]-> **Material** -[RELATES-TO]-> **Music** -[RELATES-TO]-> **Material's Mechanical Properties**
\end{lstlisting}
\end{LLMbox}

Next, the model develops an abstract reasoning pattern as follows:

\begin{LLMbox}{Abstract pattern}
\begin{lstlisting}
**Abstract Pattern:**

(*@$\alpha \rightarrow \beta \rightarrow \gamma$ @*) 

**Key Transformation Rule:**

(*@$\alpha \propto \beta $@*) 

**Essential Condition:**

(*@$\gamma \rightarrow \alpha $@*) 

**Explanation:**

- (*@$\alpha $@*) represents **Music**, (*@$\beta $@*) represents **Material**, and (*@$\gamma $@*) represents **Material's Mechanical Properties**.
- The abstract pattern (*@$\alpha \rightarrow \beta \rightarrow \gamma$ @*)  shows the direct relationship between Music, Material, and Material's Mechanical Properties.
- The key transformation rule (*@$\alpha \propto \beta $@*)  indicates that Music ((*@$\alpha $@*)) has a proportional influence on Material's Mechanical Properties ((*@$\beta $@*)).
- The essential condition (*@$\gamma \rightarrow \alpha $@*)  signifies that Material's Mechanical Properties ((*@$\gamma $@*)) are influenced by Music ((*@$\alpha $@*)), creating a feedback loop in the system.

Pattern Context:
Inspire a new method to tune and control the mechanical properties of materials by analyzing and manipulating the frequency spectrum of music. This method leverages the dynamic responses of materials to different frequencies, mimicking the effects of music on materials.

\end{lstlisting}
\end{LLMbox}

The model then proceeds to develop a series of additional structured mechanisms (reasoning steps, relevant materials, design principles, material properties, hypothesis, background), building on these initial steps. These are:

\begin{LLMbox}{Additional thinking steps}
\begin{lstlisting}
**Reasoning Steps**:

1. Music has dynamic responses on materials, similar to how it affects human emotions.
2. Analyze the frequency spectrum of music to understand its dynamic responses.
3. Use the frequency spectrum to tune and control the mechanical properties of materials.
4. This method leverages the relationship between music's frequency spectrum and material responses.

**Relevant Materials or Concepts**:

- Nonlinear Dynamic Response
- Frequency Spectrum
- Material Selection
- Tuning and Control of Mechanical Properties

**Design Principles**:

- **Frequency Spectrum Analysis**: Study the dynamic responses of materials to different frequencies.
- **Material Selection**: Choose materials with nonlinear dynamic responses to optimize mechanical property tuning.
- **Frequency Spectrum Manipulation**: Modify the frequency spectrum of music to influence material responses.
- **Tuning and Control**: Use the manipulated frequency spectrum to tune and control the mechanical properties of selected materials.

**Material Properties**:

- **Nonlinear Dynamic Response**: Materials exhibit nonlinear responses to external stimuli, such as vibrations or frequencies.
- **Frequency Dependence**: Material responses vary with different frequencies, allowing for tuning and control.
- **Mechanical Property Tuning**: Materials' mechanical properties can be adjusted by manipulating their nonlinear dynamic responses to specific frequencies.

**Hypothesis**:

"Music's frequency spectrum can tune and control the nonlinear dynamic responses of materials, mimicking the effects of music on materials."

**Additional Background**:

The relationship between music and materials involves analyzing and manipulating the dynamic responses of materials to different frequencies. This can be achieved by studying the nonlinear dynamic responses of materials to external stimuli, such as vibrations or frequencies. By understanding how materials react to various frequencies, it's possible to tune and control their mechanical properties. This approach can be applied to materials used in various applications, from engineering to biomedical devices. The key is to select materials that exhibit significant nonlinear responses to specific frequencies, allowing for targeted manipulation of their properties.
\end{lstlisting}
\end{LLMbox}

All of the above content was enclosed within thinking tokens. Once these are completed, the model produces the final answer, shown in Text Box~\ref{TexBox:final_answer_after_thinking_ex_1}.

We can see that the output features rich content that we can analyze. First, Figure~\ref{fig:fig_10} depicts a visualization of the knowledge graph and abstract reasoning pattern.  We find that the relationship between music and materials is formalized through an abstract pattern derived from their fundamental interactions, where the model proposes the triple system $(\alpha, \beta, \gamma)$ where $\alpha$ represents music as an audio signal, $\beta$ represents the material as a physical substance, and $\gamma$ represents the material's mechanical properties. The core relationship follows the pattern $\alpha \rightarrow \beta \rightarrow \gamma$ with the key transformation rule $\alpha \propto \beta$ and essential condition $\gamma \rightarrow \alpha$, proposing a closed feedback loop. 

This system manifests through nonlinear dynamic responses of materials to different frequencies, where the frequency spectrum of music directly influences material properties. The relationship leverages the dynamic responses of materials to different frequencies, mimicking the effects of music on materials. Through frequency spectrum analysis and manipulation, the proposed method enables tuning and control of mechanical properties such as possibly damping, stiffness, and damping capacity. This approach opens possibilities for non-destructive material testing, property tuning, and biomedical applications, where the nonlinear dynamic response provides a mechanism for controlled material modification through specific frequency interactions.

Thinking about the idea more deeply, the proposal's main novelty lies in its symmetrical conceptualization of music-material interaction as a feedback loop. Rather than just treating music as an input that affects materials (which would be more obvious), it proposes a cyclic system where material properties can also influence the musical response. The proposal relies heavily on well-known concepts like nonlinear dynamic responses and frequency analysis, but the practical implementation focuses on fairly standard approaches (frequency spectrum analysis, material selection).  The proposed methodology presents  notable innovations in relating music and materials, particularly through its biomimetic foundation that draws parallels between material responses and human emotional reactions to music ($\alpha \rightarrow \text{response}$). The approach extends beyond conventional frequency-response studies by establishing a comprehensive framework that encompasses specific technical implementations through dynamic mechanical analysis (DMA), concrete property targets (damping, stiffness), and notably, potential biomedical applications for cancer and neurological disorders. The methodology's novelty lies not only in its theoretical framework but in its practical manifestation through non-contact and non-destructive testing methods. The proposal demonstrates depth in addressing implementation challenges, particularly in material selection optimization and frequency spectrum manipulation. Critically, it acknowledges and provides direction for key technical hurdles, including the scalability of the method to larger structures, the complexity of identifying materials with appropriate nonlinear dynamic responses, and the need for robust theoretical models to predict material behavior under musical frequency stimulation. The potential impact is amplified by its extension into biomedical applications, suggesting a bridge between materials science and medical diagnostics through music-inspired techniques. This comprehensive treatment of both theoretical foundations and practical challenges, coupled with the biomimetic inspiration and medical applications, establishes the proposal as a significant innovation in material property tuning methodologies, transcending traditional approaches to frequency-based material manipulation.

The innovation's fundamental strength emerges from its adherence to the abstract pattern $\alpha \rightarrow \beta \rightarrow \gamma \rightarrow \alpha$, where the cyclic relationship enables a unique feedback system. The proportionality rule $\alpha \propto \beta$ manifests in the biomimetic response mechanism, while the essential condition $\gamma \rightarrow \alpha$ is realized through the material's frequency-dependent behavior. This creates a dynamic system where music ($\alpha$) influences material properties ($\gamma$) through material response ($\beta$), while the resulting property changes create new response patterns, forming a continuous adaptive loop. As an interpretation, the relationship can be viewed as a coupled system where each component's influence is nonlinearly dependent on the others: 
\[
\frac{\partial \gamma}{\partial t} = f(\alpha, \beta) \hspace{0.5cm} \textrm{ and } \hspace{0.5cm} \frac{\partial \beta}{\partial t} = g(\gamma, \alpha), 
\]
capturing both the immediate response and long-term evolution of material properties under musical influence. This mathematical framework elegantly supports the proposal's practical applications while maintaining the fundamental symmetry of the original abstract pattern.

\subsection{\textit{In-situ} graph generation and recursive reasoning through multi-agent modeling}

The earlier example showed that the model was able to expand its capabilities beyond the materials-science focused training data and was able to successfully integrate different domains into the structured reasoning paradigm. In the next experiment, we task the model as follows:

\begin{LLMbox}{User task}
\begin{lstlisting}
(*@\bf\hlred{User:}@*) Integrate a snowflake and ant behavior to design a new tough material made from protein.
\end{lstlisting}
\end{LLMbox}

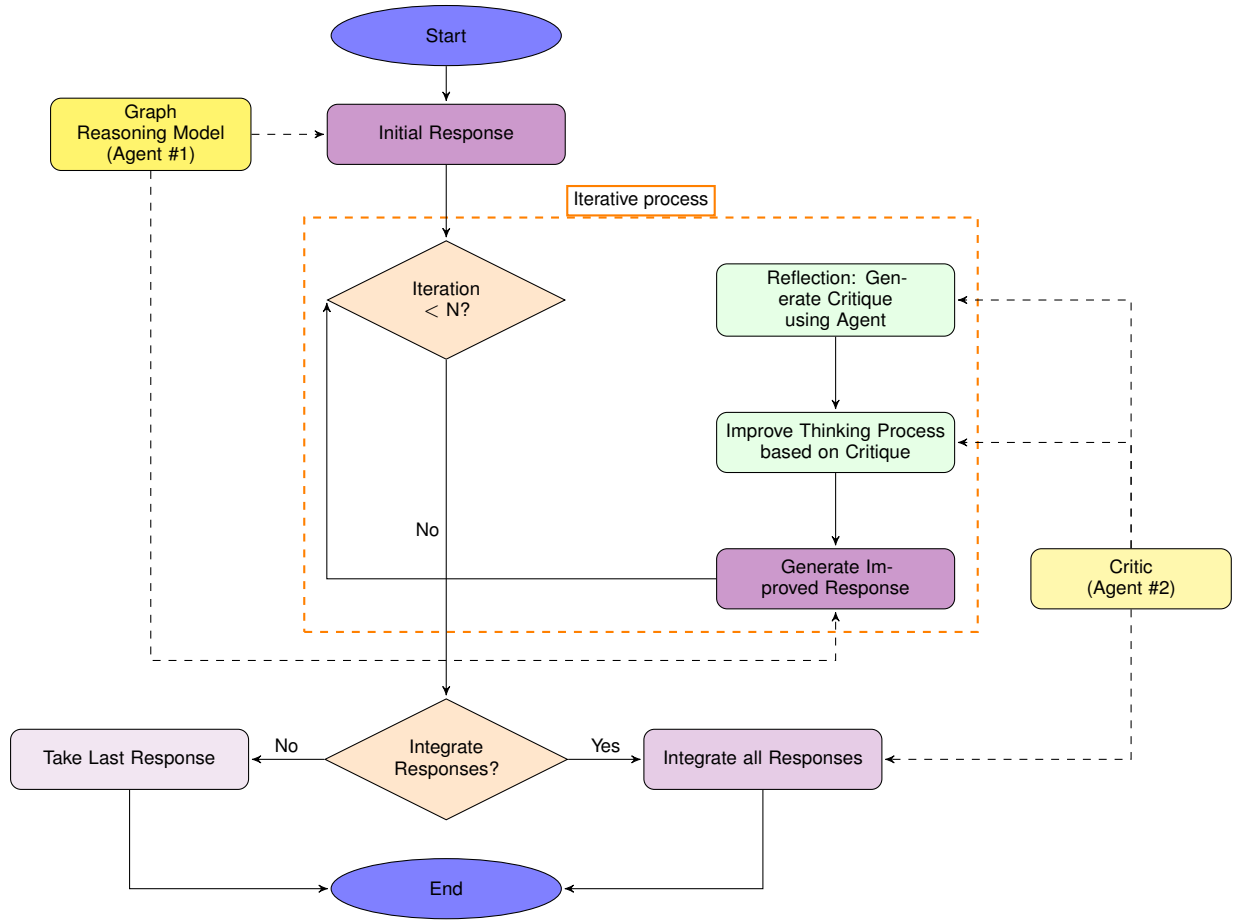
\begin{figure}
\sffamily 
\scriptsize
\sansmath 
\begin{tikzpicture}[node distance=1cm and 2cm, auto,
    block/.style={rectangle, draw, fill=violet!40, text width=3cm, text centered, rounded corners, minimum height=0.8cm},
    block2/.style={rectangle, draw, fill=violet!20, text width=3cm, text centered, rounded corners, minimum height=0.8cm},
    block3/.style={rectangle, draw, fill=violet!10, text width=3cm, text centered, rounded corners, minimum height=0.8cm},
    block4/.style={rectangle, draw, fill=green!10, text width=3cm, text centered, rounded corners, minimum height=0.8cm},
    decision/.style={diamond, draw, fill=orange!20, text width=2cm, text centered, inner sep=1pt, aspect=2},
    cloud/.style={draw, ellipse, fill=blue!50, text width=2cm, text centered, minimum height=0.8cm},
    model/.style={rectangle, draw, fill=yellow!40, text width=2.5cm, text centered, rounded corners, minimum height=0.8cm},
    model2/.style={rectangle, draw, fill=yellow!70, text width=2.5cm, text centered, rounded corners, minimum height=0.8cm},
    line/.style={draw, -stealth', shorten >=1pt},
    dashed line/.style={draw, dashed, -stealth', shorten >=1pt}]
    \node [cloud] (start) {Start};
    \node [block, below=of start, yshift=.5cm] (init) {Initial Response};
    \node [decision, below=of init] (iterate) {Iteration \\ $<$ N?};
    \node [block4, right=of iterate] (extract) {Reflection: Generate Critique \\using Agent};
    \node [block4, below=of extract] (improve) {Improve Thinking Process \\based on Critique};
    \node [block, below=of improve] (generate) {Generate Improved Response};
    \node [decision, below=of iterate, yshift=-3.5cm] (final_decision) {Integrate \\ Responses?};
    \node [block3, left=of final_decision, xshift=1cm] (last_response) {Take Last Response};
    \node [block2, right=of final_decision, xshift=-1cm] (integrate) {Integrate all Responses};
    \node [cloud, below=of final_decision, yshift=.5cm] (end) {End};
    \node [model2, left=of init, xshift=1cm] (reasoning) {Graph \\ Reasoning Model \\ (Agent \#1)};
    \node [model, right=of generate, xshift=-1cm] (critic) {Critic \\ (Agent \#2)};
    \path [line] (start) -- (init);
    \path [line] (init) -- (iterate);
    \path [line] (extract) -- (improve);
    \path [line] (improve) -- (generate);
    \path [line] (generate) -| (iterate.west);
    \path [line] (iterate) -- node [left] {No} (final_decision);
    \path [line] (final_decision) -- node [above] {No} (last_response);
    \path [line] (final_decision) -- node [above] {Yes} (integrate);
    \path [line] (last_response) |- (end);
    \path [line] (integrate) |- (end);
    \path [dashed line] (reasoning) -- (init);
    \path [dashed line] (reasoning) |- ([yshift=-7.cm]reasoning -| generate.south) -- (generate.south);
    \path [dashed line] (critic) |- (extract.east);
    \path [dashed line] (critic) |- (improve.east);
    \path [dashed line] (critic) |- (integrate);
    \begin{pgfonlayer}{background}
        \node [draw=orange, thick, dashed, fit=(iterate) (extract) (improve) (generate), inner sep=0.3cm, 
               label={[fill=white, draw=orange, thick]above:Iterative process}] {};
    \end{pgfonlayer}
\end{tikzpicture}
\sffamily 
\caption{Graph-PRefLexOR Recursive Reasoning Algorithm, using graph reasoning and abstract representations of relational mechanics, using a multi-agent system with Agent \#1 being the Graph Reasoning model, and Agent \#2 being a general-purpose critic model. The reflection is generated using the Critic agent and then used to improve the thinking process. This resembles an iterative approach leveraging the Reasoning Model and a general-purpose Critic Model to generate, refine, and optionally integrate responses. As before, the process ultimately involves generating initial responses, extracting reflections, improving thinking processes, and creating new responses based on refined thinking, with an optional final integration step. The algorithm relies on extracting thinking processes (indicated via \texttt{<|thinking|>}..\texttt{<|/thinking|>}) and reflection processes generated by the Critic. The sampled responses can either be used in their final state or integrated into an amalgamated response that shows very rich facets in the scientific process.  
}
\label{fig:PreFlexOR_recursive}

\end{figure}

However, unlike before where one can respond with a single shot, here we use recursive reasoning by using a two-agent setup where the thinking steps are critiqued, then improved, and fed back to the model. Ultimately, an integrated response is developed by the model that incorporates the various ideas, concepts, and details developed during the recursive process. Figure~\ref{fig:PreFlexOR_recursive} shows a flowchart of the process used.  In the example discussed below, we use $N=3$ iterations until the final answer is produced through multiple refinement steps. 

\begin{figure}
	\centering
	\includegraphics[width=1.\linewidth]{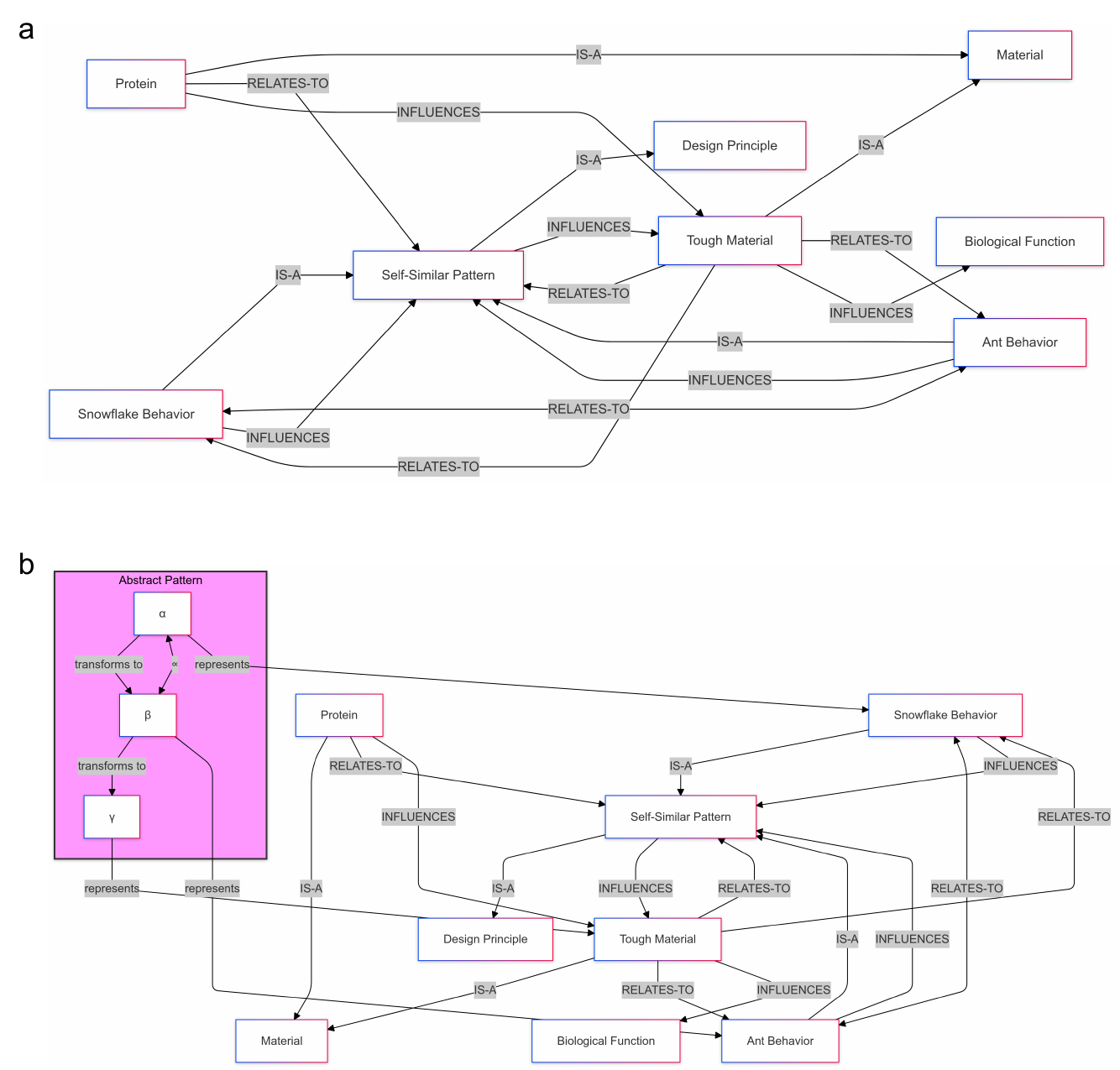}
        \caption{Overview of the graph data generated in response to the question: \texttt{Integrate a snowflake and ant behavior to design a new tough material made from protein.} The figure depicts both the knowledge graph alone (panel a) and an integration of the knowledge graph with the abstract pattern representation (panel b). Results after the first iteration, $i=1$, are shown.}
	\label{fig:fig_15}
\end{figure}

Each iteration in the algorithm produces an intermediate result that is critiqued and improved as delineated in Figure~\ref{fig:PreFlexOR_recursive}. The final answer after iteration $N=3$ is shown in Text Box~\ref{TexBox:final_answer_after_it_3}.

Once the maximum number of iterations has been reached, we develop an integrated answer that uses the responses from all three iterations to develop the final response to the task. For the example discussed here, the final, integrated answer is:

\begin{LLMbox}{Final integrated answer from recursive Graph-PReFLexOR}
\begin{lstlisting}
Designing a new tough material made from protein inspired by snowflake and ant behavior involves creating a protein-based material with fractal patterns and hierarchical structures, similar to snowflakes and ants, to enhance toughness, adaptability, and diversity in applications. The design principle for creating such a material is to mimic fractal patterns and hierarchical structures, inspired by snowflakes, and self-organization, inspired by ants, to enhance toughness, adaptability, and diversity in applications.

The key components of this design are:

1. **Fractal Patterns (Snowflake Inspiration):** Incorporate fractal patterns and self-similarity to create intricate, complex structures. Use branching patterns and scaling laws to create patterns at multiple length scales. This will enhance the material's toughness and adaptability.

2. **Self-Organization (Ant Inspiration):** Design the protein material to mimic the self-organization and hierarchical structures found in ant colonies. Encourage self-organization through hierarchical structures at multiple length scales. This will promote adaptability and diversity in the material's properties.

3. **Hierarchical Structure:** Incorporate multiple length scales to create complex, tough materials. This will enhance the material's toughness and adaptability.

The design process involves:

1. **Design Protein Material:** Create a protein-based material with fractal patterns and hierarchical structures. Incorporate self-organization and hierarchical structures at multiple length scales.

2. **Test and Refine:** Test the protein material's toughness, adaptability, and diversity in properties. Refine the design based on the test results to enhance the material's performance.

The material properties of the designed protein-based material are:

- **Fractal Pattern Mimicry:** Inspired by snowflakes, the protein material exhibits fractal patterns and self-similarity.
- **Hierarchical Structure:** Mimics ant behavior by creating hierarchical structures at multiple length scales.
- **Self-Organization:** Encourages self-organization, similar to ants, to enhance toughness and adaptability.
- **Toughness Enhancement:** Combines fractal patterns and hierarchical structures to create tough, protein-based materials.
- **Adaptability:** Designed for diverse applications by mimicking snowflake and ant patterns.

The potential applications of the designed protein-based material are:

- **Biomedical Applications:** Inspired by snowflake and ant patterns, protein-based materials can be used in biomedical applications such as tissue engineering, wound dressing, and drug delivery.
- **Biodegradable Materials:** Protein-based materials can be designed to be biodegradable, providing a sustainable alternative to traditional materials.
- **Soft Robotics:** Inspired by ant behavior, protein-based materials can be used in soft robotics, enabling more flexible and adaptable robotic systems.
- **Biomineralization:** Protein-based materials can mimic natural biomineralization processes, providing insights into bone and tooth formation.

The challenges and opportunities associated with designing protein-based materials inspired by snowflake and ant behavior are:

- **Scalability:** Maintaining fractal patterns and hierarchical structures at multiple length scales while scaling up the material.
- **Self-Organization:** Encouraging self-organization in protein materials to enhance toughness and adaptability.
- **Biocompatibility:** Ensuring the protein material is biocompatible and non-toxic for potential applications.
- **Synthesis Methods:** Developing efficient and scalable methods for synthesizing protein-based materials with fractal patterns and hierarchical structures.

Future research directions include:

- **Fractal Geometry of Protein Materials:** Investigating the fractal geometry of protein materials and its role in enhancing toughness and adaptability.
- **Ant-Inspired Protein Materials:** Developing protein-based materials that mimic ant behavior, such as self-organization and hierarchical structures.
- **Biodegradable Protein Materials:** Designing protein-based materials that are biodegradable and sustainable.
- **Synthesis Methods for Fractal Protein Materials:** Developing efficient and scalable methods for synthesizing protein-based materials with fractal patterns and hierarchical structures.

In conclusion, designing protein-based materials inspired by snowflake and ant behavior offers a promising approach to creating tough, adaptable, and diverse materials. By mimicking fractal patterns and hierarchical structures, and self-organization, protein materials can be engineered to enhance toughness, adaptability, and diversity in applications. However, challenges such as scalability, self-organization, biocompatibility, and synthesis methods need to be addressed. Future research should focus on fractal geometry, ant-inspired protein materials, biodegradable protein materials, and efficient synthesis methods to unlock the full potential of protein-based materials.
\end{lstlisting}
\end{LLMbox}

For a comparison, the answer from a standard (non-fine-tuned) model is shown in Text Box~\ref{TexBox:non-fine-tuned}. We note that this response does not include any thinking section (and hence, no graph reasoning or abstraction), and solely consists of the text as provided.

We first compare the results from the first iteration of the graph reasoning model with the standard model. From the standpoint of a scientist at a high technical level, the graph reasoning based response demonstrates superior intellectual depth and reasoning quality. The response provides a highly systematic framework for material design, detailing explicit parameters (e.g., protein selection, fractal pattern generation, hierarchical structure formation) and an iterative validation process that reflects rigorous scientific methodology and clear delineation in graph form (Figure~\ref{fig:fig_15}). This structured approach aligns well with the expectations of graduate-level research, as it includes clear constraints, performance criteria, and testing protocols akin to what one might find in peer-reviewed  scientific literature. By contrast, the response from the standard model showcases a more conceptual, broader-stroke approach and lacks the comprehensive methodological granularity seen in the other response. While it does incorporate topics such as self-healing peptides and adaptive camouflage, its overall elaboration remains more conceptual than procedural. Thus the graph reasoning result more closely meets the benchmarks of high rigor by systematically addressing design parameters, experimental validation, and scalability concerns.

We compare the two responses using~\texttt{GPT-4o} as evaluator, with results shown in Table~\ref{tab:response_comparison_with_graph}. Graph-PReFLexOR systematically addresses the design principles (fractals, hierarchical structures, self-organization) and walks through potential challenges (scalability, synthesis methods, biocompatibility) with a more methodical cause-and-effect flow and hence offers a deeper theoretical reasoning overall. In contrast, the comparison with the non-fine tuned model is much shorter and more general. It does not dig as deeply into the theoretical underpinnings, deep materials-specific reasoning, or longer-term challenges. The graph-based response offers more robust, multi-layered reasoning. 

\begin{table}[h!]
\centering
\small
\begin{tabular}{|p{4cm}|p{2cm}|p{2cm}|p{7cm}|}
\hline
\textbf{Criteria}              & \textbf{Graph reasoning response (Response 1)} & \textbf{Reference model (Response 2)} & \textbf{Explanation} \\ \hline
Intellectual Depth             & 10/10                & 7/10                & Response 1 combines text and the graph, providing a multi-layered, systematic approach to design. Response 2 lacks visual support and focuses on textual descriptions. \\ \hline
Reasoning Quality              & 10/10                & 8/10                & The graph in Response 1 supports robust reasoning, connecting abstract concepts to practical design principles. Response 2 demonstrates logical reasoning but without visual reinforcement. \\ \hline
Logic                          & 10/10                & 7/10                & Response 1’s graph enhances the logical connections between abstract patterns, protein structures, and material properties. Response 2 lacks a structured framework to visually link ideas. \\ \hline
Use of Academic Measures       & 10/10               & 6/10                & Response 1 leverages the graph to convey systematic relationships and integrates validation criteria effectively. Response 2 is descriptive and less rigorous in methodology. \\ \hline
Clarity and Organization       & 8/10                & 9/10                & While Response 1 provides depth, the graph may overwhelm casual readers. Response 2 is simpler and more accessible but less comprehensive. \\ \hline
Potential Applicability        & 10/10                & 8/10                & Response 1 explicitly ties abstract and biological principles to practical applications, supported by the graph. Response 2 mentions potential applications but without detailed pathways. \\ \hline
Overall Integration of Graph   & 10/10                & 0/10                 & The graph in Response 1 enhances intellectual depth, reasoning, and academic rigor. Response 2 lacks a graph entirely. \\ \hline
Overall Score                  & 9.7/10              & 6.4/10              & Response 1 excels in intellectual depth, reasoning, and academic rigor due to the graph and systematic methodology. Response 2, while clear and concise, falls short in depth and structure. \\ \hline
\end{tabular}
\caption{Comparison of responses, as obtained using \texttt{GPT-4o} as a judge. Graph-PReFLexOR responses are delineated as `Response 1', and the reference model as `Response 2'.  The graph reasoning model performs much better overall, with an average score of 9.7 vs. 6.4.}
\label{tab:response_comparison_with_graph}
\end{table}

\subsection{Growing a knowledge garden via \textit{in-situ} graph generation and knowledge expansion}

In this next experiment we use the capability of the graph reasoning model to generate graphs that describe phenomena in a relational manner and apply it to explore ever-expanding graph structures by repeatedly prompting the model. This allows us to recursively grow a graph, starting from an initial concept or response to a task, and to examine whether the model can blend creative and analytical reasoning. The initial task is:

\begin{LLMbox}{User task}
\begin{lstlisting}
(*@\bf\hlred{User:}@*) Write a song by incorporating snow flakes and flower petals.
\end{lstlisting}
\end{LLMbox}

\begin{LLMbox}{Knowledge graph}
\begin{lstlisting}
**Knowledge Graph:**

1. **Snow Flakes** -[IS-A]-> **Natural Phenomenon**
2. **Flower Petals** -[IS-A]-> **Natural Phenomenon**
3. **Snow Flakes** -[RELATES-TO]-> **Winter**
4. **Flower Petals** -[RELATES-TO]-> **Spring**
5. **Snow Flakes** -[INFLUENCES]-> **Mood** (e.g., Serene, Calming)
6. **Flower Petals** -[INFLUENCES]-> **Mood** (e.g., Delicate, Whimsical)
7. **Snow Flakes** -[INFLUENCES]-> **Nature's Beauty**
8. **Flower Petals** -[INFLUENCES]-> **Nature's Beauty**
9. **Snow Flakes** -[INFLUENCES]-> **Seasonal Changes**
10. **Flower Petals** -[INFLUENCES]-> **Seasonal Changes**
\end{lstlisting}
\end{LLMbox}

The abstract patterns are:

\begin{LLMbox}{Abstract Patterns}
\begin{lstlisting} 
**Abstract Pattern:**

(*@$\alpha \rightarrow \beta$@*) 
(*@$\beta \rightarrow \gamma$@*) 
(*@$\alpha \rightarrow \delta$@*) 
(*@$\beta \rightarrow \delta$@*) 
(*@$\alpha \rightarrow \varepsilon$@*) 
(*@$\beta \rightarrow \varepsilon$@*) 
(*@$\alpha \rightarrow \zeta$@*)
(*@$\beta \rightarrow \zeta$@*)

**Key Transformation Rule:**

If (*@$\alpha \rightarrow \delta$@*) and (*@$\beta \rightarrow \delta$@*) then (*@$\delta \rightarrow \varepsilon$@*) 

**Essential Condition:**

(*@$\alpha \neq \beta$@*) 

Pattern Context:
The song incorporates snowflakes and flower petals to represent the contrasting yet harmonious aspects of nature's beauty and seasonal changes. Snowflakes symbolize the serenity and uniqueness of winter, while flower petals signify the delicacy and renewal of spring. Together, they highlight the dynamic interplay between these elements, creating a balanced and evocative musical composition.
\end{lstlisting}
\end{LLMbox}

\begin{figure}
	\centering
	\includegraphics[width=1.\linewidth]{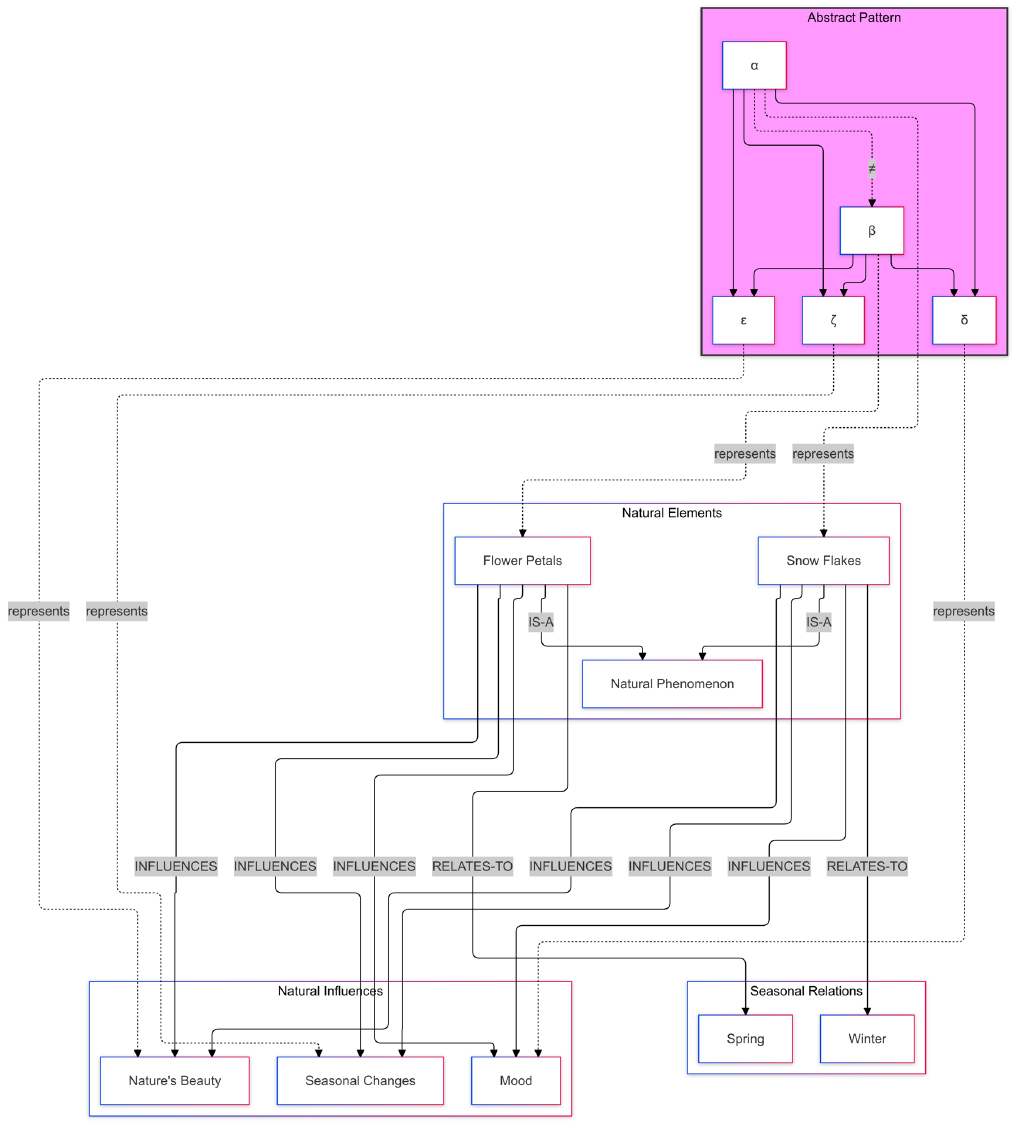}
        \caption{Resulting graph in response to the task: \texttt{Write a song by incorporating snow flakes and flower petals.} The visualization depicts the knowledge graph integrated with the abstract pattern representation.}
	\label{fig:fig_25}
\end{figure}

This task results in the graph as depicted in Figure~\ref{fig:fig_25}.  There are other elements in the response, as reflected here:

\begin{LLMbox}{Reasoning steps}
\begin{lstlisting} 
**Reasoning Steps**:

1. Snowflakes and flower petals are both natural phenomena.
2. Snowflakes are associated with winter, while flower petals are associated with spring.
3. Snowflakes evoke a sense of serenity and uniqueness, while flower petals symbolize delicacy and renewal.
4. Combining snowflakes and flower petals in a song can create a harmonious representation of nature's beauty and seasonal changes.
5. This combination can inspire a balanced and evocative musical composition.

**Relevant Materials or Concepts**:

- Snowflakes
- Flower Petals
- Nature's Beauty
- Seasonal Changes
- Serenity
- Uniqueness
- Delicacy
- Renewal

**Design Principles**:

- **Contrast**: Snowflakes and flower petals represent opposing seasons, creating a striking contrast.
- **Harmony**: Together, they create a harmonious representation of nature's beauty and seasonal changes.
- **Evocative**: Each element evokes distinct emotions and sensations, inspiring a balanced and evocative musical composition.

**Material Properties**:

- **Snowflakes**:
  - **Appearance**: Delicate, intricate, crystalline structure
  - **Sound**: Soft, gentle, repetitive
  - **Emotional Connotation**: Serene, calming, unique

- **Flower Petals**:
  - **Appearance**: Delicate, colorful, soft
  - **Sound**: Light, airy, whimsical
  - **Emotional Connotation**: Delicate, whimsical, renewal

**Hypothesis**:

"Incorporating snowflakes and flower petals in a song will create a harmonious representation of nature's beauty and seasonal changes, evoking distinct emotions and sensations."

**Additional Background**:

To incorporate snowflakes and flower petals into a song, one could consider the following elements:

1. **Imagery**: Use vivid and evocative language to describe the intricate patterns and beauty of snowflakes and flower petals. This can be achieved through lyrics or even instrumental descriptions.

2. **Metaphors**: Draw parallels between the delicate nature of snowflakes and flower petals and the human experience. For instance, snowflakes could represent individuality and uniqueness, while flower petals could symbolize growth and renewal.

3. **Seasonal themes**: Explore the contrast between winter (snowflakes) and spring (flower petals) to create a narrative that highlights the cyclical nature of life and the beauty in each season.

4. **Musical motifs**: Use specific musical motifs or melodies to evoke the feeling of snowflakes (e.g., soft, repetitive patterns) and flower petals (e.g., light, airy, whimsical).

5. **Collaboration**: Consider co-writing with a poet or lyricist to craft lyrics that incorporate these natural elements in a meaningful and impactful way.
[...]
\end{lstlisting}
\end{LLMbox}

The final response after the thinking section is: 

\begin{LLMbox}{Final response}
\begin{lstlisting} 
**Song: "Snowflakes and Flower Petals"**

**Verse 1**
In winter's chill, where snowflakes fall
Each one unique, a delicate call
Dance upon the breeze, a fleeting sight
A moment's beauty, lost in the night

**Chorus**
But even in the cold, there's warmth to find
In the petals of the flower, left behind
Delicate, colorful, soft to the touch
A symbol of renewal, in the winter's hush

**Verse 2**
In spring's awakening, where petals unfold
A tapestry of colors, a story to be told
Whimsical, light, a gentle caress
A promise of new life, in every tender dress

**Chorus**
But even in the warmth, there's serenity to find
In the snowflakes that fall, in the stillness of the mind
Unique, a work of art, a fleeting sight
A symbol of peace, in the quiet of the night

**Bridge**
In the contrast, we find harmony
A balance of opposites, a symphony
Snowflakes and flower petals, a dance so fine
A celebration of life, in all its seasons' rhyme

**Chorus**
In the snowflakes and flower petals, we find
A reflection of ourselves, in nature's design
A reminder to cherish, each moment we share
A symbol of hope, in the beauty that's always there
\end{lstlisting}
\end{LLMbox}

The answer also includes additional facets and analyses, as follows. These additional details and rational developments are critical for the proposed approach as they provide a traceable, detailed and interpretable analysis of how the answers were developed. 

\begin{LLMbox}{Design principles and background}
\begin{lstlisting} 
**Design Principles**:

- **Contrast**: Snowflakes and flower petals represent opposing seasons, creating a striking contrast.
- **Harmony**: Together, they create a harmonious representation of nature's beauty and seasonal changes.
- **Evocative**: Each element evokes distinct emotions and sensations, inspiring a balanced and evocative musical composition.

**Material Properties**:

- **Snowflakes**:
  - **Appearance**: Delicate, intricate, crystalline structure
  - **Sound**: Soft, gentle, repetitive
  - **Emotional Connotation**: Serene, calming, unique

- **Flower Petals**:
  - **Appearance**: Delicate, colorful, soft
  - **Sound**: Light, airy, whimsical
  - **Emotional Connotation**: Delicate, whimsical, renewal

**Hypothesis**:

"Incorporating snowflakes and flower petals in a song will create a harmonious representation of nature's beauty and seasonal changes, evoking distinct emotions and sensations."

**Additional Background**:

To incorporate snowflakes and flower petals into a song, consider using metaphors, imagery, and musical motifs to evoke the emotions and sensations associated with these natural elements. You could also explore the contrast between the two elements to create a narrative that highlights the cyclical nature of life and the beauty in each season. Additionally, consider using seasonal themes and lyrics that reflect the unique characteristics of each element to create a rich and evocative musical composition.
\end{lstlisting}
\end{LLMbox}

While this section focuses on growing the graph iteratively, the depth of the response merits further analysis as it sheds light on the capabilities of the model. First, we note that the model's approach to the musical composition task demonstrates sophisticated pattern recognition 
capabilities through multiple analytical layers, and a capability to blend creative and analytical reasoning. For a deeper analysis, let $\mathcal{M}$ represent the model's mapping function, where $\mathcal{M}: \text{Natural Elements} \rightarrow \text{Musical Structure}$. The model proposes that the primary recognition pattern can be formalized as $\mathcal{P}(\alpha, \beta) = \{(\alpha \rightarrow x) \iff (\beta \rightarrow x) \mid x \in \text{Influences}\}$, where $\alpha$ represents snowflakes and $\beta$ represents flower petals. This symmetrical mapping demonstrates the model's ability to identify and preserve parallel structures while maintaining essential distinctions.

We can see that the model's pattern recognition operated on multiple levels: $\mathcal{L}_1: \text{Categorical} 
\rightarrow \{\text{Natural Phenomenon}\}$, $\mathcal{L}_2: \text{Temporal} \rightarrow 
\{\text{Winter}, \text{Spring}\}$, and $\mathcal{L}_3: \text{Influence} \rightarrow \{\text{Mood}, 
\text{Nature's Beauty}, \text{Seasonal Changes}\}$. The abstraction process generated a formal 
framework where $\forall x \in \mathcal{L}_3: (\alpha \rightarrow x) \land (\beta \rightarrow x)$ 
while maintaining the essential condition $\alpha \neq \beta$. This demonstrates sophisticated 
categorical reasoning while preserving compositional possibilities.

The response reveals remarkable symmetrical structures between winter and spring phenomena, characterized 
by parallel relationships that can be formalized as $\mathcal{S} = \{(s_w, s_p) \in \text{Phenomena} 
\times \text{Seasons}\}$, where snowflakes ($s_w$) and flower petals ($s_p$) exhibit isomorphic influence 
patterns. Each element maintains identical categorical relationships, with $(s_w, s_p) \in 
\text{Natural Phenomenon}$, while preserving distinct seasonal associations: $s_w \mapsto \text{Winter}$, 
$s_p \mapsto \text{Spring}$. The influence structure demonstrates perfect symmetry across three domains, 
expressible as $\forall x \in \{\text{Mood}, \text{Nature's Beauty}, \text{Seasonal Changes}\}: 
(s_w \rightarrow x) \iff (s_p \rightarrow x)$. This symmetry manifests despite the temporal opposition 
of their respective seasons, suggesting a fundamental balance in nature's organizational structure. The 
relationship can be further abstracted into a commutative diagram where both elements, while maintaining 
$s_w \neq s_p$, exhibit identical transformative properties: $\mathcal{T}: \{s_w, s_p\} \times 
\{\text{Mood}, \text{Nature's Beauty}, \text{Seasonal Changes}\} \rightarrow \{0,1\}$. This structural 
symmetry underlies the aesthetic and phenomenological balance observed in seasonal transitions.

As shown in Figure~\ref{fig:fig_25}, the responses introduce an abstract pattern framework $\mathcal{P} = \{\alpha, \beta, \delta, \epsilon, \zeta\}$ where 
$\alpha, \beta$ represent distinct natural phenomena ($\alpha \neq \beta$) that exhibit parallel transformative 
properties. The system is characterized by a set of mappings $\{\alpha, \beta\} \rightarrow \{\delta, \epsilon, \zeta\}$, 
with each mapping representing a direct transformative relationship. The key transformation rule 
$(\alpha \rightarrow \delta) \land (\beta \rightarrow \delta) \implies (\delta \rightarrow \epsilon)$ establishes a 
conditional cascade effect. This creates a unique convergent structure where two distinct source elements 
$\{\alpha, \beta\}$ maintain symmetric relationships with target elements while preserving their fundamental 
inequality. The pattern demonstrates both first-order transformations 
$T_1: \{\alpha, \beta\} \times \{\delta, \epsilon, \zeta\} \rightarrow \{0,1\}$ and second-order conditional  transformations $T_2: \delta \times \epsilon \rightarrow \{0,1\}$ under specific antecedent conditions, forming  a hierarchical transformation framework.

The model prioritized structural understanding over domain-specific musical application, evidenced by 
its construction of the abstract framework $\mathcal{P} = \{\alpha, \beta, \delta, \epsilon, \zeta\}$ 
before considering musical implementation. This approach demonstrates strong logical reasoning but 
suggests potential limitations in capturing the temporal and emotional nuances essential to musical 
composition. The symmetrical framework, while mathematically elegant ($\forall x \in \mathcal{L}_3: 
s_w \rightarrow x \iff s_p \rightarrow x$), might benefit from additional parameters to capture the 
dynamic nature of musical expression.

\subsubsection{Knowledge expansion growth phase}

\begin{figure}[h!]
\centering
\begin{tikzpicture}[
    node distance=.75cm and .75cm,
    prompt/.style={rectangle, draw=blue!50, fill=blue!10, thick, text width=12cm, align=center, rounded corners, inner sep=5pt},
    line/.style={draw, thick, ->, >=stealth, black!70},
    every node/.append style={font=\footnotesize}
]

\node[prompt] (P1) {1. Write a song by incorporating snowflakes and flower petals.};
\node[prompt, below=of P1] (P2) {2. How do the molecular symmetry principles that create snowflake patterns influence nature's beauty in other seasonal phenomena?};
\node[prompt, below=of P2] (P3) {3. How does the cyclical pattern of seasonal changes shape human understanding of impermanence and renewal?};
\node[prompt, below=of P3] (P4) {4. What role does nanopatterning play in flower petals?};
\node[prompt, below=of P4] (P5) {5. Make a connection between 'Self-Recognition and Adhesion Behavior' and Magister Ludi in Hesse's Glass Bead Game.};
\node[prompt, below=of P5] (P6) {6. Make a connection between 'Glass Beads' and 'Nanopatterning' in the context of music.};
\node[prompt, below=of P6] (P7) {7. Propose an integration of fracture and music, considering nanopatterning.};

\path[line] (P1) edge (P2);
\path[line] (P2) edge (P3);
\path[line] (P3) edge (P4);
\path[line] (P4) edge (P5);
\path[line] (P5) edge (P6);
\path[line] (P6) edge (P7);

\begin{scope}[on background layer]
    \node[draw=blue!50, rounded corners, inner sep=10pt, fit=(P1)(P7)] (bg) {};
\end{scope}

\end{tikzpicture}
\caption{Logical sequential flow of prompts used to grow the knowledge graph successively through recursive reasoning. This diagram illustrates the progression of ideas, starting with creative tasks (e.g., composing a song using snowflakes and flower petals) and advancing through scientific and philosophical inquiries, such as molecular symmetry, seasonal patterns, and nanopatterning. The sequence of questions can either be generated by human collaborators, or by AI. In the latter case, an `infinite' loop of expanding the knowledge graph can be constructed.  The example prompt trajectory shown here was developed via human-AI collaboration. }
\label{fig:prompts-flow}
\end{figure}
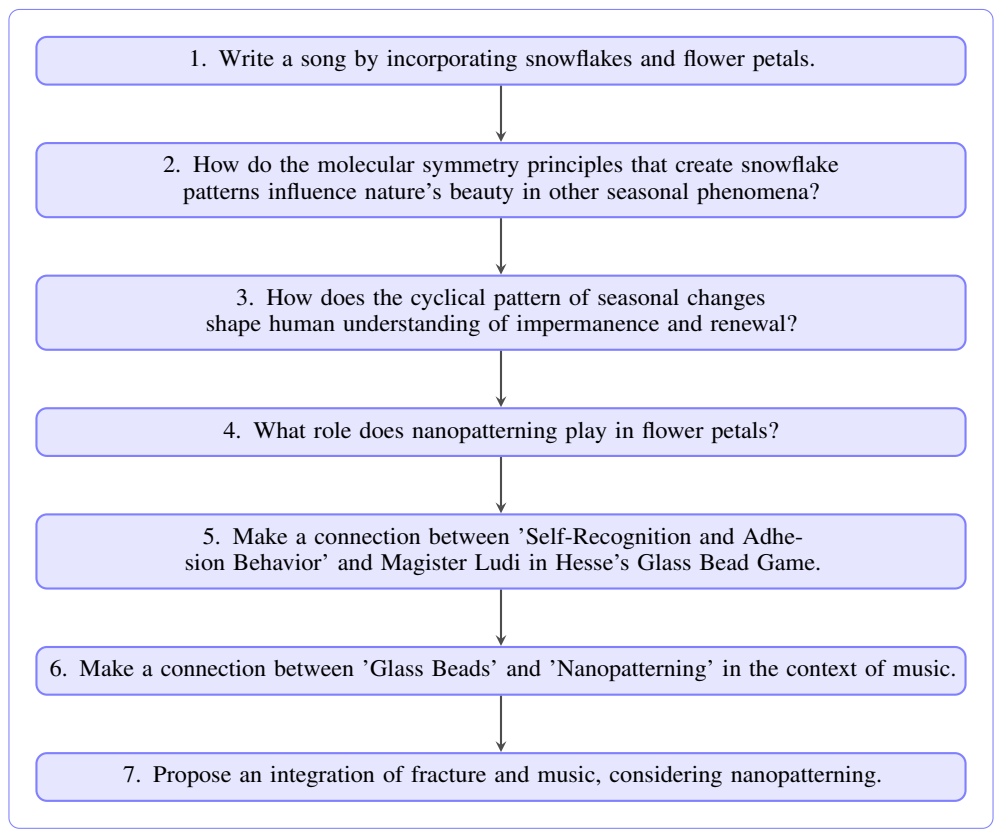

An interesting aspect of the capability of our graph reasoning model is  its ability to grow graphs dynamically to extract, generate and design new knowledge.  To do this we iteratively task the model to generate graphs, where new prompts are developed based on earlier ones to expand the answer in new directions. Since the model produces a graph at every iteration, all these graphs can ultimately be integrated into one large graph that spans diverse topics. We use the following series of prompts as visualized in Figure~\ref{fig:prompts-flow}.

\begin{figure}
	\centering
	\includegraphics[width=.9\linewidth, angle=90]{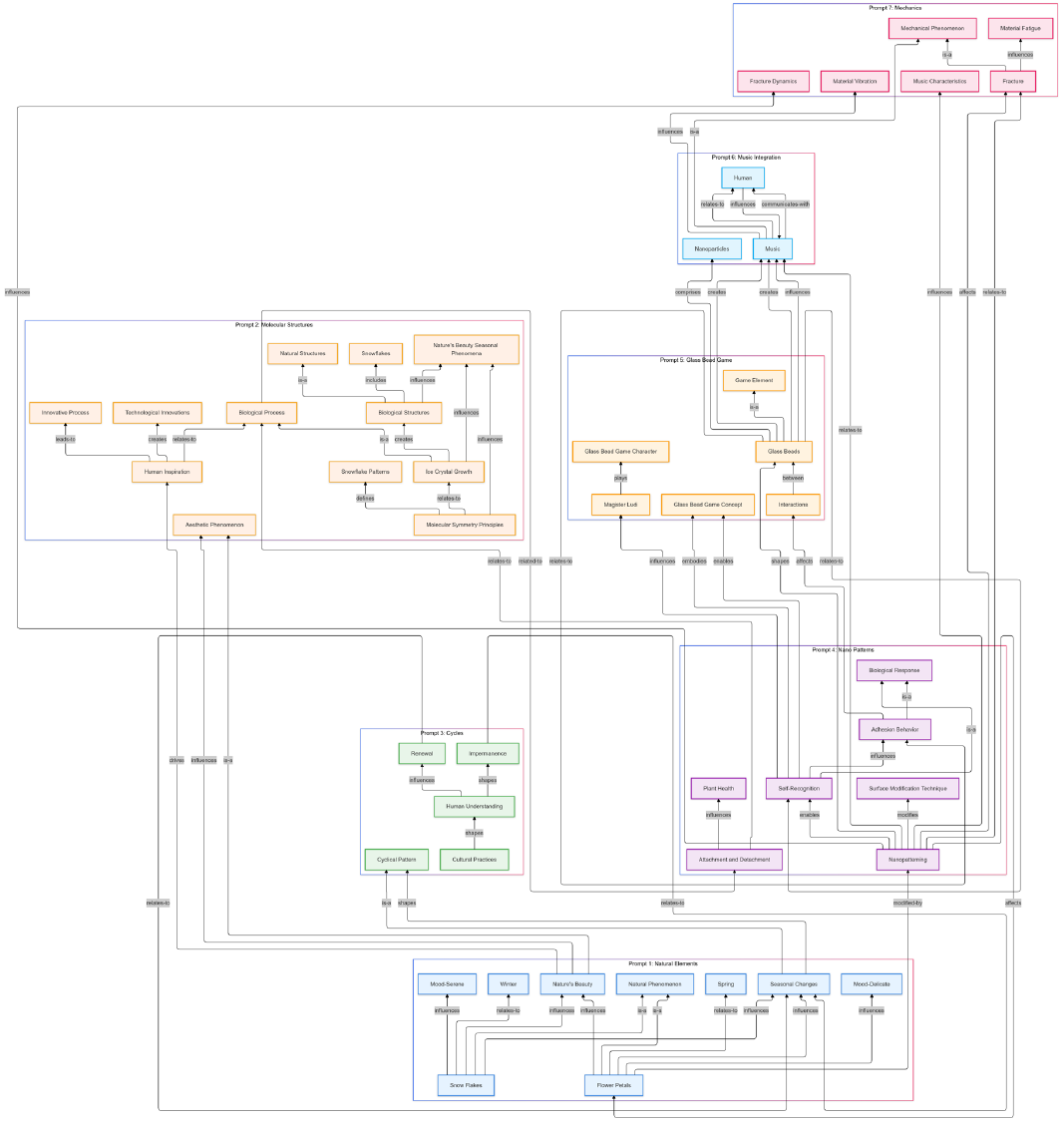}
        \caption{Visualization of the resulting integrated knowledge graph that emerges after the series of prompts delineated in Figure~\ref{fig:prompts-flow}. To start, we started tasking the model to \texttt{Write a song by incorporating snowflakes and flower petals.} The resulting graphs are organized by prompts but we show the connections that emerge between the sub-graphs.}
	\label{fig:fig_11}
\end{figure}

Figures~\ref{fig:fig_11} and~\ref{fig:fig_12} present complementary visualizations of the concatenated knowledge graphs derived from the series of prompts outlined in Figure~\ref{fig:prompts-flow}. Figure~\ref{fig:fig_11} organizes the knowledge graphs by individual prompts, emphasizing the structure and interconnections within each sub-graph. This visualization highlights the local relationships and the specific thematic focus of each prompt while illustrating how these ideas are interconnected across different sub-graphs. In contrast, Figure~\ref{fig:fig_12} integrates the same data into a unified knowledge graph, providing a holistic view of the interplay among all the prompts. By merging the sub-graphs into a single structure, this representation reveals overarching connections and emergent patterns that may not be immediately apparent when examining the sub-graphs in isolation. Together, these figures offer a layered perspective on the relationships within the dataset, from prompt-specific insights to a system-level understanding.

\begin{figure}
	\centering
	\includegraphics[width=0.85\linewidth,  ]{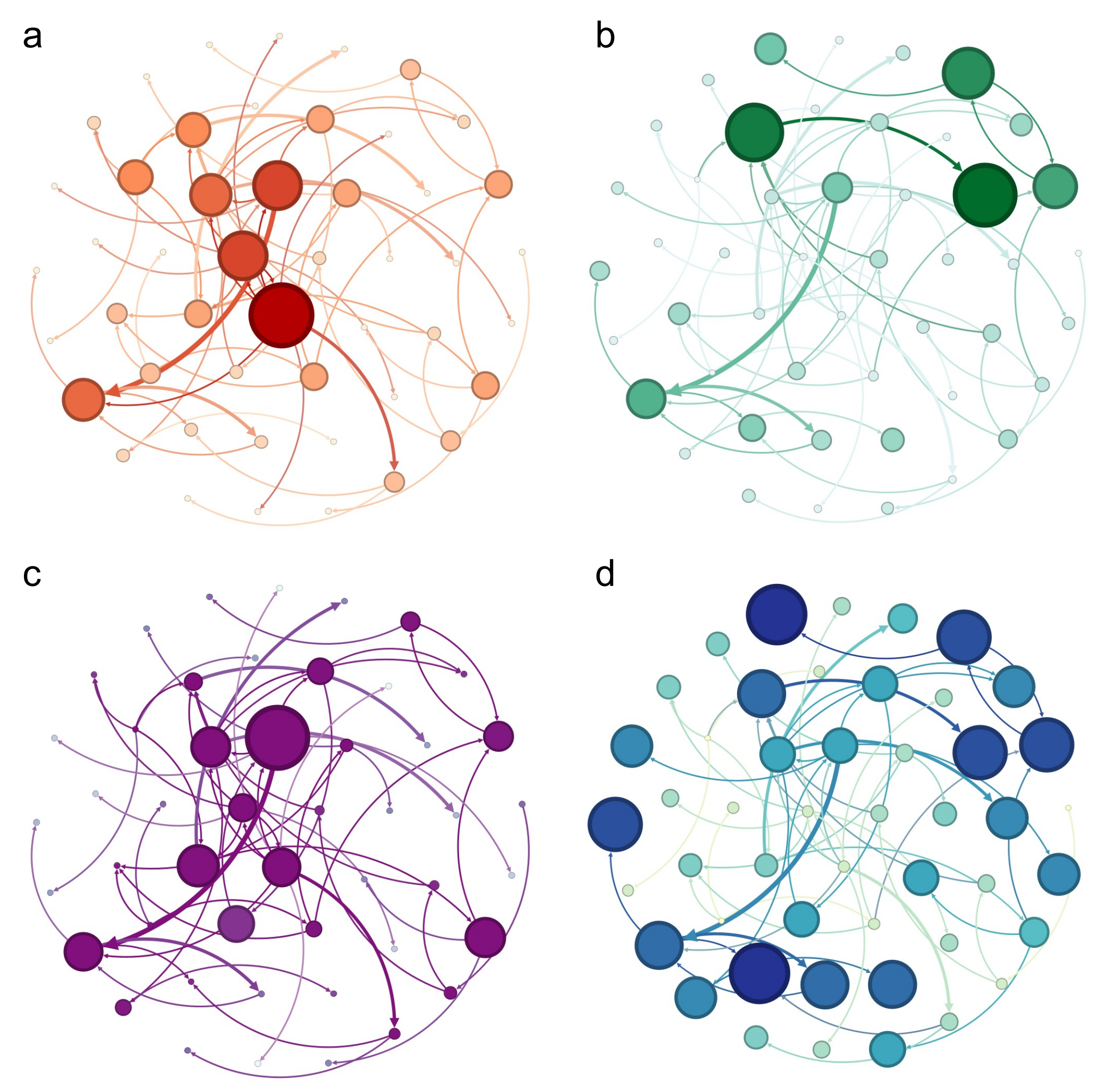}
        \caption{Visualization of the integrated knowledge graph created based ion the series of prompts delineated in Figure~\ref{fig:prompts-flow}, starting with: \texttt{Write a song by incorporating snowflakes and flower petals.} The data is the same as shown in Figure~\ref{fig:fig_11}, but organized here as an integrated graph rather than by prompt, and laid out using the Fruchterman Reingold layout algorithm~\cite{Fruchterman1991}. Panel a, node size by node degree. Panel b, node size by page rank. Panel c, node size by bridging coefficient. Panel d, node size by domain prestige (metric defined by fraction of nodes within a network that are directly od indirectly pointing to it). Table~\ref{tab:node_metrics} lists the top 5 nodes per each of these measures.}
	\label{fig:Fig_12_abstract}
\end{figure}

Figure~\ref{fig:Fig_12_abstract} depicts the graph in a more abstract representations, visualized per distinct layouts and different node properties.  Data is shown in node degree, PageRank~\cite{Brin1998}, and bridging coefficient displays. These are fundamental measures in network analysis, each capturing different aspects of a node's role within a graph, where the analyses point to the powerful insights that can be generated based on the model output. Node degree represents the number of connections a node has, providing a simple yet powerful indicator of local importance; in directed graphs, this can be further divided into in-degree (incoming connections) and out-degree (outgoing connections). 

As an alternative, PageRank offers a more global perspective, measuring a node's importance based on the significance of its neighbors. Originally developed for ranking web pages, PageRank emphasizes the quality of connections, assigning higher scores to nodes linked by other highly-ranked nodes. PageRank is a measure of a node's global importance in a network, originally designed to rank web pages in search engines. It assigns a score to each node based on the principle that connections from important nodes contribute more to a node's rank than connections from less important nodes. The algorithm works iteratively, where a node's PageRank is proportional to the sum of the PageRank scores of its neighbors, weighted by the number of links those neighbors have. This approach captures both the quantity and quality of connections, emphasizing nodes that are linked by highly influential entities. A damping factor is typically introduced to account for random jumps, ensuring the scores converge and the network can handle dangling nodes (nodes with no outgoing links). In the context of knowledge graphs, PageRank is particularly relevant for identifying the most semantically important entities or concepts. For instance, in a knowledge graph where nodes represent entities and edges represent relationships, PageRank can highlight nodes that are central to the graph’s structure, such as frequently referenced concepts or pivotal entities connecting different subdomains. This can aid in prioritizing entities for tasks like question answering, reasoning, or graph traversal. Moreover, PageRank can uncover influential nodes that might serve as anchors for linking related knowledge, enhancing both the efficiency of queries and the understanding of the graph’s overall topology.

The bridging coefficient highlights a node’s role in connecting otherwise unconnected or loosely connected regions of the graph. Nodes with high bridging coefficients serve as critical links between clusters, facilitating information flow and structural integration. Together, these metrics provide complementary insights into the structure and dynamics of complex networks. The bridging coefficient measures how well a node serves as a bridge between otherwise unconnected or loosely connected regions of a graph, emphasizing the node's role in maintaining structural integrity and facilitating information flow. In the context of knowledge graphs, nodes with high bridging coefficients are often critical for linking distinct clusters of knowledge, such as different domains or subgraphs. These bridging nodes play a crucial role in enabling cross-domain reasoning and ensuring the graph remains well-connected. For example, in a knowledge graph representing scientific research, a bridging node might connect distinct areas like biology and materials science, fostering interdisciplinary insights. By identifying nodes with high bridging coefficients, one can uncover concepts or entities that act as gateways to new knowledge, prioritize areas for graph enrichment, or ensure robustness in query pathways. This measure is particularly useful for tasks like ontology alignment, where structural gaps between subdomains need to be bridged effectively.

\begin{figure}
	\centering
	\includegraphics[width=0.85\linewidth,  ]{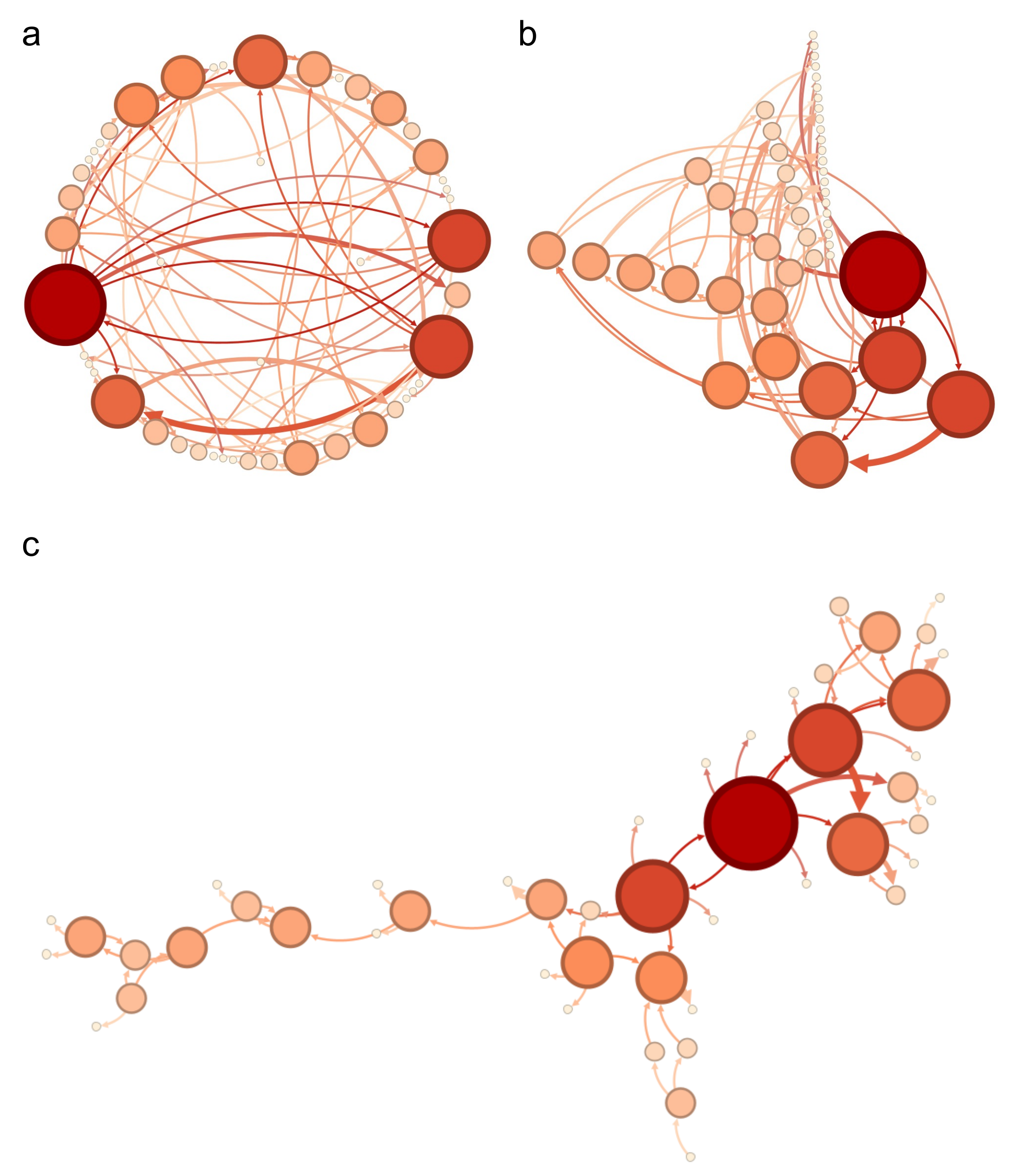}
        \caption{Visualization of the integrated knowledge graph created based ion the series of prompts delineated in Figure~\ref{fig:prompts-flow} (starting with: \texttt{Write a song by incorporating snowflakes and flower petals.}). The underlying data is the same as shown in Figure~\ref{fig:fig_11}, but organized here effectively via circular layout (panel a), radial axis layout (panel b), and Yifan Hu layout~\cite{Hu2005} (panel c). Color and size of each node is scaled by node degree. The layout in panel c visualizes the great distance traversed in the integrated graph, providing far-ranging relationships developed by the model. }
	\label{fig:Fig_12_variant}
\end{figure}

Domain prestige is a measure of how accessible a node is within a directed graph, calculated as the fraction of all nodes that can reach a given node either directly or indirectly. It captures a node’s global influence by accounting for indirect paths, providing a more comprehensive view of its accessibility. In the context of knowledge graphs, this measure highlights key entities that are reachable by a large portion of the graph, making them central for reasoning, inference, or information propagation. Nodes with domain prestige act as crucial hubs or reference points, while nodes with low values are likely isolated or peripheral. This metric is particularly valuable for identifying entities that facilitate connectivity and play pivotal roles in the overall graph structure. Table~\ref{tab:node_metrics} lists the top 5 nodes for each of these measures, complementing the visual representation in Figure~\ref{fig:Fig_12_abstract} to indicate the most significant nodes in each case and metric. For alternative layouts, Figure~\ref{fig:Fig_12_variant} depicts various layout choices for comparison. 
 
\begin{table}[h!]
\centering
\small
\begin{tabular}{|l|l|l|l|}
\hline
\textbf{Node Degree} & \textbf{Page Rank}         & \textbf{Bridging Coefficient} & \textbf{Domain Prestige} \\ \hline
Music                & Cyclical Pattern           & Spring  &  Plant Health  \\ \hline
Self-Recognition     & Seasonal Changes           &  Mood-Delicate  &  Mechanical Phenomenon    \\ \hline
Flower Petals        & Attachment and Detachment  & Surface Modification Technique    & Cyclical Pattern   \\ \hline
Glass Beads          & Biological Process         & Fracture Dynamics  & Material Vibration  \\ \hline
Nanopatterning       & Music                      & Music Characteristics  &  Attachment and Detachment \\ \hline
\end{tabular}
\caption{Comparison of top nodes in the generated knowledge graph for different measures, following the four panels in Figure~\ref{fig:Fig_12_abstract}. }
\label{tab:node_metrics}
\end{table}

The knowledge expansion process showcased in these repeated prompts demonstrates the ability of Graph-PReFLexOR to iteratively generate and connect knowledge graphs based on a series of related prompts. Beginning with a creative task (writing a song incorporating snowflakes and flower petals), the model progressively expands its knowledge base by addressing scientific and philosophical inquiries stemming from the initial prompt. Some unique insights/ideas generated during the recursive knowledge expansion process include:

\begin{itemize}
    \item Symmetrical structures in nature: The model identifies symmetrical influence patterns between snowflakes (associated with winter) and flower petals (associated with spring). Despite their temporal opposition, both exhibit isomorphic influence on mood, nature’s beauty, and seasonal changes. This reveals a fundamental balance in nature's organization across different seasons.
    
    \item Abstraction of natural phenomena: The model constructs an abstract pattern framework representing the relationships between distinct natural phenomena. This framework highlights both first-order and second-order conditional transformations, demonstrating the model's capacity for high-level reasoning and abstraction.
    
    \item Integration of diverse concepts: The knowledge graph expands to incorporate concepts like molecular symmetry principles, nanopatterning in flower petals, and the philosophical idea of impermanence and renewal, showcasing the ability to connect ideas across diverse disciplines from materials science to music to literature and philosophy~\cite{hesse2000glass,Ziolkowski1965} .
   
    \item Emergent patterns: By visualizing the integrated knowledge graph, overarching connections and patterns emerge that might not be apparent when examining individual sub-graphs. This highlights the system's ability to synthesize knowledge and reveal hidden relationships.
     
\end{itemize}

The knowledge expansion process exemplifies the strength of Graph-PReFLexOR in generating a network of interconnected ideas, moving beyond isolated responses to build a complex and evolving knowledge base. This capability holds significant potential for interdisciplinary research and exploration.

\subsection{Autonomously growing knowledge garden}

\begin{figure}
	\centering
	\includegraphics[width=1.\linewidth, 
    ]{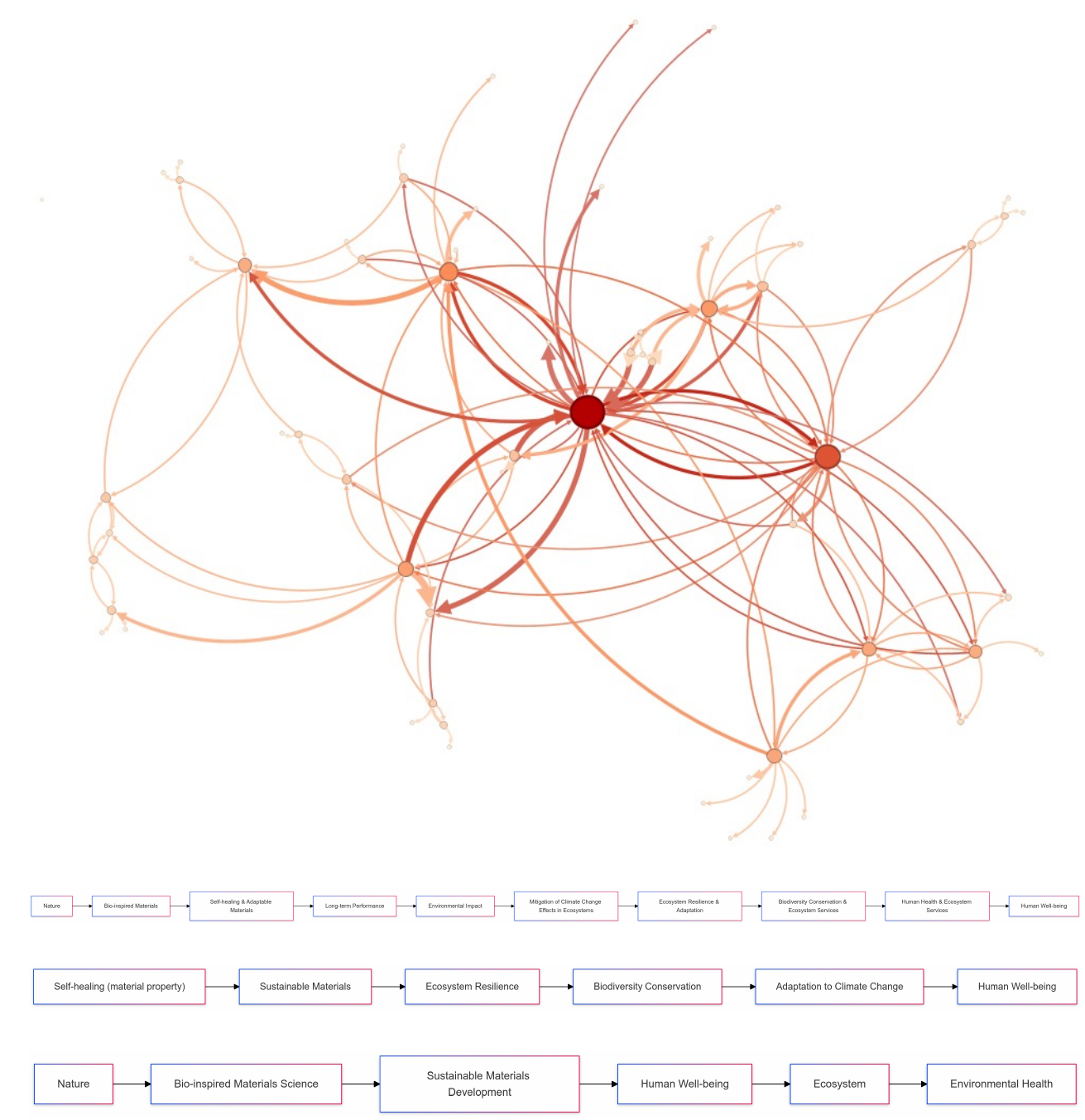}
        \caption{An autonomously grown knowledge graph, where a multi-agent AI system creates questions and answers, each with a new knowledge subgraph, which is then concatenated into an integrated graph and converted into GraphML format using \texttt{GPT-4o} for analysis using \texttt{networkx} and \texttt{Gephi}. The upper panel shows the graph, and the lower panel a selection of interesting paths identified in the overall graph. The starting prompt was chosen to be: \texttt{Discuss an interesting idea in bio-inspired materials science.} The model then iterated for $N=12$ iterations to grow the graph. The nodes with the highest degrees are `Sustainable Materials', `Self-healing and Adaptable Materials', and `Ecosystems'.}
	\label{fig:fig_1000}
\end{figure}

For a fully autonomous generation of graphs without human input for prompt generation, the system can be modified easily by using an LLM to develop new prompts autonomously based on previous results, to create an ever-expanding abstraction of relationships. Figure~\ref{fig:fig_1000} shows the results of an autonomously grown knowledge graph. The upper panel shows the graph, and the lower panel depicts a selection of interesting paths identified in the integrated graph. The starting prompt, the only one generated by a human, was chosen to be: 

\begin{LLMbox}{User task}
\begin{lstlisting}
(*@\bf\hlred{User:}@*) Discuss an interesting idea in bio-inspired materials science.
\end{lstlisting}
\end{LLMbox}

The model then iterated for $N=12$ iterations to grow the graph. Figure~\ref{fig:bio-inspired-flow} shows the series of prompts developed and used by the model to grow the graph (we emphasize that only the first prompt was provided by the human user).

\begin{figure}[h!]
\centering
\begin{tikzpicture}[
    node distance=0.5cm and 0.5cm,
    promptblue/.style={rectangle, draw=blue!50, fill=blue!10, thick, text width=12cm, align=center, rounded corners, inner sep=5pt},
    promptorange/.style={rectangle, draw=orange!50, fill=orange!10, thick, text width=12cm, align=center, rounded corners, inner sep=5pt},
    line/.style={draw, thick, ->, >=stealth, black!70},
    every node/.append style={font=\footnotesize}
]

\node[promptblue] (Q1) {Discuss an interesting idea in bio-inspired materials science.};
\node[promptorange, below=of Q1] (Q2) {How do the properties of **Bio-inspired Materials** derived from **Nature**, such as self-healing and adaptability, impact the development of **Sustainable Materials** for **Human Well-being** in **Ecosystems**?};
\node[promptorange, below=of Q2] (Q3) {What role do **Self-healing and Adaptability** play in the development of **Sustainable Materials** for **Human Well-being in Ecosystems**, and how do these properties impact the long-term performance and environmental impact of bio-inspired materials?};
\node[promptorange, below=of Q3] (Q4) {How do self-healing and adaptability properties in bio-inspired materials influence the development of sustainable materials for human well-being in ecosystems, particularly in relation to long-term performance and environmental impact?};
\node[promptorange, below=of Q4] (Q5) {What role do **Adaptability** and **Self-healing** play in the development of **Sustainable Materials** for **Human Well-being in Ecosystems**, and how do these properties influence the **Environmental Impact** of these materials?};
\node[promptorange, below=of Q5] (Q6) {What role do **Adaptability** and **Self-healing** play in the development of **Sustainable Materials** for **Human Well-being in Ecosystems**, and how do these properties impact the **Environmental Impact** of these materials?};
\node[promptorange, below=of Q6] (Q7) {How do the properties of self-healing and adaptability in bio-inspired materials influence their potential to mitigate the effects of climate change on ecosystems, and what role do sustainable materials play in this process?};
\node[promptorange, below=of Q7] (Q8) {How do **Self-healing and Adaptable Materials** contribute to the **Mitigation of Climate Change Effects** in **Ecosystems**, and what role do they play in **Sustainable Materials** for **Human Well-being in Ecosystems**?};
\node[promptorange, below=of Q8] (Q9) {How do self-healing and adaptable materials influence the development of sustainable materials for human well-being in ecosystems, particularly in relation to mitigating the effects of climate change on ecosystem resilience and biodiversity conservation?};
\node[promptorange, below=of Q9] (Q10) {What role do **Bio-inspired Materials** play in **Ecosystem Resilience**, and how do they contribute to the development of **Sustainable Materials** for mitigating the effects of **Climate Change**?};
\node[promptorange, below=of Q10] (Q11) {How do bio-inspired materials influence the development of sustainable materials for human well-being in ecosystems, particularly in the context of mitigating the effects of climate change?};
\node[promptorange, below=of Q11] (Q12) {What role do bio-inspired materials play in mitigating the effects of climate change on ecosystems, and how can their development contribute to sustainable materials for human well-being in ecosystems?};

\path[line] (Q1) edge (Q2);
\path[line] (Q2) edge (Q3);
\path[line] (Q3) edge (Q4);
\path[line] (Q4) edge (Q5);
\path[line] (Q5) edge (Q6);
\path[line] (Q6) edge (Q7);
\path[line] (Q7) edge (Q8);
\path[line] (Q8) edge (Q9);
\path[line] (Q9) edge (Q10);
\path[line] (Q10) edge (Q11);
\path[line] (Q11) edge (Q12);

\begin{scope}[on background layer]
    \node[draw=blue!50, rounded corners, inner sep=10pt, fit=(Q1)(Q12)] (bg) {};
\end{scope}

\end{tikzpicture}
\caption{Logical sequential flow of prompts exploring bio-inspired materials. The progression begins with a broad question (prescribed by the human: \texttt{Discuss an interesting idea in bio-inspired materials science.}) and transitions to increasingly detailed and interconnected inquiries into adaptability, self-healing, and their roles in mitigating climate change and enhancing ecosystem resilience. This structure exemplifies recursive reasoning. These autonomously generated prompts are used for constructing knowledge graphs that are then assembled into an integrated graph.}
\label{fig:bio-inspired-flow}
\end{figure}

\begin{figure}
	\centering
	\includegraphics[width=.8\linewidth, 
    ]{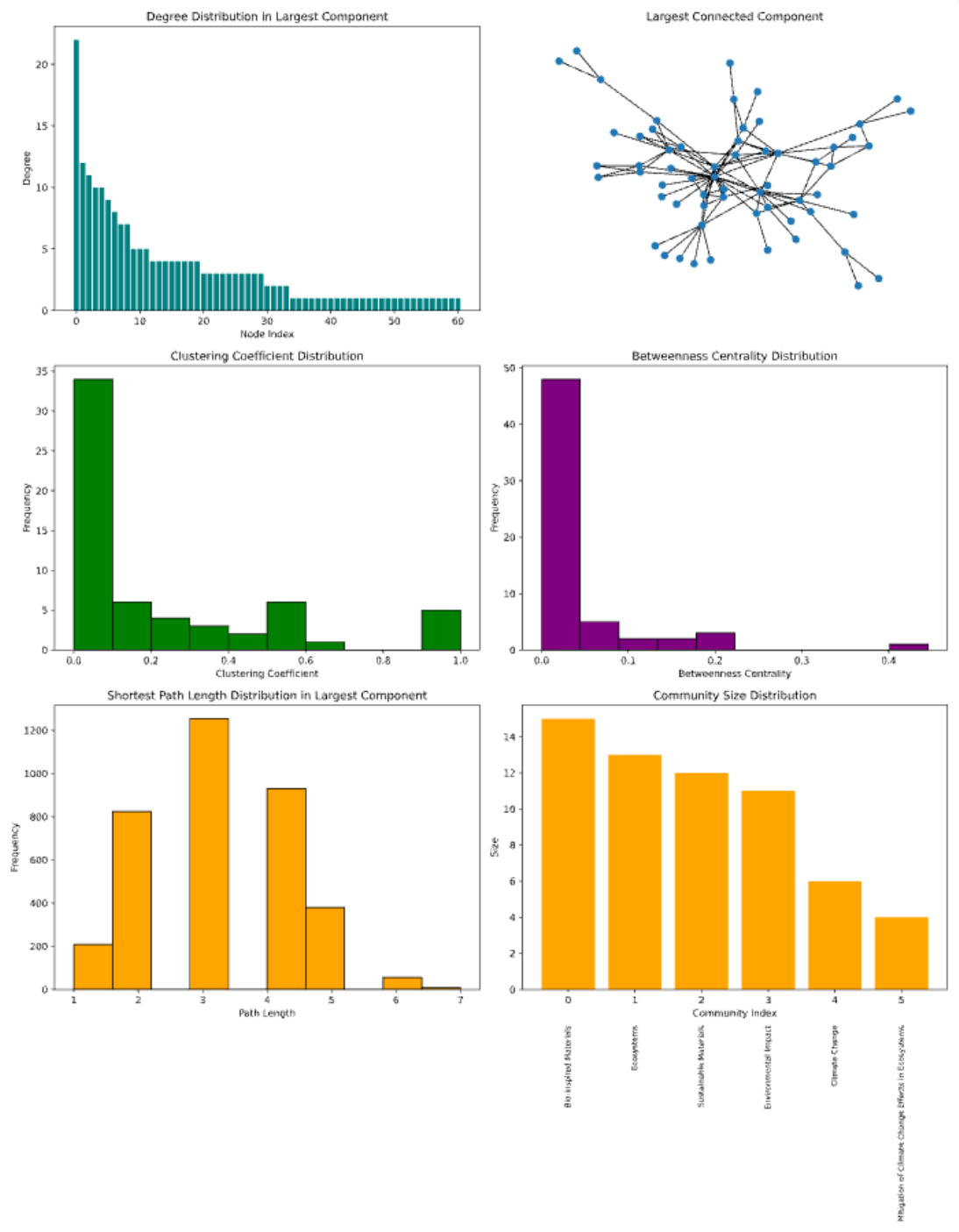}
        \caption{Visualization of the graph analysis of the autonomously grown graph depicted in Figure~\ref{fig:fig_1000}. 
(Top Left) Degree distribution shows a heterogeneous structure with prominent hubs and many low-degree nodes. 
(Top Right) Visualization of the largest connected component reveals its modular structure. 
(Middle Left) Clustering coefficient distribution highlights sparse local connectivity, with some nodes in dense clusters. 
(Middle Right) Betweenness centrality distribution indicates a small subset of critical nodes facilitating connectivity. 
(Bottom Left) Shortest path length distribution suggests a small-world structure with most nodes separated by a few hops. 
(Bottom Right) Community size distribution, with central nodes labeled, reflects varying sizes and roles within the graph.
The analysis collectively underscores the complex topology of the graph, including heterogeneity, modularity, and small-world characteristics.
}
	\label{fig:fig_1001}
\end{figure}

The analysis presented in Figure~\ref{fig:fig_1001} provides a comprehensive understanding of the graph's structure and dynamics. The degree distribution (Top Left) highlights the presence of a scale-free topology, characterized by a few highly connected hubs that dominate the graph's connectivity, alongside a majority of low-degree nodes. This structure suggests robustness to random node failures but potential vulnerability to targeted attacks on the hubs.  The visualization of the largest connected component (Top Right) reveals its modular nature, with densely connected subregions surrounded by sparsely connected nodes. This organization suggests a hierarchical topology, where clusters are linked by intermediary nodes, enabling efficient navigation through the network.
The clustering coefficient distribution (Middle Left) emphasizes the variation in local connectivity. While most nodes exhibit low clustering coefficients, indicating sparse local neighborhoods, the few nodes with high clustering coefficients are likely critical to maintaining the coherence of tightly-knit subcommunities.
The betweenness centrality distribution (Middle Right) further illustrates the graph's structural dependencies. A small subset of nodes exhibits significantly higher centrality, underscoring their role as key connectors that facilitate information flow between otherwise disjoint regions of the graph. These nodes act as bridges, ensuring overall network connectivity and efficiency.
The shortest path length distribution (Bottom Left) demonstrates that the graph exhibits small-world characteristics, with most nodes separated by only a few hops. This property enables rapid information transfer across the graph, a hallmark of efficient networks in both natural and engineered systems.
The community size distribution (Bottom Right) provides insights into the modular organization of the graph. Communities vary widely in size, with larger communities potentially playing a dominant role in global connectivity. Central nodes within each community, annotated at the base of the bars, represent the most connected nodes within their respective communities. These nodes likely serve as local hubs, facilitating intra-community interactions and linking smaller, peripheral nodes.
This analysis reveals a graph that is highly efficient, modular, and robust, with a clear interplay between global connectivity and local structure. The presence of scale-free properties, small-world behavior, and modular organization suggests that the grown graph's topology is optimized for both stability and adaptability.

In the next experiment, we prompt the model with this initial prompt, by specifically focusing the algorithm to develop new tasks at the intersection with distinct knowledge areas to foster exploration of dissimilar ideas, such as art and science (see, Materials and Methods, section~\label{garden_prompt}):

\begin{LLMbox}{User task}
\begin{lstlisting}
(*@\bf\hlred{User:}@*) Discuss protein design based on collagen and elastin.
\end{lstlisting}
\end{LLMbox}

\begin{figure}[h!]
\centering
\begin{tikzpicture}[
    node distance=.25cm and .25cm,
    promptblue/.style={rectangle, draw=blue!50, fill=blue!10, thick, text width=15cm, align=center, rounded corners, inner sep=5pt},
    promptorange/.style={rectangle, draw=orange!50, fill=orange!10, thick, text width=14.5cm, align=center, rounded corners, inner sep=5pt},
    line/.style={draw, thick, ->, >=stealth, black!70},
    every node/.append style={font=\footnotesize}
]

\node[promptblue] (Q1) {Discuss protein design based on collagen and elastin.};
\node[promptorange, below=of Q1] (Q2) {Can the properties of hydrogel materials, inspired by the hierarchical structures of collagenous proteins and elastin-like proteins, be used to create a new form of 'bioluminescent biomaterial' that can adapt to and respond to environmental stimuli, blurring the lines between biology, art, and technology?};
\node[promptorange, below=of Q2] (Q3) {Can a bioluminescent biomaterial, influenced by environmental stimuli, be used as a medium for an immersive, site-specific art installation that explores the relationship between living organisms and their surroundings, blurring the boundaries between nature and art?};
\node[promptorange, below=of Q3] (Q4) {How might the intersection of bioluminescent biomaterials, nature-inspired art, and the blurring of boundaries between art and nature influence the development of new, site-specific, and immersive installations that not only interact with the environment but also challenge our understanding of the boundaries between living organisms and inanimate objects?};
\node[promptorange, below=of Q4] (Q5) {Can a bioluminescent biomaterial be used to create an immersive, site-specific installation that blurs the boundaries between art and nature, challenging our understanding of the relationship between living organisms, materials, and the environment, and ultimately, forcing us to reevaluate the notion of 'art' as a static, human-created entity?};
\node[promptorange, below=of Q5] (Q6) {Can a bioluminescent biomaterial be used to create an immersive, site-specific installation that challenges the traditional understanding of art as a static entity, and in doing so, blurs the boundaries between art and nature, ultimately influencing our perception of the relationship between living organisms, materials, and the environment?};
\node[promptorange, below=of Q6] (Q7) {How can the concept of 'Bioluminescent Biomaterial' be used to challenge traditional notions of 'Art' and 'Nature' in a way that blurs the boundaries between the two, and ultimately, what implications does this have on our understanding of the relationship between living organisms, materials, and the environment?};
\node[promptorange, below=of Q7] (Q8) {How might the incorporation of bioluminescent biomaterials into site-specific, immersive art installations challenge our understanding of the relationship between living organisms, materials, and the environment, ultimately blurring the boundaries between art, nature, and the human experience, and prompting a reevaluation of the role of art in shaping our perception of the natural world?};
\node[promptorange, below=of Q8] (Q9) {How might the incorporation of bioluminescent biomaterials into site-specific, immersive art installations challenge our understanding of the boundaries between art, nature, and the human experience, ultimately blurring the lines between the organic and the synthetic, and forcing us to reevaluate the role of art in the natural world?};
\node[promptorange, below=of Q9] (Q10) {How can the concept of 'bioluminescent biomaterials' as a medium for site-specific, immersive art installations challenge our understanding of the boundaries between art, nature, and human experience, and ultimately redefine the role of art in the natural world?};
\node[promptorange, below=of Q10] (Q11) {Can the boundaries between art, nature, and human experience be further blurred by incorporating bioluminescent biomaterials into a 'mythological' art installation that reimagines the relationship between the human body and the natural world, drawing inspiration from ancient mythologies and the concept of 'thin places' where the boundaries between reality and the divine are said to be at their thinnest?};
\node[promptorange, below=of Q11] (Q12) {How can the concept of ``thin places'' in mythology, where the boundaries between the natural and human-made worlds are blurred, be applied to the design of immersive, site-specific art installations that incorporate bioluminescent biomaterials, to create a new form of experiential art that challenges our understanding of the relationship between nature, art, and human experience?};

\path[line] (Q1) edge (Q2);
\path[line] (Q2) edge (Q3);
\path[line] (Q3) edge (Q4);
\path[line] (Q4) edge (Q5);
\path[line] (Q5) edge (Q6);
\path[line] (Q6) edge (Q7);
\path[line] (Q7) edge (Q8);
\path[line] (Q8) edge (Q9);
\path[line] (Q9) edge (Q10);
\path[line] (Q10) edge (Q11);
\path[line] (Q11) edge (Q12);

\begin{scope}[on background layer]
    \node[draw=blue!50, rounded corners, inner sep=10pt, fit=(Q1)(Q12)] (bg) {};
\end{scope}
 
\end{tikzpicture}
\caption{Logical sequential flow of prompts exploring protein design and bioluminescent biomaterials. The progression begins with protein design and expands into speculative questions about bioluminescent biomaterials, their artistic and ecological applications, and their integration into immersive art installations. These prompts illustrate how the intersection of biology, technology, and art can be explored recursively to generate novel insights and creative concepts.}
\label{fig:bioluminescent-flow}
\end{figure}
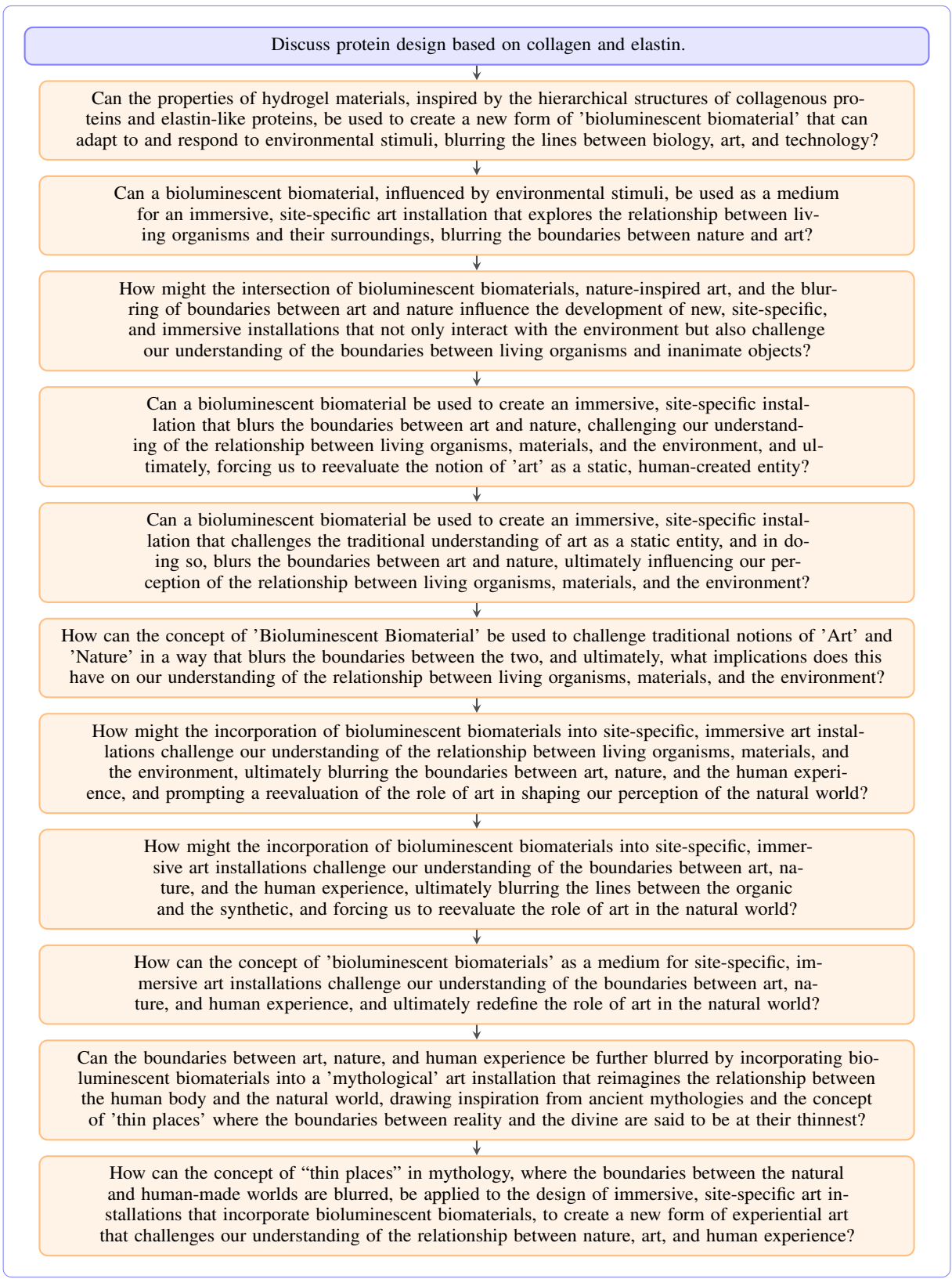

\begin{figure}
	\centering
	\includegraphics[width=.95\linewidth, 
    ]{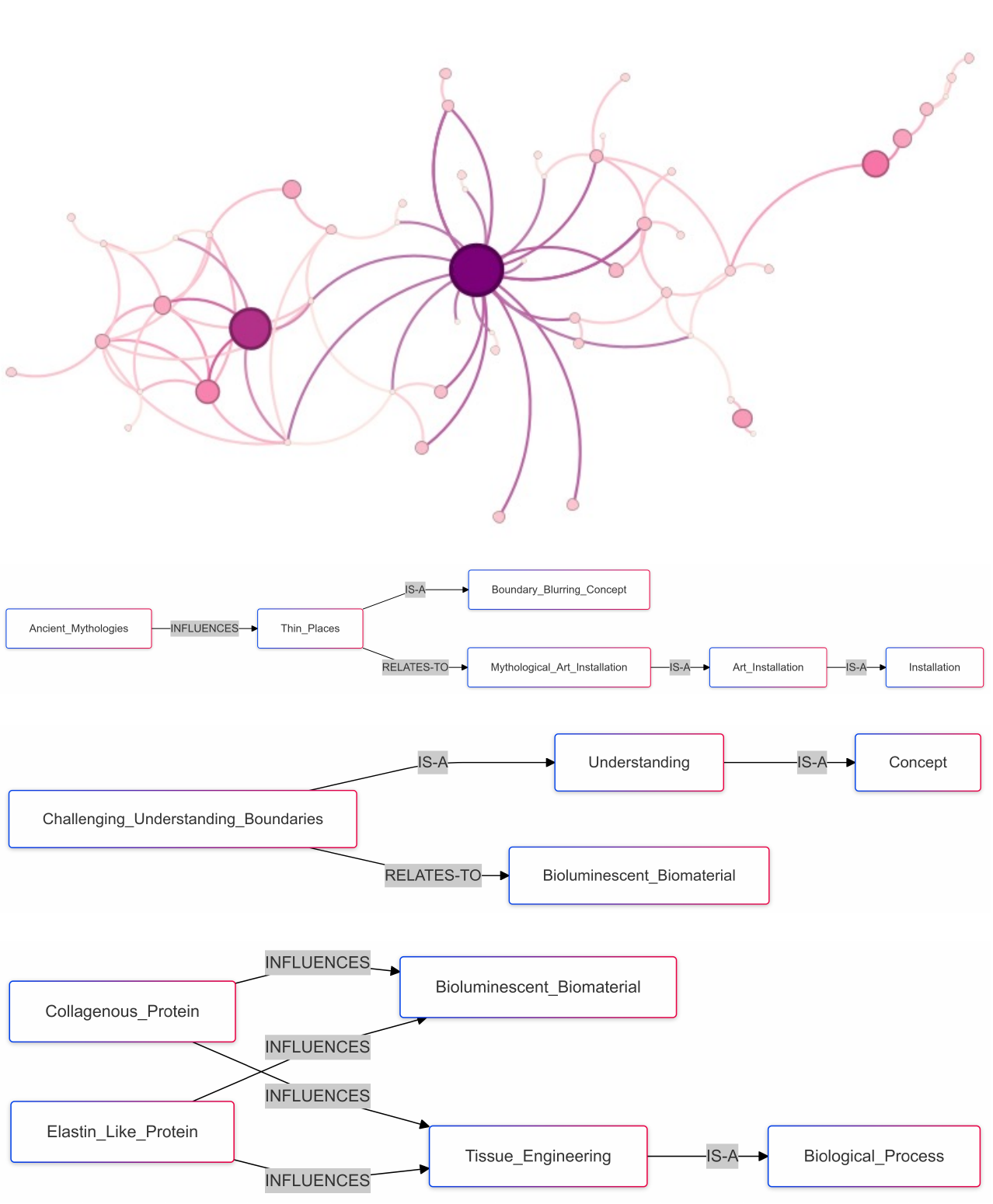}
        \caption{Grown knowledge graph based on the prompt \texttt{Discuss protein design based on collagen and elastin.} Here, the agentic system is specifically charged to develop new questions (see, Figure~\ref{fig:bioluminescent-flow}) that integrate dissimilar fields such as philosophy or art. The visualization depicts nodes sized and colored by page rank, with top nodes being `Biolumniscent Biomaterial' (most significant node also per node degree), `Protein', `Installation', `Elastin' and `Concept'. A unique feature is the incorporation of the concept of `thin places', a concept drawn from various mythological and spiritual traditions that invokes sites or moments where the boundary between the physical world and a transcendent realm is perceived to be exceptionally thin or permeable. The model identifies this as an interesting association for this particular task, which is particularly interesting as a way to create a new concept that incorporates highly multidisciplinary relationships. 
}
	\label{fig:fig_2000}
\end{figure}

The algorithm develops a series of prompts as shown in Figure~\ref{fig:bioluminescent-flow}, resulting in the integrated graph depicted in Figure~\ref{fig:fig_2000}.  A deeper analysis of the resulting graph is provided in Figure~\ref{fig:fig_1001_other_exp}.

We highlight one unique insight developed by the model, to relate the concept of `thin places' with `biological materials' and `art'. Thin places is a concept drawn from various mythological and spiritual traditions, notably Celtic lore, referring to sites or moments where the boundary between the physical world and a transcendent realm is perceived to be exceptionally thin or permeable. In our integrated graph, these \emph{thin places} are not only connected to site-specific, mythological art installations but also echo the conceptual blurring of boundaries integral to protein design and biomaterial engineering. Such a connection is unusual because, at first glance, mythic realms and advanced biotechnology inhabit entirely distinct spheres. Yet the same drive to challenge and dissolve traditional boundaries--whether between art and nature, or between synthetic and organic matter--underlies both phenomena. Consequently, the metaphor of thin places illuminates how immersive, boundary-blurring experiences in mythological art might inform innovative approaches to tissue engineering and bioluminescent biomaterials, bridging realms often seen as incompatible. This underscores the novel insight that creative, myth-inspired concepts can resonate meaningfully with the cutting edges of scientific inquiry.

Naturally, the output of these reasoning steps and larger graph generated can be used for further analysis. While a broader expansion is beyond the scope of this paper, we show one example here. For instance, we experimented with charging the \texttt{o1} model~\cite{o1-model-card-2024} to propose a new theory or concept that incorporates the results produced by Graph-PReFLexOR. The results are shown in Text Box~\ref{TexBox:o1_liminality}, showing how other reasoning models can utilize the graph reasoning output and produce concise delineation of ideas.

\section{Conclusion}

This work introduced Graph-PReFLexOR, a framework that integrates \textit{in-situ} graph reasoning, symbolic abstraction, and recursive reflection into the generative modeling paradigm (Figures~\ref{fig:fig_01} and \ref{fig:fig_5}). By embedding graph-based intermediate representations within LLMs, Graph-PReFLexOR advances beyond the limitations of purely linguistic systems to tackle complex tasks requiring relational reasoning, multi-step deduction, and adaptive knowledge synthesis.  Our experiments were designed to assess particularly how well the model could generalize, as the training data consisted of ~1,000 papers on biological materials~\cite{Luu2023BioinspiredLLM:Materials}, whereas the tasks were constructed at the interface of science and other disciplines including design, music, and philosophy to probe the model's generalization capability beyond the technical-focused training data. We find exceptional performance throughout and see that the model could very well generalize, follow the structured reasoning it learned during training (see, Figure~\ref{fig:fig_01}), and even construct highly complex graphs that interface a myriad of disciplines (see, e.g. Figure~\ref{fig:fig_11} and ~\ref{fig:fig_12}), and other results. 

A particularly compelling application was the development of the knowledge garden concept, building on the ability of the model to grow knowledge graphs dynamically and iteratively, by adding new relational insights and abstractions. This resulted in a series of experiments where we expanded graphs upon an initial simple task, yielding complex graph structures that themselves could be the basis of further research, inquiry and reasoning (e.g., Figures~\ref{fig:Fig_12_abstract} and \ref{fig:Fig_12_variant}). Much future work can be conducted based on this method. Some initial experimentation on growing graphs autonomously yielded interesting results, such as shown in Figures~\ref{fig:fig_1000} and \ref{fig:fig_1001}, where we identified a capability of the model to search and expand topics and connect complex ideas. 

By explicitly constructing and abstracting relational graphs, the Graph-PReFLexOR approach provides a more structured foundation than standard sequence‐oriented Transformer training. In particular, preserving connectivity among entities and relations makes it easier to detect and exploit universal features such as isomorphisms and recurrent subgraph patterns. Unlike pure next‐token objectives, which largely rely on distributional cues hidden in token sequences, graph abstractions bring structural commonalities to the forefront, enabling more systematic identification of shared algebraic forms, relational templates, and higher‐order symmetries. As a result, symbolic rewriting and generalization become more direct, since the underlying topology is explicitly represented rather than merely inferred. Consequently, the resulting embeddings capture deeper, domain‐invariant regularities that might otherwise remain implicit or fragmentary in a purely sequence‐based approach.

\subsection{Mathematical and Logical Framework}

Graph-PReFLexOR establishes a pipeline of thinking and a unified framework that bridges symbolic reasoning and dynamic graph-based abstraction to tackle the complexity of scientific inquiry. At its core, the framework formalizes reasoning as a multi-layered mapping as introduced in equation~\ref{eq:M_T_G_P} where a task \( \mathcal{T} \) produces a knowledge graph \( \mathcal{G} \), abstract patterns \( \mathcal{P} \), and final answers \( \mathcal{A} \). The knowledge graph \( \mathcal{G} = (V, E) \) encodes concepts as nodes \( V \) and relationships as directed edges \( E \), such as \texttt{IS-A}, \texttt{RELATES-TO}, and \texttt{INFLUENCES}. The system derives abstract patterns \( \mathcal{P} \) by identifying higher-order dependencies, structured as transformations:
\[
\alpha \to \beta \to \gamma \to \delta \to \epsilon,
\]
with proportional relationships such as:
\[
\alpha \propto \epsilon,
\]
indicating how the initial state \( \alpha \) contributes to the final transformation \( \epsilon \). Recursive reflection as shown in Figure~\ref{fig:PreFlexOR_recursive} refines these outputs iteratively, producing an optimized answer \( \mathcal{A} \).

Inspired by category theory, Graph-PReFLexOR emphasizes relational over intrinsic properties, allowing the abstraction of domain-specific patterns into reusable, transferable frameworks. For example, graph symmetries enable translational reasoning across materials science, bioengineering, and philosophy, breaking traditional siloed boundaries. The recursive refinement process ensures these abstractions remain grounded and interpretable, aligning outputs with task-specific goals while introducing new hierarchies of understanding, specifically creating shared embedding representations with multidimensional mappings.

Another key aspect of the framework lies in its iterative, knowledge-expanding capability. By incorporating feedback-driven graph updates and dynamically integrating symbolic patterns, Graph-PReFLexOR demonstrates the potential for \textit{in-situ} knowledge growth. This positions it uniquely to address challenges in hypothesis generation, interdisciplinary exploration, and adaptive learning—tasks that require not just computational power but conceptual flexibility. Ultimately, Graph-PReFLexOR presents opportunities to expand the paradigm of AI-driven reasoning by fostering transparent, interpretable models of discovery, laying the groundwork for a new era of autonomous scientific and creative inquiry.

\subsection{Advances Over Existing Approaches}

Compared to traditional LLMs that rely on unstructured token-level generation, Graph-PReFLexOR introduces:
\begin{itemize}
    \item \textbf{Structured Intermediate Representations:} Unlike models that output responses directly, Graph-PReFLexOR generates a knowledge graph \( \mathcal{G} \) to explicitly represent relationships and dependencies. This improves interpretability and ensures reasoning consistency, which can be adapted easily to other scientific, technical or other domains, and formalizes shared representations.
    \item \textbf{Symbolic and Connectionist Integration:} While most transformer-based models do not focus specifically on symbolic reasoning capabilities, Graph-PReFLexOR bridges the gap by combining linguistic fluency with graph-based reasoning. This aligns with hybrid approaches but demonstrates superior adaptability to novel tasks.
    \item \textbf{Recursive Refinement:} Recursive reasoning enables multi-step improvement by iteratively refining knowledge graphs and abstract patterns. This dynamic process enhances response quality and ensures alignment with task objectives. 
    \item \textbf{Scalability and Adaptability:} Graph-PReFLexOR extends traditional methods by dynamically expanding its knowledge graph to adapt to evolving prompts, facilitating knowledge transfer across domains.
\end{itemize}

Our experimental results demonstrate that Graph-PReFLexOR outperforms baseline models in reasoning depth, knowledge transfer, and adaptability. For instance, the recursive reasoning mechanism (equation~\ref{eq:recursive}) achieves higher reasoning depth and adaptability compared to static methods, as we demonstrated in Table~\ref{tab:response_comparison_with_graph}. For scientific applications, and special cases like mathematics, the critic function $f_\text{critic}$ could  involve specific fact-checking or consistency assessments, or even executing simulations to inject new data or physical insights. These can be incorporated during training or inference and ensure that such a recursive reasoning system produces accurate and consistent results. This flexibility offers significant potential for scientific applications.

\subsection{Quantum-Inspired Metaphor for Transformers and Graph-Based Reasoning}

\begin{figure}[h!]
\centering
\sffamily 
\scriptsize
\sansmath 

\begin{tikzpicture}[
    >=stealth,
    node distance=3cm,
    every node/.style={font=\small},
    scale=1.0,
    transform shape
]

\tikzstyle{block} = [draw, rectangle, minimum width=2.cm, minimum height=1.25cm, align=center]
\tikzstyle{cloud} = [draw, ellipse, 
                     minimum width=2.5cm, minimum height=1.5cm, align=center]

\node[cloud, fill=gray!10] (superposition) {
    \begin{tabular}{c}
    Latent Superposition \\
    (\emph{All Potential Answers})
    \end{tabular}
};

\node[above of=superposition, yshift=-.5cm, font=\normalsize] (transformer) {Transformer Model: Potentiality};

\node[block, fill=yellow!20, below of=superposition] (measurement) {
    \begin{tabular}{c}
    Measurement \\
    (\emph{Task Constraint})
    \end{tabular}
};

\node[block, fill=blue!10, right of=measurement, node distance=5cm] (graph) {
    \begin{tabular}{c}
    Graph-Based \\  
    Reflection \& Reasoning
    \end{tabular}
};

\node[block, fill=green!15, right of=graph, node distance=5cm] (answer) {
    \begin{tabular}{c}
    Optimal \\
    Answer
    \end{tabular}
};

\draw[-, thick, dashed] (transformer) -- (superposition)
    node[midway, right, xshift=0.2cm, align=center] {\scriptsize};
\draw[->, thick] (superposition) -- (measurement)
    node[midway, left, xshift=-0.1cm, align=center] {\scriptsize \emph{Task}\\[-4pt]\scriptsize \emph{Query}};
\draw[->, thick] (measurement) -- (graph)
    node[midway, above, align=center] {};
\draw[->, thick] (graph) -- (answer)
    node[midway, above, align=center] {\scriptsize \emph{Refined}\\[-4pt]\scriptsize \emph{Solution}};

\begin{scope}[shift={(4.5,-1.5)}] 
    \node[circle, fill=white, draw, inner sep=1pt] (gn0) at (-.6,-.3) {};
    \node[circle, fill=white, draw, inner sep=1pt] (gn1) at (0,0) {};
    \node[circle, fill=white, draw, inner sep=1pt] (gn2) at (0.4,0.5) {};
    \node[circle, fill=white, draw, inner sep=1pt] (gn3) at (0.8,-0.35) {};
    \node[circle, fill=white, draw, inner sep=1pt] (gn4) at (1.2,1.) {};
    \draw (gn1) -- (gn2);
    \draw (gn2) -- (gn3);
    \draw (gn1) -- (gn3);
    \draw (gn1) -- (gn2);
    \draw (gn4) -- (gn2);
    \draw (gn0) -- (gn1);
    
\end{scope}

\end{tikzpicture}
\caption{A conceptual illustration of how a Transformer-based system starts with a latent superposition of potentiality that encompass all potential answers, in a lose analogy to quantum systems. When a task is posed, it ``measures'' these possibilities to produce an output. In Graph-PReFLexOR, we leverage graph-based reflection and reasoning to converge on an optimal answer by imposing structured mechanisms of reasoning.}
\label{fig:conceptual_illustration_measurement}
\end{figure}
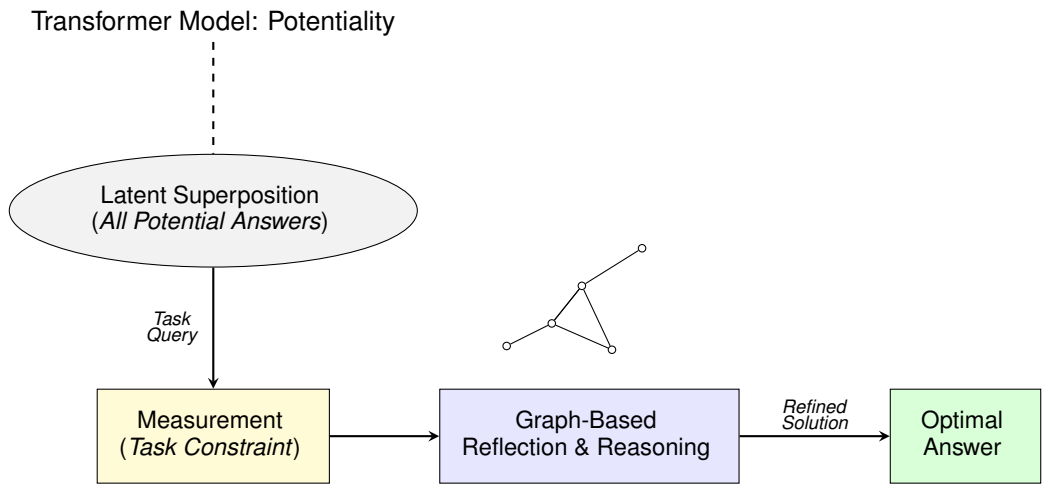

The Transformer architecture, at its core, mirrors certain fundamental principles of quantum mechanics, particularly in its representation and refinement of knowledge. While the analogy is not perfect and it is crucial to recognize that this remains a metaphor and not an actual quantum mechanical process, it allows us to explore some new associations. In a Transformer, before a specific task or question is posed, the system exists in a state analogous to quantum superposition, where all possible answers coexist as potential outcomes (Figure~\ref{fig:conceptual_illustration_measurement}). Formally, let the system’s latent state be represented as a superposition of knowledge embeddings:
\[
|\Psi_0\rangle = \sum_k c_k |k\rangle,
\]
where each basis state \(|k\rangle\) corresponds to a potential reasoning path or answer, and \(c_k\) reflects the weight or amplitude of that path based on prior knowledge. The multi-head attention mechanism introduces a form of entanglement, binding token representations through a context-sensitive tensor product:
\[
|\Psi\rangle = |\psi_1\rangle \otimes |\psi_2\rangle \otimes \cdots \otimes |\psi_N\rangle,
\]
where tokens dynamically interact to refine the system’s focus. This process can be seen as reducing the system’s uncertainty, akin to minimizing entropy in quantum systems.

In the case of Graph-PReFLexOR, graph-based reasoning emerges as one expression of this general quantum-like behavior. Here, the knowledge graph serves as a concrete structure for representing interdependencies between concepts, evolving iteratively through recursive refinement as shown in Figure~\ref{fig:PreFlexOR_recursive}. Each refinement step is analogous to the application of a quantum operator, governed by reasoning rules encoded in a Hamiltonian:
\[
|\psi_{i+1}\rangle = \hat{H} |\psi_i\rangle,
\]
where \(|\psi_i\rangle\) represents the graph’s state at step \(i\), and \(\hat{H}\) encodes the principles of relational optimization (e.g., maximizing coherence or minimizing contradictions). The final knowledge graph, after sufficient refinement, reflects a collapsed eigenstate, optimized for the given question.

Principally, this analogy extends to the entire Transformer framework. The recursive refinement within Graph-PReFLexOR (e.g., Figures~\ref{fig:fig_01}, \ref{fig:fig_5} and \ref{fig:PreFlexOR_recursive}) exemplifies how transformers resolve complex queries by dynamically collapsing potential answers into structured outputs through an iterative process. The abstraction mechanisms used here are just one specific instantiation of this broader capability, where relationships between nodes represent entangled dependencies that become more expressive, and accessible, through shared representations that encourage the model think in isomorphisms. In more abstract terms, the system applies interference patterns via attention weights to amplify relevant paths and suppress contradictory ones:
\[
c_k^{(i+1)} = \sum_j \mathcal{A}_{kj} c_j^{(i)},
\]
where \(\mathcal{A}_{kj}\) are attention weights that act analogously to quantum amplitudes, modulating the influence of each path.

By connecting graph reasoning to these foundational quantum-like behaviors, Graph-PReFLexOR illustrates how Transformers can model tasks requiring multi-step abstraction, recursive refinement, and contextual reasoning. The graph serves as a symbolic representation of the underlying quantum-inspired mechanisms, offering interpretability and structure while adhering to the general principles governing the model’s operation.

The reasoning process can be viewed as evolving from a state of superposition \( |\Psi_0\rangle = \sum_k c_k |k\rangle \), where potential reasoning paths are explored. Through recursive refinement, the framework updates the reasoning state iteratively, $\mathbf{R}_{i+1} = f_\text{critic}(\mathbf{R}_i, \mathbf{F}_i)$,  
analogous to quantum state evolution governed by external feedback \( \mathbf{F}_i \). The integration of reasoning paths into a unified representation aligns with the mapping  defined in equation~\ref{eq:M_T_G_P}, where a task \( \mathcal{T} \) produces a knowledge graph \( \mathcal{G} \), abstract patterns \( \mathcal{P} \), and final answers \( \mathcal{A} \). Just as the quantum system collapses into a measurable eigenstate, Graph-PReFLexOR resolves these intermediate abstractions into a final response: $\mathbf{A} = g(\mathbf{R}_N)$, providing a coherent and interpretable output.  

Transformers are unique in developing this analogy, as unlike traditional neural networks, Transformers explicitly model relationships between inputs through their self-attention mechanism, creating a dynamic superposition of weighted dependencies across all tokens in a sequence. This mechanism reflects quantum superposition, where all possible relationships coexist as potentialities before being resolved. Multi-headed attention further enhances this analogy by enabling independent exploration of diverse patterns, akin to quantum entanglement, where multiple states are interdependent. Additionally, the Transformer’s decoding process parallels wavefunction collapse, as the model resolves probabilistic attention distributions into specific outputs. This explicit modeling of interdependencies and dynamic evolution of representations makes Transformers inherently suited to the quantum analogy, more so than other architectures like convolutional or recurrent neural networks, which rely on static filters or sequential processing, respectively. Dense neural networks rely on fixed weight matrices to transform inputs in a static, layer-by-layer manner, without explicitly capturing interdependencies between different elements of the input. Each transformation is local to the input vector as a whole, akin to a single deterministic operation, rather than a superposition of potential relationships. Unlike transformers, dense networks do not employ mechanisms like self-attention to dynamically assign weights to input relationships, making it difficult to draw analogies to quantum phenomena such as superposition or entanglement. Furthermore, dense networks lack the iterative refinement of representations seen in transformers, where relational dependencies evolve layer by layer, mirroring quantum state evolution.  

While the quantum analogy offers conceptual insights, it does not literally apply to Transformers, which operate on real-valued vectors rather than complex Hilbert spaces and do not follow quantum mechanical laws such as unitarity or the Born rule. Consequently, terms like ``superposition,'' ``entanglement,'' and ``collapse'' are used figuratively, reflecting iterative reasoning rather than actual quantum processes. This can be misleading if taken literally, as Transformers remain classical computational systems. Still, these discussions give us insights into how Transformers can be viewed, and potentially improved, to better model physical processes. 

\subsection{Challenges and Future Opportunities}

Future work could focus on scaling the framework to larger datasets and more complex models, addressing challenges such as interpretability in deeply interconnected graphs. Integrating Graph-PReFLexOR with state-of-the-art architectures, such as multi-modal transformers and graph neural networks, will further expand its applicability to domains like bioinformatics, materials science, and hypothesis-driven discovery. Graph-PReFLexOR represents an advancement in reasoning frameworks, combining symbolic and connectionist paradigms. By embedding explicit graph reasoning and recursive reflection into LLMs, it sets a new benchmark for scientific discovery, unlocking opportunities for transformative research across disciplines.

The ability to construct structured and symbolically focused reasoning strategies within the flexible framework of LLMs is appealing as an alternative to conventional symbolic mechanisms and strategies. It also allows for the construction of powerful agentic frameworks. For instance, the approach taken in this paper demonstrated a strategy to create an iterative prompting to expand knowledge graphs; this can be done via human-AI collaboration or via AI-based reasoning only. It can also incorporate specific targets, such as to expand the knowledge graph towards specific directions (e.g. introduce art, music, specific technologies, etc.) to drive discovery and mimic some mechanisms seen in directed diffusion. Other strategies may invoke more complex approaches to avoid existing known relationships to drive the model towards unexplored domains. Other aspects could involve adding retrieval-augmented assessments against literature (e.g. using Semantic Scholar) or the internet, and even include assessments of feasibility as done in other agentic frameworks that operate at the interface of AI and physics~\cite{ghafarollahi2024sciagentsautomatingscientificdiscovery,Ghafarollahi2024,ghafarollahi2024atomagentsalloydesigndiscovery}.
This helps us to expand the use of AI tools as powerful assistants to connect ideas and explore new frontiers of knowledge that were previously not accessible. Broad access to such tools is essential, especially democratization access of AI for broad audiences, as this holds a key to vast unchartered knowledge. AI is no longer merely a tool but a partner in new creative pursuits of humans.

\section{Materials and Methods}

We describe key materials and methods developed in this section. 

\subsection{Graph-PReFLexOR model development}

\subsubsection{Knowledge Graph Generation and Question-Answering Framework}
This section details the key algorithms developed for generating a knowledge graph, enhancing context, and producing structured question-answering outputs during the training phase. The training dataset is the raw text of around 1,000 scientific papers, as detailed in earlier papers~\cite{Luu2023BioinspiredLLM:Materials,lu2024finetuninglargelanguagemodels}. 

During \textit{in-situ} dataset generation during training,~\cite{buehler2024preflexorpreferencebasedrecursivelanguage} the knowledge graph generation algorithm creates a focused graph that includes key concepts and their relationships, such as classification (IS-A), influence (INFLUENCES), and connections (RELATES-TO). Retrieval-Augmented Generation (RAG) is used to identify abstract patterns and dependencies, enhancing the conceptual representation, using the original dataset of raw scientific papers. As discussed in the original paper, the use of RAG is significant since it provides a direct, deep and structured connection to related concepts in the training corpus to ensure full and complete reasoning paths are developed during training.

The process proceeds in a systematic fashion via distinct phases of construction. 

\paragraph{Question-and-Answer Generation Algorithm}
This algorithm synthesizes knowledge graph generation, enriched context, and question-answer generation into a unified process. It produces challenging questions, detailed correct answers, and rejected answers for evaluation.

An abstract representation of the workflow is as follows:

1. Retrieve a randomly selected context $T$ (from all raw data) as a sequence of tokens $T = \{t_1, t_2, \dots, t_N\}$ from the knowledge index.
\[
T = \bigcup_{i=1}^k \text{Text}(n_i).
\]

2. Generate a question $Q$ based on $T$ by maximizing relevance:
\[
Q = \operatorname*{argmax}_q \text{Relevance}(q, T).
\]

3. Enrich $T$ using RAG, yielding $T' = T + \text{RAG}(T)$. 
Enrichment algorithms use RAG to add supplementary insights to the retrieved context, to connect specific text chunks used for generating the question with the entire dataset, ensuring that the answer and reasoning steps incorporate a global perspective.

4. Construct a knowledge graph $G$ from $T'$.

5. Extract reasoning steps $S$ and generate an answer $A$ as:
\[
A = \operatorname*{argmax}_a \text{Quality}(a | Q, S, G).
\]

6. Generate a rejected answer $A'$ by perturbing $A$ or through direct model prompts.

\textbf{Details on Graph Construction:}
Given a set of extracted concepts $C = \{c_1, c_2, \dots, c_n\}$ and relationships $R = \{r_1, r_2, \dots, r_m\}$, the knowledge graph $G = (V, E)$ is constructed as:
\begin{itemize}
    \item $V$: Nodes representing the concepts $C$.
    \item $E$: Directed edges $(c_i, c_j, r_k)$ where $r_k$ specifies the relationship type between $c_i$ and $c_j$.
\end{itemize}

The relationship types $r_k$ are encouraged to be within a set of categories (e.g., IS-A, RELATES-TO, INFLUENCES), albeit the training process may yield alternative relationship in the data. 

\paragraph{Abstract Pattern Generation Using the Generative Framework}

Abstract patterns are derived using RAG by identifying higher-order dependencies and summarizing them as:
\[
P(C, R) \sim \sum_{i,j} f(c_i, c_j; r_k),
\]
where $f$ quantifies the semantic strength of the relationship $r_k$ between $c_i$ and $c_j$.
Abstract patterns are constructed by analyzing hierarchical dependencies within the enriched context and summarizing key relationships into symbolic representations. The process begins with a knowledge graph \( G = (V, E) \), where \( V \) represents concepts and \( E \) defines relationships. The abstract pattern generation algorithm operates as follows:

First, we identify core concepts \( \{c_1, c_2, \dots, c_n\} \) from \( V \). Second, we evaluate the relevance of the relationships \( r_k \in E \) based on semantic weights:
\[
r_k = \arg \max_{r} f(c_i, c_j; r),
\]
where \( f \) is a scoring function quantifying the significance of \( r \) between concepts \( c_i \) and \( c_j \).
Third, we construct a directed sequence of abstract states \( \alpha, \beta, \gamma, \delta, \epsilon \) representing transformations, where:
\[
\alpha \to \beta \to \gamma \to \delta \to \epsilon.
\]
Finally we derive proportional relationships between states, such as:
\[
\alpha \propto \epsilon,
\]
indicating the dependence of the initial state \( \alpha \) on the final transformation \( \epsilon \).

This framework synthesizes complex relationships into interpretable symbolic patterns, enabling high-level reasoning and abstraction. The above steps are conducted via repeated prompting an LLM, with RAG, to construct data in the form delineated based on raw data.

\subsubsection{Integrated Dataset Generation Algorithm}
Using the above steps, this algorithm creates datasets for training or evaluation by generating question-answer pairs and applying optional filtering based on specific criteria.

\textbf{Abstract Workflow:}

1. Iterate over topics $T_k$ or randomly sample contexts.

2. For each $T_k$:
   - Generate a question $Q_k$, correct answer $A_k$, and rejected answer $A_k'$.
   - Optionally filter pairs where $A_k$ does not meet predefined correctness criteria.

3. Formulate the dataset as:
\[
\mathcal{D} = \{(Q_k, A_k, A_k') \mid k = 1, \dots, N\}.
\]

4. Apply post-processing $\mathcal{P}$ if specified, yielding the final dataset:
\[
\mathcal{D}' = \mathcal{P}(\mathcal{D}).
\]

The dataset generation algorithm supports training and evaluation by producing structured data that combines knowledge graph reasoning with contrastive answer evaluation.

\subsubsection{Example of Resulting \textit{in-situ} generated Data for Training}

The \textit{in-situ}  generated data consists of structured text representing user queries, reasoning steps, knowledge graphs, and responses. Below is a snippet of the resulting data.

First, the question generated is:

\begin{LLMbox}{Training data: Question}
\begin{lstlisting}
What is the primary advantage of using Origami and Kirigami Engineering (OKN) for fabricating microscale 3D objects compared to traditional top-down approaches?
\end{lstlisting}
\end{LLMbox}

The answer is developed first within a thinking section, followed by the actual answer. The thinking sectionb begins with the generation of a knowledge graph and reasoning, like so: 

\begin{LLMbox}{Training data: Knowledge graph/reasoning}
\begin{lstlisting}
**Knowledge Graph**:
Core Concepts and Relationships:
- OKN -IS-A- Origami Engineering, Kirigami Engineering
- OKN -RELATES-TO- Microscale 3D Objects (Enables fabrication)
- OKN -INFLUENCES- Traditional Top-Down Approaches (Overcomes limitations)
- OKN -RELATES-TO- Dynamic Transitions (From 2D patterns to 3D constructs)

**Reasoning Steps**:
1. OKNs enable the creation of 3D structures from 2D precursors.
2. They allow for reconfigurable structures surpassing traditional method constraints.
3. OKNs facilitate dynamic transitions and geometry changes.

**Additional Background**:
Origami and Kirigami Engineering enable dynamic geometry changes with unique properties, overcoming the limitations of traditional top-down approaches.
\end{lstlisting}
\end{LLMbox}

The abstract pattern identified is:

\begin{LLMbox}{Training data example: Abstract pattern}
\begin{lstlisting}
**Abstract Pattern:** 

(*@$\alpha \rightarrow \beta \rightarrow \gamma \rightarrow \delta \rightarrow \varepsilon$@*) 

**Key Transformation Rule:** 

(*@$\alpha \rightarrow \beta $@*): OKN enables fabrication of microscale 3D objects ((*@$\beta $@*)) 
(*@$\beta \rightarrow \gamma $@*): Reconfigurability influences OKN ((*@$\gamma $@*)) 
(*@$\gamma \rightarrow \delta $@*): OKN influences traditional top-down approaches ((*@$\delta $@*)) 
(*@$\delta \rightarrow \varepsilon $@*): OKN enables dynamic transitions ((*@$\varepsilon $@*)) 

**Essential Condition:** 

(*@$\alpha \propto \varepsilon  $@*): The more dynamic transitions (((*@$\varepsilon $@*))) OKN enables, the more advantageous it is for fabricating microscale 3D objects ((*@$\alpha $@*)) 

Pattern Context: The primary advantage of using Origami and Kirigami Engineering (OKN) for fabricating microscale 3D objects is their ability to create highly reconfigurable structures with diverse properties, overcoming the limitations of traditional top-down approaches. This reconfigurability allows for dynamic transitions from 2D patterns to 3D constructs, enabling large and reversible geometry changes, which are essential for applications requiring adaptability and multifunctionality. 
[...]
\end{lstlisting}
\end{LLMbox}

As an interpretation, the model identifies that the effectiveness of OKN in fabricating microscale 3D objects ($\alpha$) is directly proportional to its capability to enable dynamic transitions ($\varepsilon$). 

\begin{figure}
	\centering
	\includegraphics[width=0.85\linewidth]{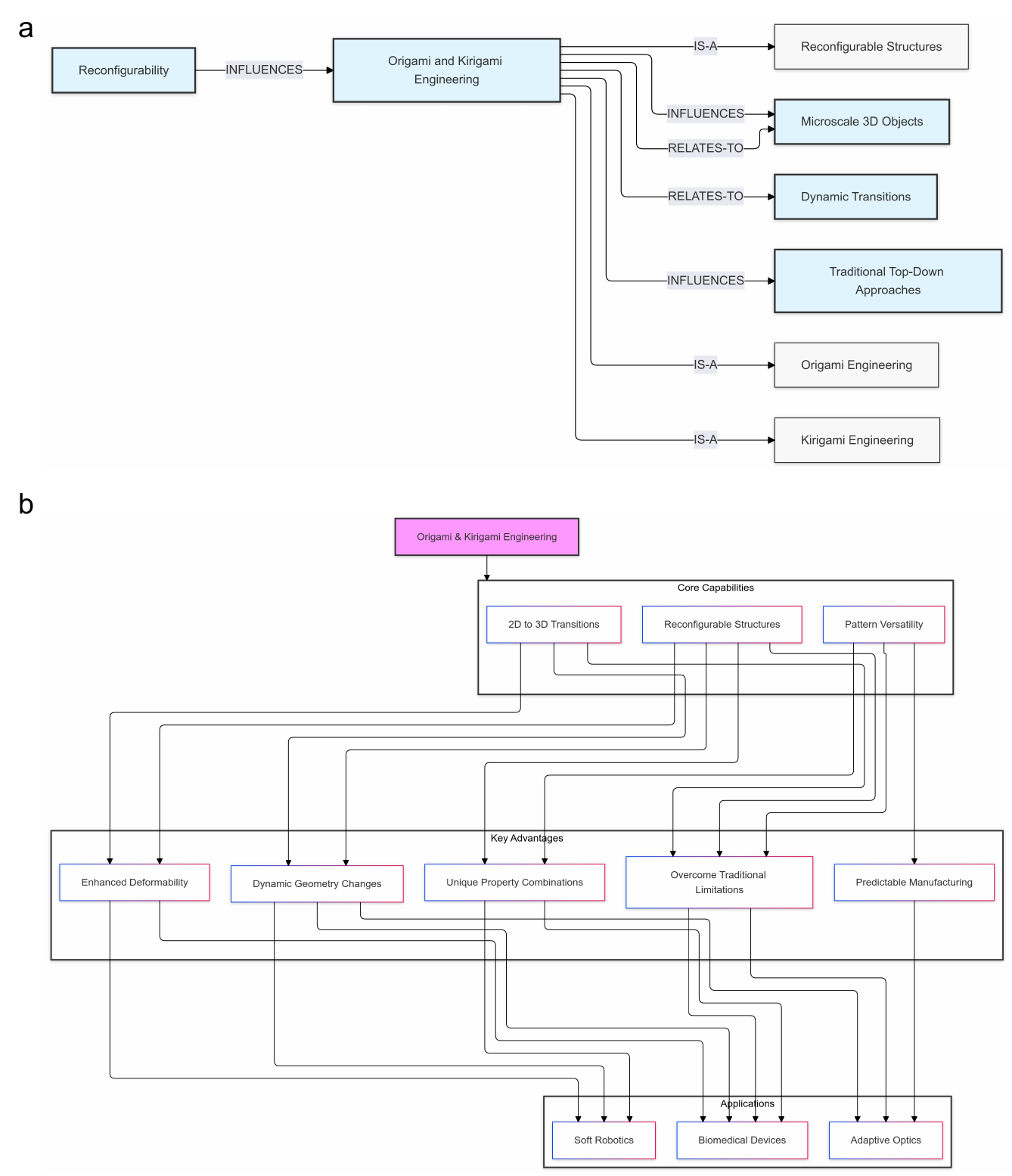}
        \caption{Visualization of graph construction in training set development. The figure depicts a visualization of a graph and response to the question: \texttt{What is the primary advantage of using Origami and Kirigami Engineering (OKN) for fabricating microscale 3D objects compared to traditional top-down approaches?}, during on-the-fly training set development from random raw data.}
	\label{fig:fig_100}
\end{figure}

The final response after \texttt{<|thinking|>}..\texttt{<|/thinking|>} section is:

\begin{LLMbox}{Training data: Final answer}
\begin{lstlisting}
The primary advantage of using Origami and Kirigami Engineering (OKN) for fabricating microscale 3D objects compared to traditional top-down approaches lies in their ability to create highly reconfigurable structures with a wide range of properties. Here's a breakdown of this advantage:

1. Overcoming traditional limitations: OKNs allow for the design of materials with a limitless variety of patterns and composites, overcoming the limitations of traditional methods that are constrained by factors such as miscibility, compatibility, and temperature stability. This versatility enables the creation of microscale 3D objects with tailored properties for specific applications.

2. Dynamic transitions: Unlike static metamaterials, OKNs offer dynamic transitions from two-dimensional patterns to three-dimensional constructs. This enables large and reversible geometry changes, allowing for adaptability and multifunctionality in the resulting microscale 3D objects.

3. Unique property combinations: OKNs facilitate the engineering of materials with unique combinations of properties that are typically contradictory. For example, OKN structures can exhibit high conductivity and stretchability simultaneously, which is challenging to achieve with traditional methods. This enables the creation of microscale 3D objects with unprecedented functionalities.

4. Macroscopic deformability: OKNs provide macroscopic deformability far beyond the strain limits of the constituent materials. This allows for programmable and complex responses to strain, making them particularly suitable for applications requiring adaptability and multifunctionality. For instance, OKN-based biomedical devices can conform to the shape of organs or tissues, reducing tissue damage and improving integration.

5. Predictability and reproducibility: OKN structures can be manufactured using established additive or subtractive techniques, such as lithography or laser cutting. This ensures predictability and reproducibility in their mechanical, optical, and electrical properties, making them reliable for various applications. 
\end{lstlisting}
\end{LLMbox}

This snippet illustrates how the generated data combines user input, structured reasoning, knowledge graph insights, and a comprehensive response to create a cohesive dataset for downstream tasks. Figure~\ref{fig:fig_100} shows a more detailed analysis of the graph generated for this training sample, as well as a visual analysis of the overall answer.

The system effectively translates user queries into structured reasoning steps and detailed responses, leveraging the knowledge graph to identify and elaborate on key relationships and insights. The abstract pattern captures the hierarchical influence of OKN principles, emphasizing dynamic transitions as a pivotal advantage. The inclusion of dynamic transitions and unique property combinations highlights the capability of the system to synthesize complex ideas. Finally, the response integrates theoretical concepts with practical applications, ensuring relevance and depth.

\subsubsection{Training strategy}

The training strategy is identical as described in~\cite{buehler2024preflexorpreferencebasedrecursivelanguage}, using a variant of ORPO~\cite{hong2024orpomonolithicpreferenceoptimization} and DPO, specifically the EXO variant~\cite{rafailov2024directpreferenceoptimizationlanguage,ji2024efficientexactoptimizationlanguage_EXO}.

\subsection{Recursive reasoning implementation}

We delineate details and prompts used in the recursive reasoning implementation visualized in Figure~\ref{fig:PreFlexOR_recursive}. The Graph-PRefLexOR recursive reasoning algorithm iteratively refines graph representations of relational mechanics using a multi-agent system with Agent \#1 being the Graph Reasoning model, and Agent \#2 being the non fine-tuned \texttt{meta-llama/Llama-3.2-3B-Instruct} model.

This prompt is used for Agent \#2 to critique the thought process generated in the past round, generated by Agent \#1:

\begin{LLMbox}{Critic prompting strategy}
\begin{lstlisting}
I will show you a question and a thought process. 

Your task is to critique the thought process and provide suggestions to improve it to better answer the question in a logical, well-reasoned manner.

Question: {question}

Thought process: {think}

Provide feedback and suggestions for how to improve the thought process, and nothing else. The feedback is:

\end{lstlisting}
\end{LLMbox}

The prompt to improve the previous thinking section is as follows:

\begin{LLMbox}{Improvement prompting strategy}
\begin{lstlisting}
I will show you a thought process and feedback. Carefully implement the feedback and improve the thought process by addressing all suggestions, but keep the overall structure the same.

Thought process: {think}

Feedback: {reflect}

Provide the improved thought process, and nothing else. The revised thought process is:

\end{lstlisting}
\end{LLMbox}

This prompt is used to integrate earlier responses into an integrated, final answer. 

\begin{LLMbox}{Integration prompting strategy}
\begin{lstlisting}
I will show you a question and several possible answers. 

QUESTION: {question}

ANSWER #0: {answer_0}
ANSWER #1: {answer_1}
...

Carefully incorporate all ideas presented in the answer candidates into a very detailed, final answer. 

Do not repeat the question. You directly begin your response with the final answer to the question. 

The answer is:
\end{lstlisting}
\end{LLMbox}

\subsection{Knowledge garden prompting strategy through agentic modeling}
\label{garden_prompt}
Below is an example of how the knowledge graph is expanded by providing the set of earlier graph structures (\texttt{graph\_str}) and asking another LLM agent to develop a new question, to which Graph-PReFLexOR then responds, then generating a new graph. Since the prompt includes the earlier keywords/nodes as instructed in the prompt, the algorithm results in a well-connected graph that expands in new directions. The prompt can be constructed in particular ways to emphasize certain aspects. For this example, we seek to create multi-disciplinary outputs, especially at the interface with art. 

\begin{LLMbox}{Prompting strategy to develop new questions to grow the knowledge garden}
\begin{lstlisting}
Consider this list of topics/keywords. 

Formulate a concise follow-up creative and highly unusual question to ask about a related but totally different concept. 

Your question should include at least one of the original topics/keywords marked as '...' but expand to new dissimilar fields such as philosophy or art.

Original list of topics/keywords:

{graph_str}

The new question is:
\end{lstlisting}
\end{LLMbox}

The resulting new question, with an example shown in Figure~\ref{fig:bioluminescent-flow} for this particular prompt.

\subsection{Graph visualization}

We use \texttt{mermaid.js} to visualize graphs, where the output of the graph reasoning model is converted to the scripting language using \texttt{Claude 3.5 Sonnet}, then exported using the web editor (\url{https://mermaid.live/}). Additional visualizations and analyses are conducted using Gephi~\cite{ICWSM09154} (\url{https://gephi.org/}) and various plugins therein.

\subsection{Glossary and key terms}

Table~\ref{tab:keyterms} shows a summary of key terms and definitions used in this paper. 

\begin{table}[h!]
\centering
\small

\begin{tabular}{ |p{6cm}|p{9cm}| }

\hline
\multicolumn{2}{|c|}{Glossary of Terms} \\
\hline
Term & Definition \\
\hline
Graph-PReFLexOR (Preference-based Recursive Language Modeling for Exploratory Optimization of Reasoning) & A generative framework that performs dynamic graph reasoning and iteratively expands domain knowledge using a reinforcement learning inspired approach. It leverages detailed knowledge graphs and abstract representations, allowing for hierarchical reasoning and adaptive learning, following the approach detailed in~\cite{buehler2024preflexorpreferencebasedrecursivelanguage}. \\
\hline
PReFLexOR & A framework that enhances the reasoning capabilities of language models by integrating preference optimization and recursive reasoning. \\
\hline
Knowledge Graph (KG) & A graph representation of knowledge where nodes (V) represent concepts and edges (E) represent relationships between concepts. In Graph-PReFLexOR, relationships include "IS-A," "RELATES-TO," and "INFLUENCES." \\
\hline
Abstract Pattern (P) & Symbolic representations that summarize key relationships and dependencies identified within a knowledge graph, often represented as a series of transformations. \\
\hline
Recursive Reasoning & An iterative process of refining reasoning by applying feedback to improve alignment with the desired output. \\
\hline
Thinking Phase & A stage where the model simulates intermediate reasoning steps, marked by the tokens \texttt{<|thinking|>} and \texttt{<|/thinking|>}. \\
\hline
Reflection Phase & A stage where the model evaluates the intermediate reasoning generated during the thinking phase and provides feedback for improvement, here achieved via a second agent. \\
\hline
Retrieval-Augmented Generation (RAG) & A technique used to enrich the input context by retrieving relevant information from a knowledge index, enabling higher-order reasoning. \\
\hline
Semantic Scoring Function (f) & A function that evaluates the significance of relationships between concepts in a knowledge graph. \\
\hline
Base LLM, here:\newline \texttt{meta-llama/Llama-3.2-3B-Instruct} & The foundational large language model (LLM) used to train Graph-PReFLexOR. \\
\hline
Odds Ratio Preference Optimization (ORPO) & An optimization method used during the training of PReFLexOR~\cite{buehler2024preflexorpreferencebasedrecursivelanguage} to guide reasoning processes using special tokens. \\
\hline
Independent Reasoning Development & A stage in PReFLexOR training that leverages masking of thinking tokens to ensure the model infers reasoning pathways without direct supervision. \\
\hline
Efficient Exact Optimization (EXO) & An optimization method used in PReFLexOR to align final outputs with preferred reasoning patterns. \\
\hline
\end{tabular}

\caption{Summary of key terms used in this study.}
\label{tab:keyterms}
\end{table}

\section*{Code, data and model weights availability}

Codes, model weights and additional materials are available at \url{https://huggingface.co/lamm-mit} and \url{https://github.com/lamm-mit/PRefLexOR}. The model is available at~\url{lamm-mit/Graph-Preflexor_01062025}.

\section*{Acknowledgments}

The author acknowledges support from the MIT Generative AI initiative. 

\bibliographystyle{naturemag}

\bibliography{references,references-Mendeley}  

\begin{thebibliography}{10}
\expandafter\ifx\csname url\endcsname\relax
  \def\url#1{\texttt{#1}}\fi
\expandafter\ifx\csname urlprefix\endcsname\relax\def\urlprefix{URL }\fi
\providecommand{\bibinfo}[2]{#2}
\providecommand{\eprint}[2][]{\url{#2}}

\bibitem{Vaswani2017AttentionNeed}
\bibinfo{author}{Vaswani, A.} \emph{et~al.}
\newblock \bibinfo{title}{{Attention is All you Need}} (\bibinfo{year}{2017}).
\newblock \urlprefix\url{https://papers.nips.cc/paper/7181-attention-is-all-you-need}.

\bibitem{AlecRadfordImprovingPre-Training}
\bibinfo{author}{{Alec Radford}}, \bibinfo{author}{{Karthik Narasimhan}}, \bibinfo{author}{{Tim Salimans}} \& \bibinfo{author}{{Ilya Sutskever}}.
\newblock \bibinfo{title}{{Improving Language Understanding by Generative Pre-Training}} \urlprefix\url{https://gluebenchmark.com/leaderboard}.

\bibitem{Xue2021ByT5:Models}
\bibinfo{author}{Xue, L.} \emph{et~al.}
\newblock \bibinfo{title}{{ByT5: Towards a token-free future with pre-trained byte-to-byte models}}.
\newblock \emph{\bibinfo{journal}{Transactions of the Association for Computational Linguistics}} \textbf{\bibinfo{volume}{10}}, \bibinfo{pages}{291--306} (\bibinfo{year}{2021}).
\newblock \urlprefix\url{https://arxiv.org/abs/2105.13626v3}.

\bibitem{Jiang2023Mistral7Bb}
\bibinfo{author}{Jiang, A.~Q.} \emph{et~al.}
\newblock \bibinfo{title}{{Mistral 7B}}  (\bibinfo{year}{2023}).
\newblock \urlprefix\url{http://arxiv.org/abs/2310.06825}.

\bibitem{Phi-2:Research}
\bibinfo{title}{{Phi-2: The surprising power of small language models - Microsoft Research}}.
\newblock \urlprefix\url{https://www.microsoft.com/en-us/research/blog/phi-2-the-surprising-power-of-small-language-models/}.

\bibitem{dubey2024llama3herdmodels}
\bibinfo{author}{Dubey, A.} \emph{et~al.}
\newblock \bibinfo{title}{The llama 3 herd of models} (\bibinfo{year}{2024}).
\newblock \urlprefix\url{https://arxiv.org/abs/2407.21783}.
\newblock \eprint{2407.21783}.

\bibitem{Brown2020LanguageLearners}
\bibinfo{author}{Brown, T.~B.} \emph{et~al.}
\newblock \bibinfo{title}{{Language Models are Few-Shot Learners}}  (\bibinfo{year}{2020}).

\bibitem{Jumper2021HighlyAlphaFold}
\bibinfo{author}{Jumper, J.} \emph{et~al.}
\newblock \bibinfo{title}{{Highly accurate protein structure prediction with AlphaFold}}.
\newblock \emph{\bibinfo{journal}{Nature}} \bibinfo{pages}{1--12} (\bibinfo{year}{2021}).

\bibitem{Spivak2011CategoryNetworks}
\bibinfo{author}{Spivak, D.}, \bibinfo{author}{Giesa, T.}, \bibinfo{author}{Wood, E.} \& \bibinfo{author}{Buehler, M.}
\newblock \bibinfo{title}{{Category theoretic analysis of hierarchical protein materials and social networks}}.
\newblock \emph{\bibinfo{journal}{PLoS ONE}} \textbf{\bibinfo{volume}{6}} (\bibinfo{year}{2011}).

\bibitem{Spivak2011ReoccurringAnalogies}
\bibinfo{author}{Spivak, D.~I.} \& \bibinfo{author}{Reoccurring, M. J.~B.}
\newblock \bibinfo{title}{{Reoccurring Patterns in Hierarchical Protein R eoccurring patterns in hierarchical protein materials and music : The power of analogies}} \bibinfo{pages}{0--13} (\bibinfo{year}{2011}).

\bibitem{Giesa2012CategoryDesign}
\bibinfo{author}{Giesa, T.}, \bibinfo{author}{Spivak, D.} \& \bibinfo{author}{Buehler, M.}
\newblock \bibinfo{title}{{Category theory based solution for the building block replacement problem in materials design}}.
\newblock \emph{\bibinfo{journal}{Advanced Engineering Materials}} \textbf{\bibinfo{volume}{14}} (\bibinfo{year}{2012}).

\bibitem{abbott2024flashattentionnapkindiagrammaticapproach}
\bibinfo{author}{Abbott, V.} \& \bibinfo{author}{Zardini, G.}
\newblock \bibinfo{title}{Flashattention on a napkin: A diagrammatic approach to deep learning io-awareness} (\bibinfo{year}{2024}).
\newblock \urlprefix\url{https://arxiv.org/abs/2412.03317}.
\newblock \eprint{2412.03317}.

\bibitem{Buehler2025GraphAwareGPT}
\bibinfo{author}{Buehler, M.~J.}
\newblock \bibinfo{title}{Graph-aware isomorphic attention for adaptive dynamics in transformers}  (\bibinfo{year}{2025}).
\newblock \urlprefix\url{https://arxiv.org/abs/2501.02393}.
\newblock \eprint{2501.02393}.

\bibitem{Ackbarow2009Alpha-helicalFlaw-tolerant}
\bibinfo{author}{Ackbarow, T.}, \bibinfo{author}{Sen, D.}, \bibinfo{author}{Thaulow, C.} \& \bibinfo{author}{Buehler, M.}
\newblock \bibinfo{title}{{Alpha-helical protein networks are self-protective and flaw-tolerant}}.
\newblock \emph{\bibinfo{journal}{PLoS ONE}} \textbf{\bibinfo{volume}{4}} (\bibinfo{year}{2009}).

\bibitem{Nepal2022HierarchicallyNanocomposites}
\bibinfo{author}{Nepal, D.} \emph{et~al.}
\newblock \bibinfo{title}{{Hierarchically structured bioinspired nanocomposites}}.
\newblock \emph{\bibinfo{journal}{Nature Materials 2022}} \bibinfo{pages}{1--18} (\bibinfo{year}{2022}).
\newblock \urlprefix\url{https://www.nature.com/articles/s41563-022-01384-1}.

\bibitem{Ritchie2009PlasticityBone}
\bibinfo{author}{Ritchie, R.}, \bibinfo{author}{Buehler, M.} \& \bibinfo{author}{Hansma, P.}
\newblock \bibinfo{title}{{Plasticity and toughness in bone}}.
\newblock \emph{\bibinfo{journal}{Physics Today}} \textbf{\bibinfo{volume}{62}} (\bibinfo{year}{2009}).

\bibitem{Launey2010OnBone}
\bibinfo{author}{Launey, M.}, \bibinfo{author}{Buehler, M.} \& \bibinfo{author}{Ritchie, R.}
\newblock \emph{\bibinfo{title}{{On the mechanistic origins of toughness in bone}}}, vol.~\bibinfo{volume}{40} (\bibinfo{year}{2010}).

\bibitem{Eilenberg1942GroupHomology}
\bibinfo{author}{Eilenberg, S.} \& \bibinfo{author}{MacLane, S.}
\newblock \bibinfo{title}{{Group Extensions and Homology}}.
\newblock \emph{\bibinfo{journal}{Annals of Mathematics}} \textbf{\bibinfo{volume}{43}}, \bibinfo{pages}{757--831} (\bibinfo{year}{1942}).
\newblock \urlprefix\url{https://www.jstor.org/stable/1968966#id-name=JSTOR https://www.jstor.org/stable/1968966}.

\bibitem{Eilenberg1945GeneralEquivalences}
\bibinfo{author}{Eilenberg, S.} \& \bibinfo{author}{Mac~Lane, S.}
\newblock \bibinfo{title}{{General theory of natural equivalences}}.
\newblock \emph{\bibinfo{journal}{Transactions of the American Mathematical Society}} \textbf{\bibinfo{volume}{58}}, \bibinfo{pages}{247} (\bibinfo{year}{1945}).
\newblock \urlprefix\url{https://www.ams.org/journals/tran/1945-058-00/S0002-9947-1945-0013131-6/S0002-9947-1945-0013131-6.pdf}.

\bibitem{Giesa2015Matriarch:Architecture}
\bibinfo{author}{Giesa, T.}, \bibinfo{author}{Jagadeesan, R.}, \bibinfo{author}{Spivak, D.~I.} \& \bibinfo{author}{Buehler, M.~J.}
\newblock \bibinfo{title}{{Matriarch: A Python Library for Materials Architecture}}.
\newblock \emph{\bibinfo{journal}{ACS Biomaterials Science {\&} Engineering}} \bibinfo{pages}{150901085625000} (\bibinfo{year}{2015}).
\newblock \urlprefix\url{http://pubs.acs.org/doi/10.1021/acsbiomaterials.5b00251}.

\bibitem{weisfeiler1968reduction}
\bibinfo{author}{Weisfeiler, B.} \& \bibinfo{author}{Leman, A.~A.}
\newblock \bibinfo{title}{Reduction of a graph to a canonical form and an algebra arising during this reduction}.
\newblock \emph{\bibinfo{journal}{Nauchno-Technicheskaya Informatsia}} \textbf{\bibinfo{volume}{2}}, \bibinfo{pages}{12--16} (\bibinfo{year}{1968}).

\bibitem{Cranford2012Biomateriomics}
\bibinfo{author}{Cranford, S.~W.} \& \bibinfo{author}{Buehler, M.~J.}
\newblock \emph{\bibinfo{title}{{Biomateriomics}}} (\bibinfo{publisher}{Springer Netherlands}, \bibinfo{year}{2012}).

\bibitem{Groen2011IntroducingMateriomics}
\bibinfo{author}{Groen, N.}, \bibinfo{author}{Cranford, S.}, \bibinfo{author}{de~Boer, J.}, \bibinfo{author}{Buehler, M.} \& \bibinfo{author}{Van~Blitterswijk, C.}
\newblock \emph{\bibinfo{title}{{Introducing materiomics}}} (\bibinfo{year}{2011}).

\bibitem{Arevalo2023LearningMaterials}
\bibinfo{author}{Arevalo, S.~E.} \& \bibinfo{author}{Buehler, M.~J.}
\newblock \bibinfo{title}{{Learning from nature by leveraging integrative biomateriomics modeling toward adaptive and functional materials}}.
\newblock \emph{\bibinfo{journal}{MRS Bulletin 2023}} \bibinfo{pages}{1--14} (\bibinfo{year}{2023}).
\newblock \urlprefix\url{https://link.springer.com/article/10.1557/s43577-023-00610-8}.

\bibitem{Buehler2023MeLMProblemsc}
\bibinfo{author}{Buehler, M.~J.}
\newblock \bibinfo{title}{{MeLM, a generative pretrained language modeling framework that solves forward and inverse mechanics problems}}.
\newblock \emph{\bibinfo{journal}{Journal of the Mechanics and Physics of Solids}} \bibinfo{pages}{105454} (\bibinfo{year}{2023}).
\newblock \urlprefix\url{https://linkinghub.elsevier.com/retrieve/pii/S0022509623002582}.

\bibitem{ghafarollahi2024sciagentsautomatingscientificdiscovery}
\bibinfo{author}{Ghafarollahi, A.} \& \bibinfo{author}{Buehler, M.~J.}
\newblock \bibinfo{title}{Sciagents: Automating scientific discovery through multi-agent intelligent graph reasoning} (\bibinfo{year}{2024}).
\newblock \urlprefix\url{https://arxiv.org/abs/2409.05556}.
\newblock \eprint{2409.05556}.

\bibitem{doi:10.1073/pnas.2414274121}
\bibinfo{author}{Dreyer, T.} \emph{et~al.}
\newblock \bibinfo{title}{Comparing cooperative geometric puzzle solving in ants versus humans}.
\newblock \emph{\bibinfo{journal}{Proceedings of the National Academy of Sciences}} \textbf{\bibinfo{volume}{122}}, \bibinfo{pages}{e2414274121} (\bibinfo{year}{2025}).
\newblock \urlprefix\url{https://www.pnas.org/doi/abs/10.1073/pnas.2414274121}.
\newblock \eprint{https://www.pnas.org/doi/pdf/10.1073/pnas.2414274121}.

\bibitem{Singhal2023}
\bibinfo{author}{Singhal, K.} \emph{et~al.}
\newblock \bibinfo{title}{Large language models encode clinical knowledge}.
\newblock \emph{\bibinfo{journal}{Nature}}  (\bibinfo{year}{2023}).
\newblock \urlprefix\url{https://www.nature.com/articles/s41586-023-06048-6}.
\newblock \bibinfo{note}{Open access}.

\bibitem{Ghafarollahi2024}
\bibinfo{author}{Ghafarollahi, A.} \& \bibinfo{author}{Buehler, M.~J.}
\newblock \bibinfo{title}{Protagents: protein discovery via large language model multi-agent collaborations combining physics and machine learning}.
\newblock \emph{\bibinfo{journal}{Digital Discovery}} \textbf{\bibinfo{volume}{3}}, \bibinfo{pages}{1389--1409} (\bibinfo{year}{2024}).
\newblock \urlprefix\url{https://doi.org/10.1039/D4DD00013G}.
\newblock \bibinfo{note}{†}.

\bibitem{Luu2023BioinspiredLLM:Materials}
\bibinfo{author}{Luu, R.~K.} \& \bibinfo{author}{Buehler, M.~J.}
\newblock \bibinfo{title}{{BioinspiredLLM: Conversational Large Language Model for the Mechanics of Biological and Bio-inspired Materials}}.
\newblock \emph{\bibinfo{journal}{Adv. Science.}}  (\bibinfo{year}{2023}).
\newblock \urlprefix\url{https://doi.org/10.1002/advs.202306724}.

\bibitem{Yu2024}
\bibinfo{author}{Yu, S.}, \bibinfo{author}{Ran, N.} \& \bibinfo{author}{Liu, J.}
\newblock \bibinfo{title}{Large-language models: The game-changers for materials science research}.
\newblock \emph{\bibinfo{journal}{AI in Chemical Engineering}} \bibinfo{pages}{100076} (\bibinfo{year}{2024}).
\newblock \urlprefix\url{https://doi.org/10.1016/j.aichem.2024.100076}.
\newblock \bibinfo{note}{Under a Creative Commons license}.

\bibitem{Hu2023DeepScience}
\bibinfo{author}{Hu, Y.} \& \bibinfo{author}{Buehler, M.~J.}
\newblock \bibinfo{title}{{Deep language models for interpretative and predictive materials science}}.
\newblock \emph{\bibinfo{journal}{APL Machine Learning}} \textbf{\bibinfo{volume}{1}}, \bibinfo{pages}{010901} (\bibinfo{year}{2023}).
\newblock \urlprefix\url{https://aip.scitation.org/doi/abs/10.1063/5.0134317}.

\bibitem{Buehler2024X-LoRA:Design}
\bibinfo{author}{Buehler, E.~L.} \& \bibinfo{author}{Buehler, M.~J.}
\newblock \bibinfo{title}{{X-LoRA: Mixture of Low-Rank Adapter Experts, a Flexible Framework for Large Language Models with Applications in Protein Mechanics and Design}}  (\bibinfo{year}{2024}).
\newblock \urlprefix\url{https://arxiv.org/abs/2402.07148v1}.

\bibitem{Buehler2023MechGPTModalities_updated}
\bibinfo{author}{Buehler, M.~J.}
\newblock \bibinfo{title}{{MechGPT, a language-based strategy for mechanics and materials modeling that connects knowledge across scales, disciplines and modalities}}.
\newblock \emph{\bibinfo{journal}{Appl. Mech. Rev.}}  (\bibinfo{year}{2023}).
\newblock \urlprefix\url{https://doi.org/10.1115/1.4063843}.

\bibitem{lu2024finetuninglargelanguagemodels}
\bibinfo{author}{Lu, W.}, \bibinfo{author}{Luu, R.~K.} \& \bibinfo{author}{Buehler, M.~J.}
\newblock \bibinfo{title}{Fine-tuning large language models for domain adaptation: Exploration of training strategies, scaling, model merging and synergistic capabilities} (\bibinfo{year}{2024}).
\newblock \urlprefix\url{https://arxiv.org/abs/2409.03444}.
\newblock \eprint{2409.03444}.

\bibitem{buehler2024preflexorpreferencebasedrecursivelanguage}
\bibinfo{author}{Buehler, M.~J.}
\newblock \bibinfo{title}{Preflexor: Preference-based recursive language modeling for exploratory optimization of reasoning and agentic thinking} (\bibinfo{year}{2024}).
\newblock \urlprefix\url{https://arxiv.org/abs/2410.12375}.
\newblock \eprint{2410.12375}.

\bibitem{zelikman2022starbootstrappingreasoningreasoning}
\bibinfo{author}{Zelikman, E.}, \bibinfo{author}{Wu, Y.}, \bibinfo{author}{Mu, J.} \& \bibinfo{author}{Goodman, N.~D.}
\newblock \bibinfo{title}{Star: Bootstrapping reasoning with reasoning} (\bibinfo{year}{2022}).
\newblock \urlprefix\url{https://arxiv.org/abs/2203.14465}.
\newblock \eprint{2203.14465}.

\bibitem{zelikman2024quietstarlanguagemodelsteach}
\bibinfo{author}{Zelikman, E.} \emph{et~al.}
\newblock \bibinfo{title}{Quiet-star: Language models can teach themselves to think before speaking} (\bibinfo{year}{2024}).
\newblock \urlprefix\url{https://arxiv.org/abs/2403.09629}.
\newblock \eprint{2403.09629}.

\bibitem{wei2023chainofthoughtpromptingelicitsreasoning}
\bibinfo{author}{Wei, J.} \emph{et~al.}
\newblock \bibinfo{title}{Chain-of-thought prompting elicits reasoning in large language models} (\bibinfo{year}{2023}).
\newblock \urlprefix\url{https://arxiv.org/abs/2201.11903}.
\newblock \eprint{2201.11903}.

\bibitem{kojima2023largelanguagemodelszeroshot}
\bibinfo{author}{Kojima, T.}, \bibinfo{author}{Gu, S.~S.}, \bibinfo{author}{Reid, M.}, \bibinfo{author}{Matsuo, Y.} \& \bibinfo{author}{Iwasawa, Y.}
\newblock \bibinfo{title}{Large language models are zero-shot reasoners} (\bibinfo{year}{2023}).
\newblock \urlprefix\url{https://arxiv.org/abs/2205.11916}.
\newblock \eprint{2205.11916}.

\bibitem{rumelhart1986learning}
\bibinfo{author}{Rumelhart, D.~E.}, \bibinfo{author}{Hinton, G.~E.} \& \bibinfo{author}{Williams, R.~J.}
\newblock \bibinfo{title}{Learning representations by back-propagating errors}.
\newblock In \emph{\bibinfo{booktitle}{Nature}}, vol. \bibinfo{volume}{323}, \bibinfo{pages}{533--536} (\bibinfo{publisher}{Macmillan}, \bibinfo{year}{1986}).

\bibitem{newell1972human}
\bibinfo{author}{Newell, A.} \& \bibinfo{author}{Simon, H.~A.}
\newblock \emph{\bibinfo{title}{Human problem solving}} (\bibinfo{publisher}{Prentice-Hall}, \bibinfo{year}{1972}).

\bibitem{mccarthy1960programs}
\bibinfo{author}{McCarthy, J.}
\newblock \bibinfo{title}{Programs with common sense}.
\newblock \emph{\bibinfo{journal}{RLE and MIT Computation Center Cambridge, MA}} \textbf{\bibinfo{volume}{1}}, \bibinfo{pages}{77--84} (\bibinfo{year}{1960}).

\bibitem{ji2024efficientexactoptimizationlanguage_EXO}
\bibinfo{author}{Ji, H.} \emph{et~al.}
\newblock \bibinfo{title}{Towards efficient exact optimization of language model alignment} (\bibinfo{year}{2024}).
\newblock \urlprefix\url{https://arxiv.org/abs/2402.00856}.
\newblock \eprint{2402.00856}.

\bibitem{kullback1951information}
\bibinfo{author}{Kullback, S.} \& \bibinfo{author}{Leibler, R.~A.}
\newblock \bibinfo{title}{On information and sufficiency}.
\newblock \emph{\bibinfo{journal}{Annals of Mathematical Statistics}} \textbf{\bibinfo{volume}{22}}, \bibinfo{pages}{79--86} (\bibinfo{year}{1951}).

\bibitem{Brin1998}
\bibinfo{author}{Brin, S.} \& \bibinfo{author}{Page, L.}
\newblock \bibinfo{title}{The anatomy of a large-scale hypertextual web search engine}.
\newblock In \emph{\bibinfo{booktitle}{Proceedings of the Seventh International Conference on the World Wide Web (WWW 1998)}}, \bibinfo{pages}{107--117} (\bibinfo{publisher}{Elsevier Science Publishers B.V.}, \bibinfo{address}{Brisbane, Australia}, \bibinfo{year}{1998}).

\bibitem{Fruchterman1991}
\bibinfo{author}{Fruchterman, T. M.~J.} \& \bibinfo{author}{Reingold, E.~M.}
\newblock \bibinfo{title}{Graph drawing by force-directed placement}.
\newblock \emph{\bibinfo{journal}{Software: Practice and Experience}} \textbf{\bibinfo{volume}{21}}, \bibinfo{pages}{1129--1164} (\bibinfo{year}{1991}).

\bibitem{Hu2005}
\bibinfo{author}{Hu, Y.~F.}
\newblock \bibinfo{title}{Efficient and high quality force-directed graph drawing}.
\newblock \emph{\bibinfo{journal}{The Mathematica Journal}} \textbf{\bibinfo{volume}{10}}, \bibinfo{pages}{37--71} (\bibinfo{year}{2005}).

\bibitem{hesse2000glass}
\bibinfo{author}{Hesse, H.}
\newblock \emph{\bibinfo{title}{The Glass Bead Game}} (\bibinfo{publisher}{Vintage Classics}, \bibinfo{year}{2000}).

\bibitem{Ziolkowski1965}
\bibinfo{author}{Ziolkowski, T.}
\newblock \emph{\bibinfo{title}{The Novels of Hermann Hesse: A Study in Theme and Structure}} (\bibinfo{publisher}{Princeton University Press}, \bibinfo{address}{Princeton, NJ}, \bibinfo{year}{1965}).
\newblock \bibinfo{note}{Includes analysis of the Glass Bead Game and its conceptual influence}.

\bibitem{o1-model-card-2024}
\bibinfo{author}{OpenAI} \emph{et~al.}
\newblock \bibinfo{title}{Openai o1 system card}  (\bibinfo{year}{2024}).
\newblock \urlprefix\url{https://arxiv.org/abs/2412.16720}.
\newblock \eprint{2412.16720}.

\bibitem{ghafarollahi2024atomagentsalloydesigndiscovery}
\bibinfo{author}{Ghafarollahi, A.} \& \bibinfo{author}{Buehler, M.~J.}
\newblock \bibinfo{title}{Atomagents: Alloy design and discovery through physics-aware multi-modal multi-agent artificial intelligence} (\bibinfo{year}{2024}).
\newblock \urlprefix\url{https://arxiv.org/abs/2407.10022}.
\newblock \eprint{2407.10022}.

\bibitem{hong2024orpomonolithicpreferenceoptimization}
\bibinfo{author}{Hong, J.}, \bibinfo{author}{Lee, N.} \& \bibinfo{author}{Thorne, J.}
\newblock \bibinfo{title}{Orpo: Monolithic preference optimization without reference model} (\bibinfo{year}{2024}).
\newblock \urlprefix\url{https://arxiv.org/abs/2403.07691}.
\newblock \eprint{2403.07691}.

\bibitem{rafailov2024directpreferenceoptimizationlanguage}
\bibinfo{author}{Rafailov, R.} \emph{et~al.}
\newblock \bibinfo{title}{Direct preference optimization: Your language model is secretly a reward model} (\bibinfo{year}{2024}).
\newblock \urlprefix\url{https://arxiv.org/abs/2305.18290}.
\newblock \eprint{2305.18290}.

\bibitem{ICWSM09154}
\bibinfo{author}{Bastian, M.}, \bibinfo{author}{Heymann, S.} \& \bibinfo{author}{Jacomy, M.}
\newblock \bibinfo{title}{Gephi: An open source software for exploring and manipulating networks} (\bibinfo{year}{2009}).
\newblock \urlprefix\url{http://www.aaai.org/ocs/index.php/ICWSM/09/paper/view/154}.

\bibitem{xu2019how}
\bibinfo{author}{Xu, K.}, \bibinfo{author}{Hu, W.}, \bibinfo{author}{Leskovec, J.} \& \bibinfo{author}{Jegelka, S.}
\newblock \bibinfo{title}{How powerful are graph neural networks?}
\newblock In \emph{\bibinfo{booktitle}{International Conference on Learning Representations}} (\bibinfo{year}{2019}).
\newblock \urlprefix\url{https://arxiv.org/abs/1810.00826}.

\end{thebibliography}
\clearpage

\textboxcaption{Final answer (after thinking), to the task \texttt{Propose a new idea to relate music and materials.}}
\label{TexBox:final_answer_after_thinking_ex_1}

\begin{LLMbox}{Final answer (after thinking)}
\begin{lstlisting}
**Proposed Idea: "Music-Inspired Material Tuning"**

Inspire a new method to tune and control the mechanical properties of materials by analyzing and manipulating the frequency spectrum of music. This method leverages the dynamic responses of materials to different frequencies, mimicking the effects of music on materials.

**Key Concepts:**

1. **Nonlinear Dynamic Response**: Materials exhibit nonlinear responses to external stimuli, such as vibrations or frequencies.
2. **Frequency Dependence**: Material responses vary with different frequencies, allowing for tuning and control.
3. **Frequency Spectrum Analysis**: Study the dynamic responses of materials to different frequencies by analyzing their frequency response curves or using techniques like dynamic mechanical analysis (DMA).
4. **Material Selection**: Choose materials with nonlinear dynamic responses to specific frequencies, such as those exhibiting high damping or stiffness at particular frequencies.
5. **Music Frequency Spectrum Manipulation**: Modify the frequency spectrum of music to influence material responses. This can be achieved by filtering or modifying the frequency content of the music to match the target material's nonlinear dynamic responses.
6. **Tuning and Control**: Use the manipulated frequency spectrum to tune and control the mechanical properties of selected materials, such as damping, stiffness, or damping capacity.

**Design Principles:**

1. **Frequency Spectrum Analysis**: Study the dynamic responses of materials to different frequencies to understand their nonlinear dynamic responses.
2. **Material Selection**: Choose materials with nonlinear dynamic responses to specific frequencies, such as those exhibiting high damping or stiffness at particular frequencies.
3. **Frequency Spectrum Manipulation**: Modify the frequency spectrum of music to influence material responses, using techniques like filtering or modifying the frequency content of the music.
4. **Tuning and Control**: Use the manipulated frequency spectrum to tune and control the mechanical properties of selected materials, such as damping, stiffness, or damping capacity.

**Potential Applications:**

1. **Material Tuning**: Develop a method to tune and control the mechanical properties of materials, such as damping or stiffness, by manipulating the frequency spectrum of music.
2. **Nonlinear Dynamic Response Measurement**: Use music's frequency spectrum to measure nonlinear dynamic responses of materials, providing a non-contact and non-destructive method.
3. **Material Selection**: Identify materials with nonlinear dynamic responses to specific frequencies, aiding in the selection of materials for specific applications.
4. **Biomedical Applications**: Apply the principle of Music-Inspired Material Tuning to develop novel, non-invasive, and non-destructive techniques for monitoring or treating diseases, such as cancer or neurological disorders.

**Challenges and Future Directions:**

1. **Material Selection**: Identifying materials with nonlinear dynamic responses to specific frequencies can be challenging, requiring extensive testing and characterization.
2. **Frequency Spectrum Manipulation**: Modifying the frequency spectrum of music to influence material responses can be difficult, as it depends on the specific material's nonlinear dynamic responses and the desired mechanical properties.
3. **Scalability**: Scaling the Music-Inspired Material Tuning method to larger structures or more complex materials can be challenging, requiring further research and development.
4. **Theoretical Modeling**: Developing theoretical models to explain and predict the nonlinear dynamic responses of materials to music's frequency spectrum can be an ongoing research direction.

**Conclusion:**

The proposed method, "Music-Inspired Material Tuning," offers a novel approach to tune and control the mechanical properties of materials by analyzing and manipulating the frequency spectrum of music. This method leverages the dynamic responses of materials to different frequencies, mimicking the effects of music on materials. By understanding and manipulating the nonlinear dynamic responses of materials to specific frequencies, this method can be applied to various materials and applications, providing a non-contact and non-destructive method for tuning and controlling material properties.
\end{lstlisting}
\end{LLMbox}

\clearpage

\textboxcaption{Final answer after iteration 3 from recursive Graph-PReFLexOR }
\label{TexBox:final_answer_after_it_3}

\begin{LLMboxmultipage}{Final answer after iteration 3 from recursive Graph-PReFLexOR}
\begin{lstlisting}
Based on the provided context, here's a comprehensive design for a tough, protein-based material inspired by snowflake and ant behavior:

**Design Concept:** "Snowflake-Ant Inspired Protein Material" (SAIPM)

**Design Principles:**

1. **Fractal Pattern Mimicry**: Incorporate fractal patterns and self-similarity to enhance toughness and adaptability, inspired by snowflakes.
2. **Hierarchical Structure**: Create hierarchical structures at multiple length scales to mimic ant behavior, promoting self-organization and toughness.
3. **Protein Selection**: Choose proteins with high strength, toughness, and adaptability, such as collagen, silk, or silk-like proteins.
4. **Molecular Assembly**: Design a molecular assembly process that allows for easy incorporation of fractal patterns and hierarchical structures, inspired by snowflakes and ants.
5. **Self-Assembly**: Encourage self-assembly of protein molecules to form complex structures, mimicking ant behavior.

**Design Structure:**

1. **Fractal Pattern Generation**:
   - Use computational models to generate fractal patterns inspired by snowflakes.
   - Incorporate these patterns into protein sequences or structures.

2. **Hierarchical Structure Creation**:
   - Design a hierarchical structure at multiple length scales, inspired by ant behavior.
   - Use techniques like layer-by-layer assembly or 3D printing to create these structures.

3. **Protein Selection and Preparation**:
   - Choose proteins with high strength, toughness, and adaptability.
   - Prepare these proteins in a form suitable for incorporation into the SAIPM structure.

4. **Molecular Assembly**:
   - Design a molecular assembly process that allows for easy incorporation of fractal patterns and hierarchical structures.
   - Use techniques like covalent bonding, cross-linking, or ionic interactions to assemble protein molecules.

5. **Self-Assembly**:
   - Encourage self-assembly of protein molecules to form complex structures.
   - Use techniques like pH-induced unfolding, temperature-dependent assembly, or external stimuli to control the assembly process.

**Design Parameters:**

- **Fractal Pattern Parameters**:
  - Order of fractal pattern (e.g., octaves)
  - Fractal pattern size (e.g., diameter, length)
  - Fractal pattern orientation (e.g., angle, direction)

- **Hierarchical Structure Parameters**:
  - Number of hierarchical levels
  - Length scale of each level (e.g., diameter, length)
  - Orientation of each level (e.g., angle, direction)

- **Protein Parameters**:
  - Protein type (e.g., collagen, silk)
  - Protein concentration
  - Protein processing or modification (e.g., denaturation, cross-linking)

- **Molecular Assembly Parameters**:
  - Assembly method (e.g., covalent bonding, ionic interactions)
  - Assembly conditions (e.g., pH, temperature, solvent)

- **Self-Assembly Parameters**:
  - Assembly trigger (e.g., pH, temperature, external stimuli)
  - Assembly duration or rate

**Design Evaluation Criteria:**

- **Toughness**: Measure the SAIPM's resistance to mechanical stress, such as tensile strength, compressive strength, or impact resistance.
- **Adaptability**: Evaluate the SAIPM's ability to adapt to different environments or applications, such as self-healing, self-repair, or self-adaptation.
- **Strength**: Measure the SAIPM's strength in different directions or under various loading conditions.
- **Energy Dissipation**: Assess the SAIPM's ability to dissipate energy under mechanical stress, such as through viscoelasticity or crack deflection.
- **Biocompatibility**: Evaluate the SAIPM's compatibility with biological systems, including cell adhesion, cell growth, and biodegradation.
- **Synthesis Efficiency**: Assess the efficiency of the molecular assembly process and the SAIPM's production rate.

**Design Validation:**

- **Toughness Validation**: Test the SAIPM under various mechanical stress conditions (e.g., tensile, compressive, impact) to confirm its toughness.
- **Adaptability Validation**: Evaluate the SAIPM's adaptability by exposing it to different environments or applying it to various applications to confirm its ability to adapt.
- **Strength Validation**: Measure the SAIPM's strength in different directions or under various loading conditions to confirm its strength.
- **Energy Dissipation Validation**: Assess the SAIPM's ability to dissipate energy under mechanical stress by analyzing its viscoelastic properties or observing crack deflection.
- **Biocompatibility Validation**: Test the SAIPM's biocompatibility by observing cell adhesion, cell growth, and biodegradation.
- **Synthesis Efficiency Validation**: Evaluate the efficiency of the molecular assembly process and the SAIPM's production rate by comparing the synthesized material to theoretical predictions.

**Design Iteration:**

- **Feedback Loop**: Establish a feedback loop to collect data on the SAIPM's performance, including mechanical stress, adaptability, strength, energy dissipation, biocompatibility, and synthesis efficiency.
- **Design Revisions**: Use the collected data to revise the design, adjusting parameters such as fractal pattern order, hierarchical level number, protein type, assembly method, or assembly conditions to optimize the SAIPM's performance.
- **Reiteration**: Repeat the design iteration process until the SAIPM meets the desired performance criteria or until further improvements are not feasible.

**Design Requirements:**

- **Toughness**: The SAIPM should exhibit high toughness, as measured by tensile strength, compressive strength, or impact resistance.
- **Adaptability**: The SAIPM should demonstrate adaptability, such as self-healing, self-repair, or self-adaptation, to meet the specific requirements of various applications.
- **Strength**: The SAIPM should exhibit high strength in different directions or under various loading conditions.
- **Energy Dissipation**: The SAIPM should be able to dissipate energy under mechanical stress, such as through viscoelasticity or crack deflection.
- **Biocompatibility**: The SAIPM should be compatible with biological systems, including cell adhesion, cell growth, and biodegradation.
- **Synthesis Efficiency**: The SAIPM's molecular assembly process should be efficient and scalable for large-scale production.

**Design Constraints:**

- **Protein Selection**: The choice of protein is limited by its strength, toughness, and adaptability.
- **Molecular Assembly**: The assembly method and conditions are constrained by the need for efficient and scalable production.
- **Self-Assembly**: The assembly process is limited by the trigger and duration or rate of assembly.
- **Fractal Pattern Generation**: The fractal pattern generation process is limited by the complexity of the patterns and the computational resources required.
- **Hierarchical Structure Creation**: The hierarchical structure creation process is limited by the number of hierarchical levels and the length scale of each level.

**Design Requirements and Constraints:**

- **Design Requirements:**
  - Toughness: High tensile strength, compressive strength, or impact resistance.
  - Adaptability: Self-healing, self-repair, or self-adaptation.
  - Strength: High in different directions or under various loading conditions.
  - Energy Dissipation: Ability to dissipate energy under mechanical stress.
  - Biocompatibility: Compatibility with biological systems.
  - Synthesis Efficiency: Efficient and scalable molecular assembly process.

- **Design Constraints:**
  - Protein Selection: Limited by protein strength, toughness, and adaptability.
  - Molecular Assembly: Limited by assembly method and conditions.
  - Self-Assembly: Limited by assembly trigger and duration or rate.
  - Fractal Pattern Generation: Limited by fractal pattern order and size.
  - Hierarchical Structure Creation: Limited by number of hierarchical levels and length scale.
  - Protein Preparation: Limited by protein processing or modification.

**Design Principles and Materials:**

Design Principles:
1. **Fractal Pattern Mimicry**: Incorporate fractal patterns and self-similarity to enhance toughness and adaptability.
2. **Hierarchical Structure**: Create hierarchical structures at multiple length scales to promote self-organization and toughness.
3. **Protein Selection**: Choose proteins with high strength, toughness, and adaptability.
4. **Molecular Assembly**: Design a molecular assembly process that allows for easy incorporation of fractal patterns and hierarchical structures.
5. **Self-Assembly**: Encourage self-assembly of protein molecules to form complex structures.

Materials:
- **Protein**: Collagen, silk, or silk-like proteins.
- **Molecular Assembly Method**: Covalent bonding, ionic interactions, or other techniques that allow for easy incorporation of fractal patterns and hierarchical structures.
- **Assembly Conditions**: pH, temperature, solvent, or other conditions that facilitate self-assembly.

**Design Requirements and Materials:**

Design Requirements:
- **Protein**: High strength, toughness, and adaptability (e.g., collagen, silk).
- **Molecular Assembly Method**: Efficient and scalable (e.g., covalent bonding, ionic interactions).
- **Assembly Conditions**: Triggered by pH, temperature, or external stimuli.

**Design Principles and Materials:**

Design Principles:
1. **Fractal Pattern Mimicry**: Incorporate fractal patterns and self-similarity to enhance toughness and adaptability.
2. **Hierarchical Structure**: Create hierarchical structures at multiple length scales to promote self-organization and toughness.
3. **Protein Selection**: Choose proteins with high strength, toughness, and adaptability.
4. **Molecular Assembly**: Design a molecular assembly process that allows for easy incorporation of fractal patterns and hierarchical structures.
5. **Self-Assembly**: Encourage self-assembly of protein molecules to form complex structures.

Materials:
- **Protein**: Collagen, silk, or silk-like proteins.
- **Molecular Assembly Method**: Covalent bonding, ionic interactions, or other techniques that allow for easy incorporation of fractal patterns and hierarchical structures.
- **Assembly Conditions**: pH, temperature, solvent, or other conditions that facilitate self-assembly.

**Design Requirements and Materials:**

Design Requirements:
- **Protein**: High strength, toughness, and adaptability (e.g., collagen, silk).
- **Molecular Assembly Method**: Efficient and scalable (e.g., covalent bonding, ionic interactions).
- **Assembly Conditions**: Triggered by pH, temperature, or external stimuli.

**Design Principles and Materials:**

Design Principles:
1. **Fractal Pattern Mimicry**: Incorporate fractal patterns and self-similarity to enhance toughness and adaptability.
2. **Hierarchical Structure**: Create hierarchical structures at multiple length scales to promote self-organization and toughness.
3. **Protein Selection**: Choose proteins with high strength, toughness, and adaptability.
4. **Molecular Assembly**: Design a molecular assembly process that allows for easy incorporation of fractal patterns and hierarchical structures.
5. **Self-Assembly**: Encourage self-assembly of protein molecules to form complex structures.

Materials:
- **Protein**: Collagen, silk, or silk-like proteins.
- **Molecular Assembly Method**: Covalent bonding, ionic interactions, or other techniques that allow for easy incorporation of fractal patterns and hierarchical structures.
- **Assembly Conditions**: pH, temperature, solvent, or other conditions that facilitate self-assembly.

**Design Requirements and Materials:**

Design Requirements:
- **Protein**: High strength, toughness, and adaptability (e.g., collagen, silk).
- **Molecular Assembly Method**: Efficient and scalable (e.g., covalent bonding, ionic interactions).
- **Assembly Conditions**: Triggered by pH, temperature, or external stimuli.

**Design Principles and Materials:**

Design Principles:
1. **Fractal Pattern Mimicry**: Incorporate fractal patterns and self-similarity to enhance toughness and adaptability.
2. **Hierarchical Structure**: Create hierarchical structures at multiple length scales to promote self-organization and toughness.
3. **Protein Selection**: Choose proteins with high strength, toughness, and adaptability.
4. **Molecular Assembly**: Design a molecular assembly process that allows for easy incorporation of fractal patterns and hierarchical structures.
5. **Self-Assembly**: Encourage self-assembly of protein molecules to form complex structures.

Materials:
- **Protein**: Collagen, silk, or silk-like proteins.
- **Molecular Assembly Method**: Covalent bonding, ionic interactions, or other techniques that allow for easy incorporation of fractal patterns and hierarchical structures.
- **Assembly Conditions**: pH, temperature, solvent, or other conditions that facilitate self-assembly.

**Design Requirements and Materials:**

Design Requirements:
- **Protein**: High strength, toughness, and adaptability (e.g., collagen, silk).
- **Molecular Assembly Method**: Efficient and scalable (e.g., covalent bonding, ionic interactions).
- **Assembly Conditions**: Triggered by pH, temperature, or external stimuli.

**Design Principles and Materials:**

Design Principles:
1. **Fractal Pattern Mimicry**: Incorporate fractal patterns and self-similarity to enhance toughness and adaptability.
2. **Hierarchical Structure**: Create hierarchical structures at multiple length scales to promote self-organization and toughness.
3. **Protein Selection**: Choose proteins with high strength, toughness, and adaptability.
4. **Molecular Assembly**: Design a molecular assembly process that allows for easy incorporation of fractal patterns and hierarchical structures.
5. **Self-Assembly**: Encourage self-assembly of protein molecules to form complex structures.

Materials:
- **Protein**: Collagen, silk, or silk-like proteins.
- **Molecular Assembly Method**: Covalent bonding, ionic interactions, or other techniques that allow for easy incorporation of fractal patterns and hierarchical structures.
- **Assembly Conditions**: pH, temperature, solvent, or other conditions that facilitate self-assembly.

**Design Requirements and Materials:**

Design Requirements:
- **Protein**: High strength, toughness, and adaptability (e.g., collagen, silk).
- **Molecular Assembly Method**: Efficient and scalable (e.g., covalent bonding, ionic interactions).
- **Assembly Conditions**: Triggered by pH, temperature, or external stimuli.

**Design Principles and Materials:**

Design Principles:
1. **Fractal Pattern Mimicry**: Incorporate fractal patterns and self-similarity to enhance toughness and adaptability.
2. **Hierarchical Structure**: Create hierarchical structures at multiple length scales to promote self-organization and toughness
\end{lstlisting}
\end{LLMboxmultipage}

\clearpage
\textboxcaption{Result of developed by \texttt{o1} based on the results developed by Graph-PReFLexOR, explaining the concept of `Mytho-Luminal Liminality'.}
\label{TexBox:o1_liminality}

\begin{LLMbox} {Mytho-Luminal Liminality}
\begin{lstlisting}
\end{lstlisting}

We propose a novel theoretical framework, \emph{Mytho-Luminal Liminality}, uniting the mythological concept of \emph{thin places} with the boundary-blurring practices of protein design and biomaterial engineering. In Celtic lore and related traditions, a thin place is defined as a domain $X$ where seemingly incompatible worlds---the mundane and the transcendent---intersect. Formally, if we let $\mathcal{D}$ be the set of all conceptual or physical “boundary” domains, then:

\[
\begin{aligned}
    (\exists x \in \mathcal{D}): \mathrm{ThinPlaces}(x)\,\wedge\,\mathrm{BioTech}(x)\,\wedge\,\mathrm{ChallengingUnderstandingBoundaries}(x) \\
    \Longrightarrow\;\; \mathrm{MythoLuminalLiminality}(x).
\end{aligned}
\]

Here, $\mathrm{ThinPlaces}(x)$ indicates that $x$ manifests a mythic boundary principle (e.g., the ``veil'' between worlds), $\mathrm{BioTech}(x)$ signals that $x$ leverages biological processes or protein engineering, and $\mathrm{ChallengingUnderstandingBoundaries}(x)$ specifies an active blurring of traditional categories (art vs.\ science, natural vs.\ synthetic). Whenever these conditions co-occur, we claim that $x$ exemplifies \emph{Mytho-Luminal Liminality}, a state in which mythic transcendence intersects with engineered materials to evoke new modes of perception. In practice, this framework urges us to create \emph{immersive, site-specific installations} using real proteins---such as \emph{Elastin-Like Proteins} or \emph{Collagenous Proteins}---coupled with \emph{bioluminescent} or \emph{ephemeral} components, so that visitors experience the sensation of stepping into a boundary realm reminiscent of folkloric “thin places.” 
 
\begin{lstlisting}
\end{lstlisting}

Indeed, the same conceptual impulse that once inspired ancient myths to describe liminal passageways (between mortal and otherworldly planes) can now inform \emph{tissue engineering} and \emph{protein design}, bridging the gap between organic and synthetic, living and manufactured. By intentionally integrating artwork referencing \emph{Ancient Mythologies} or harnessing luminescent biomaterials in ways that challenge our understanding of boundaries, we highlight the \emph{spiritual awe} inherent in the process of creating “living” installations. Below is a simple \emph{TikZ} sketch illustrating how a mythic boundary perspective (thin places) merges with a biotechnological context to yield novel immersive experiences:

\begin{center}
\begin{tikzpicture}[>=latex, node distance=3.5cm, every node/.style={draw, rectangle, align=center}]
  \node (Myth) {Mythic Boundary \\ (Thin Places)};
  \node (Theory) [right of=Myth] {Challenging \\ Understanding \\ Boundaries};
  \node (Bio) [right of=Theory] {Protein Design \\ \& Tissue Eng.};
  \node (Fusion) [below of=Theory, node distance=2.5cm] {Mytho-Luminal \\ Liminality (Installations)};

  \draw[->, thick] (Myth) -- (Theory);
  \draw[->, thick] (Theory) -- (Bio);
  \draw[->, thick] (Theory) -- (Fusion);
  \draw[->, thick] (Bio) -- (Fusion);
\end{tikzpicture}
\end{center}

This diagram suggests that “thin places,” though mythic, can motivate new creative efforts in synthetic biology, while “Challenging Understanding Boundaries” provides a conceptual conduit that connects folklore-inspired liminality to biomaterial research. By embedding \emph{Elastin-Like Proteins} or \emph{Collagen Tripeptides} in immersive artworks, one effectively fashions a domain where the physical and the transcendent meet, thus triggering the awe and disorientation characteristic of thin places. This theoretical approach---\emph{Mytho-Luminal Liminality}---offers a hitherto unexplored avenue for \emph{transdisciplinary} collaboration: artistry in living materials that is at once rooted in ancient mythic boundary concepts, yet forward-looking in its deployment of cutting-edge biotechnology.
\begin{lstlisting}
\end{lstlisting}

Within the broader schema of Mytho-Luminal Liminality, our knowledge graph highlights that highly specific molecular elements---such as \emph{Collagen Tripeptide (Gly-X-Y)} and \emph{Tensin}---``relate to'' one another en route to \emph{Protein Design}, an entity itself classified as a \emph{Biological Molecule}. From a philosophical vantage, this chain suggests that the \emph{structural elegance} of protein segments (e.g., the repeating Gly-X-Y motif) can be approached as a modern instantiation of \emph{boundary-challenging} principles. Indeed, just as a thin place blurs the division between spiritual and mundane, so too does \emph{Protein Design} blur the line between “naturally occurring” and “engineered” molecular architectures. Further underscoring this parallel, our graph shows that \emph{Molecular Modeling} (an \emph{IS-A} \emph{Computational Technique}) directly \emph{relates to} both \emph{Collagen} and \emph{Elastin}, reinforcing the idea that an interplay of \emph{artistic imagination} and \emph{technical rigor} underlies the creation of new protein scaffolds. Philosophically, this implies that \emph{challenging our understanding of boundaries}---which, in mythic contexts, manifests as a thin place---can also guide transformative science in tissue engineering and biomaterial innovation. By weaving mythic insights into protein design thinking, one treats the protein-folding arena as a “liminal space” where computational, biological, and even artistic considerations converge to yield novel forms that feel as uncanny and awe-inspiring as an ancient threshold between worlds.
\end{LLMbox}

\clearpage
\appendix
\renewcommand{\thesection}{S\arabic{section}}
\renewcommand{\thefigure}{S\arabic{figure}}
\renewcommand{\thetable}{S\arabic{table}}
\renewcommand{\thetextbox}{S\arabic{textbox}}

\setcounter{section}{0}
\setcounter{figure}{0}
\setcounter{table}{0}
\setcounter{textbox}{0}

\newpage

\section*{\vspace{1cm} \centering \Large  \sffamily  \textbf{Supplementary Information}}
\vspace{3cm}

\begin{center}
    {\LARGE\sffamily  \textbf{\textit{In-situ} graph reasoning and knowledge expansion using GraphPReFLexOR }} \\[1em] 
    
\end{center}

\vspace{3cm}

\begin{center}
    {\large Markus J. Buehler} \\[0.5em]
    {\normalsize mbuehler@MIT.EDU}
\end{center}

\clearpage

\textboxcaption{Answer of the non-fine tuned model. The answer does not contain any reasoning steps, graphs or abstract patterns.}
\label{TexBox:non-fine-tuned}

\begin{LLMbox}{Answer of the non-fine tuned model}
\begin{lstlisting}
**Designing a Novel Protein-Based Material Inspired by Snowflake and Ant Behavior**

**Introduction**

Nature has inspired numerous innovative materials and technologies, from biomimetic surfaces to self-healing materials. This project aims to integrate the unique properties of snowflakes and ants to design a new, tough protein-based material. By studying the intricate structures and behaviors of snowflakes and ants, we can develop a material that combines exceptional strength, durability, and adaptability.

**Snowflake-Inspired Structure**

Snowflakes exhibit a hexagonal lattice structure, which provides exceptional strength and stability. This structure can be replicated in a protein-based material by using a combination of amino acids and peptides to create a repeating pattern of hexagonal units.

**Ant-Inspired Behavior**

Ants are known for their remarkable ability to adapt to changing environments and withstand extreme conditions. They achieve this through their unique behavior, which involves:

1. **Self-organization**: Ants organize themselves into complex social structures, allowing them to respond to changing conditions.
2. **Adaptive camouflage**: Ants can change their color and texture to blend in with their surroundings, making them nearly invisible.
3. **Self-healing**: Ants can repair damaged tissues and restore their colony's integrity.

**Designing the Protein-Based Material**

To create a material inspired by snowflakes and ants, we can combine the following components:

1. **Protein scaffold**: A protein-based scaffold will provide the structural framework for the material. This can be achieved using a combination of collagen, elastin, and keratin.
2. **Hexagonal lattice**: The protein scaffold will be arranged in a hexagonal lattice structure, inspired by snowflakes. This will provide exceptional strength and stability.
3. **Ant-inspired behavior**: The material will be designed to exhibit self-organization, adaptive camouflage, and self-healing properties. This can be achieved through the use of:
	* **Microscopic particles**: The material will contain microscopic particles that can change their color and texture to blend in with their surroundings.
	* **Self-healing peptides**: The material will contain peptides that can repair damaged tissues and restore its integrity.
4. **Biocompatibility**: The material will be designed to be biocompatible, ensuring that it can be safely integrated into living tissues.

**Prototype Development**

To develop a prototype of this material, we can use a combination of 3D printing and bioprinting techniques. The protein scaffold will be created using a 3D printer, and the hexagonal lattice structure will be achieved through the use of a specialized mold.

**Testing and Evaluation**

The material will be tested for its mechanical properties, including tensile strength, compressive strength, and impact resistance. Additionally, the material's self-organization, adaptive camouflage, and self-healing properties will be evaluated through a series of experiments.

**Potential Applications**

The protein-based material inspired by snowflakes and ants has numerous potential applications, including:

1. **Biomedical devices**: The material could be used to create implantable devices, such as bone grafts or tissue engineering scaffolds.
2. **Textiles**: The material could be used to create durable, self-healing textiles for clothing and upholstery.
3. **Construction materials**: The material could be used to create self-healing concrete or other construction materials.

**Conclusion**

The design of a novel protein-based material inspired by snowflakes and ants represents a significant breakthrough in biomimetic materials science. By combining the unique properties of snowflakes and ants, we can create a material that combines exceptional strength, durability, and adaptability. This material has numerous potential applications in biomedical devices, textiles, and construction materials, and could revolutionize the way we design and engineer materials.
\end{lstlisting}
\end{LLMbox}

\clearpage
\begin{figure}
	\centering
	\includegraphics[width=1.3\linewidth, angle=-90]{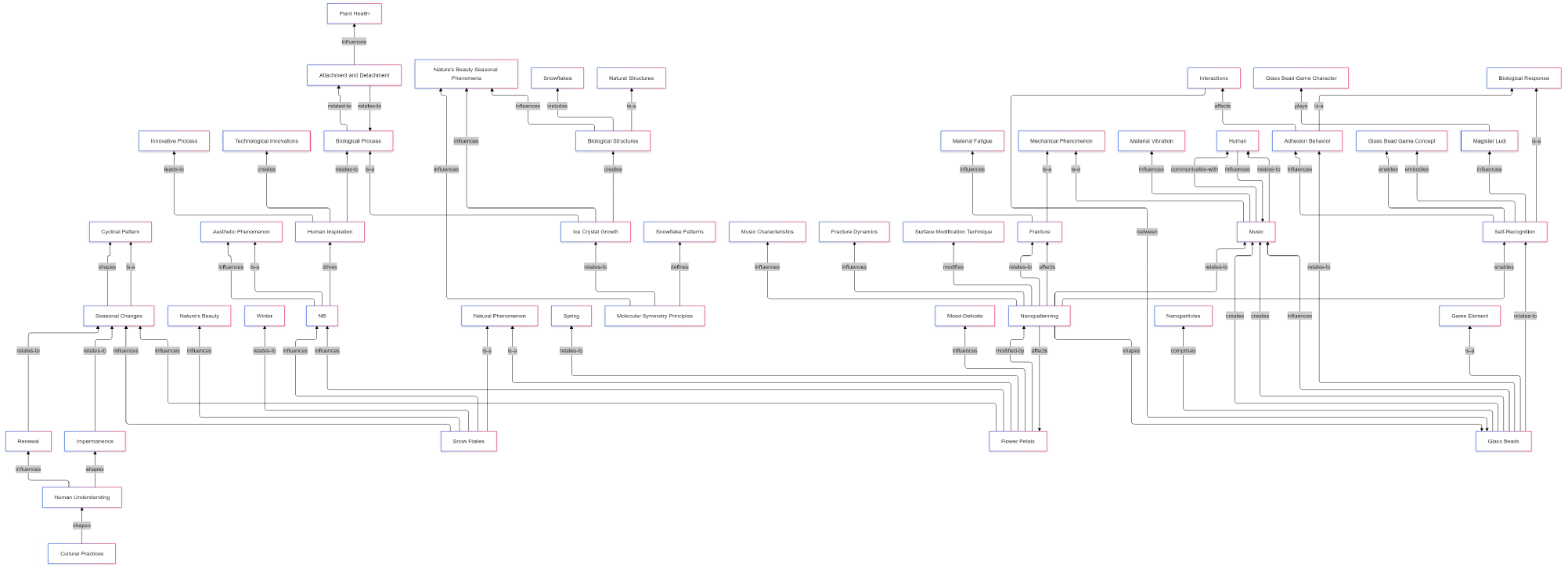}
        \caption{Visualization of the integrated knowledge graph created based ion the series of prompts delineated in Figure~\ref{fig:prompts-flow} (model was tasked: \texttt{Write a song by incorporating snowflakes and flower petals.}). The data is the same as shown in Figure~\ref{fig:fig_11}, but organized here effectively as an integrated graph rather than by prompt.}
	\label{fig:fig_12}
\end{figure}

\clearpage
\begin{figure}
	\centering
	\includegraphics[width=.8\linewidth, 
    ]{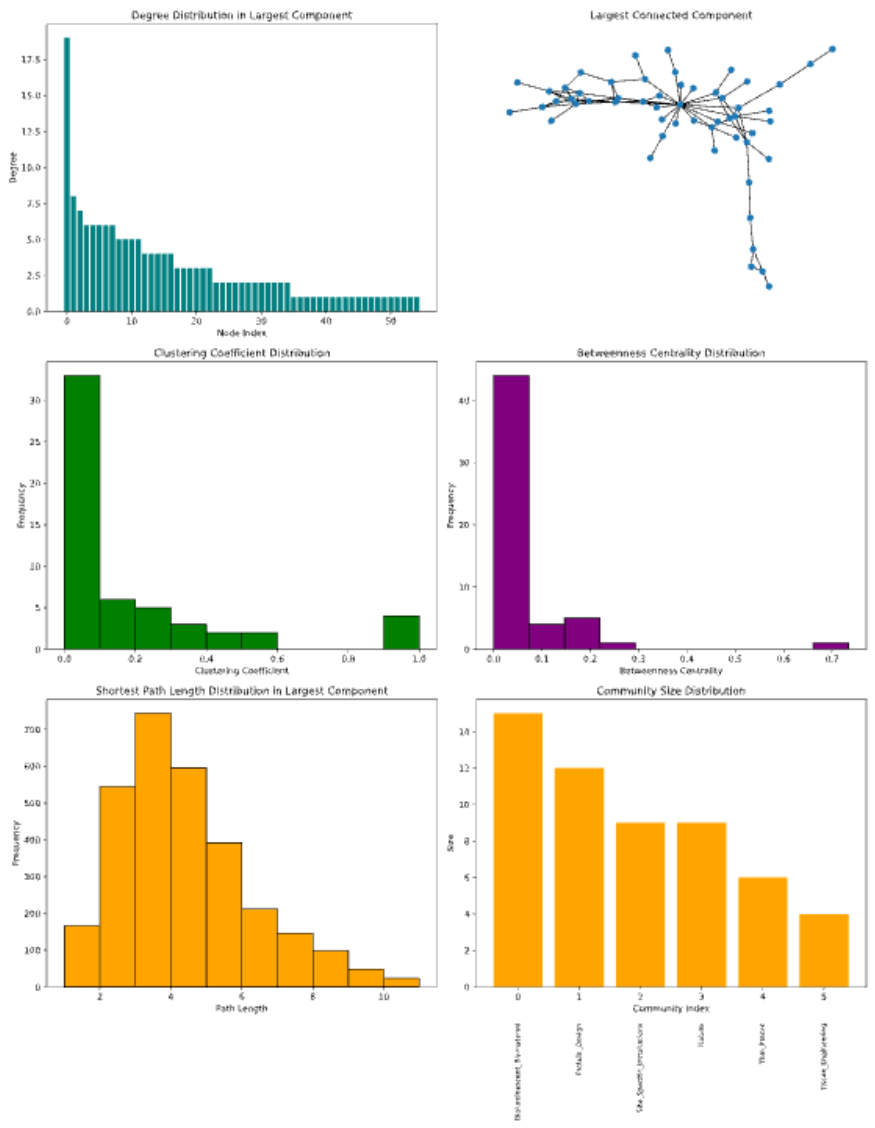}
        \caption{Visualization of the graph analysis of the autonomously grown graph depicted in Figure~\ref{fig:fig_2000}. 
(Top Left) Degree distribution shows a heterogeneous structure with prominent hubs and many low-degree nodes. 
(Top Right) Visualization of the largest connected component reveals its modular structure. 
(Middle Left) Clustering coefficient distribution highlights sparse local connectivity, with some nodes in dense clusters. 
(Middle Right) Betweenness centrality distribution indicates a small subset of critical nodes facilitating connectivity. 
(Bottom Left) Shortest path length distribution suggests a small-world structure with most nodes separated by a few hops. 
(Bottom Right) Community size distribution, with central nodes labeled, reflects varying sizes and roles within the graph.
The analysis collectively underscores the complex topology of the graph, including heterogeneity, modularity, and small-world characteristics. 
}
	\label{fig:fig_1001_other_exp}
\end{figure}

\clearpage
\newpage

\section{Example: Mapping two distinct, but isomorphic graphs into a shared representation}
\label{GIN_mapping_ex}
\normalsize

For sake of the demonstration, we have two distinct graphs that represent the equations:
\[
G_1: F = m \times a, 
\quad
G_2: V = I \times R,
\]
where \( G_1 \) and \( G_2 \) are defined as detailed below:

\paragraph{Graph Structures:}
\begin{itemize}
    \item \textbf{Graph \( G_1 \):}
    \begin{itemize}
        \item Nodes: \( F, m, a, =, \times \)
        \item Edges: \( = \to F, = \to \times, \times \to m, \times \to a, m \to \times, a \to \times \)
    \end{itemize}
    \item \textbf{Graph \( G_2 \):}
    \begin{itemize}
        \item Nodes: \( V, I, R, =, \times \)
        \item Edges: \( = \to V, = \to \times, \times \to I, \times \to R, I \to \times, R \to \times \)
    \end{itemize}
\end{itemize}

\paragraph{Initial Embeddings:}
The nodes in \( G_1 \) and \( G_2 \) are assigned distinct two-dimensional embeddings to reflect their unique roles in the equations. The initial embeddings, chosen randomly here, are referred to as iteration 0 in Table~\ref{tab:initial-distinct}. Of course, embeddings in a LLM are much higher-dimensional but the 2D illustration allows us to exemplify the approach more easily. In the scope of a LLM, the initial embeddings would reflect how tokens are mapped into the embedding space via the embedding layer from which on they are processed further. 

\begin{table}[htbp]
\centering
\caption{Distinct 2D embeddings (iteration~0) for each variable in \( G_1 \) and \( G_2 \). the variables $F$, $m$, $a$, $V$, $I$ and $R$ have different embeddings as they would in a LLM, since each token is assigned a unique embedding. The embeddings for $=$ and $\times$ are identical in both cases, as they would be in a LLM since the same tokens are mapped to the same vector. }
\label{tab:initial-distinct}
\begin{tabular}{lll}
\toprule
\textbf{Node} & \textbf{Graph} & \textbf{Initial Embedding} \\
\midrule
\(F\) & \(G_1\) & \((0.7,\; 0.3)\)\\
\(m\) & \(G_1\) & \((0.6,\; 0.9)\)\\
\(a\) & \(G_1\) & \((0.1,\; 0.9)\)\\
\(=\) & \(G_1\) & \((1.2,\; 0.5)\)\\
\(\times\) & \(G_1\) & \((0.8,\; 0.6)\)\\
\midrule
\(V\) & \(G_2\) & \((0.2,\; 0.9)\)\\
\(I\) & \(G_2\) & \((0.05,\; 0.8)\)\\
\(R\) & \(G_2\) & \((0.9,\; 0.1)\)\\
\(=\) & \(G_2\) & \((1.2,\; 0.5)\)\\
\(\times\) & \(G_2\) & \((0.8,\; 0.6)\)\\
\bottomrule
\end{tabular}
\end{table}

\paragraph{Graph Isomorphism Network (GIN) Update Mechanism:}
For details on GIN models, we refer readers to other literature~\cite{xu2019how,Buehler2025GraphAwareGPT}. For the demonstration here we assume we have a small MLP with nonlinear activations (ReLU) in our GIN update:
\[
\mathbf{h}_v^{(k+1)} = \mathrm{MLP}\bigl((1 + \epsilon) \cdot \mathbf{h}_v^{(k)} + \!\!\sum_{u \in \mathcal{N}(v)} \mathbf{h}_u^{(k)}\bigr),
\]
where \(\mathcal{N}(v)\) are the neighbors of \( v \), and
\[
\mathrm{MLP}(\mathbf{x}) 
= 
\sigma\bigl(W_L \,\sigma(W_{L-1} \dots \sigma(W_2 \,\sigma(W_1 \mathbf{x} + \mathbf{b}_1) + \mathbf{b}_2) \dots + \mathbf{b}_{L-1}) + \mathbf{b}_L\bigr).
\]
Weights/biases \((W_1, W_2, \dots \mathbf{b}_1, \mathbf{b}_2, \dots\) ) are  learnable parameters optimized during training of the model. 

\paragraph{Final Embeddings After One Iteration}

For the sake of the demonstration, we identified a particular set of weights and biases such that the embeddings of the distinct variables are mapped into the same representation by the GIN operation. Hence, after one iteration of summation and MLP, the updated embeddings \(\mathbf{h}_v^{(1)}\) are shown in Table~\ref{tab:iter1-distinct}. 

\begin{table}[htbp]
\centering
\caption{Iteration~1 embeddings after the GIN update with a nonlinear MLP. 
Notice that the embeddings for \( G_1 \) and \( G_2 \) are now aligned, respecting the functional equivalence of their structures. See Figure~\ref{fig:fig_7000} for a plot.}
\label{tab:iter1-distinct}
\begin{tabular}{cccc}
\toprule
\textbf{Node} & \textbf{Graph} & \(\mathbf{h}_v^{(1)}\) & \textbf{(Comments)}\\
\midrule
\(=\) & \(G_1\) & \((0.58,\; -0.81)\) & Matches \(=^{(1)}\) in \(G_2\)\\
\(F\) & \(G_1\) & \((-0.96,\; 0.29)\) & Matches \(V^{(1)}\) in \(G_2\)\\
\(\times\) & \(G_1\) & \((0.39,\; -0.92)\) & Matches \(\times^{(1)}\) in \(G_2\)\\
\(m\) & \(G_1\) & \((-0.70,\; 0.72)\) & Matches \(I^{(1)}\) in \(G_2\)\\
\(a\) & \(G_1\) & \((0.68,\; 0.73)\) & Matches \(R^{(1)}\) in \(G_2\)\\
\midrule
\(=\) & \(G_2\) & \((0.58,\; -0.81)\) & Matches \(=^{(1)}\) in \(G_1\)\\
\(V\) & \(G_2\) & \((-0.96,\; 0.29)\) & Matches \(F^{(1)}\) in \(G_1\)\\
\(\times\) & \(G_2\) & \((0.39,\; -0.92)\) & Matches \(\times^{(1)}\) in \(G_1\)\\
\(I\) & \(G_2\) & \((-0.70,\; 0.72)\) & Matches \(m^{(1)}\) in \(G_1\)\\
\(R\) & \(G_2\) & \((0.68,\; 0.73)\) & Matches \(a^{(1)}\) in \(G_1\)\\
\bottomrule
\end{tabular}
\end{table}

\paragraph{Discussion}

\begin{figure}
	\centering
	\includegraphics[width=.85\linewidth]{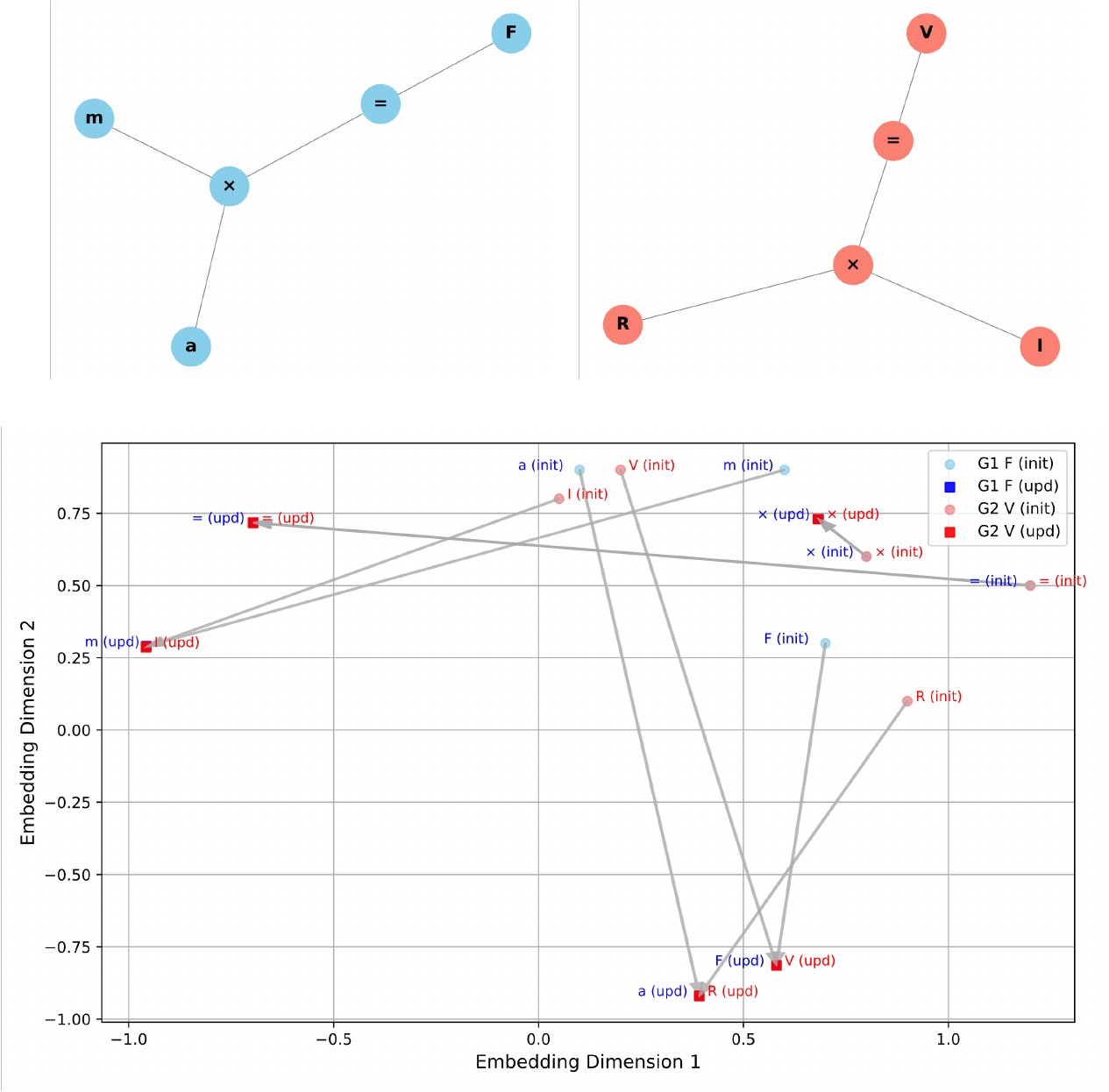}
        \caption{Visualization of transformation of two isomorphic graphs, each representing an equation ($F  = m \times a$ and $ V = I \times R$ (top), into an identical representation (bottom). As one can see, after the transformation of node embeddings by a GIN, the nodes of both graphs collapse on the same embeddings are are indistinguishable, to yield a generalized representation (akin to the concept described in Figure~\ref{fig:fig_01}, bottom). The two graphs not only have the same node embeddings but also the same graph-level embeddings, revealing that they are isomorphic.}
	\label{fig:fig_7000}
\end{figure}

This example demonstrates the power of GINs in mapping distinct concepts into a shared representation, forming one example of how abstraction can be achieved. Despite the graphs \( G_1 \) and \( G_2 \) starting with distinct initial embeddings and different structures, the GIN successfully aligned their embeddings after one iteration. This alignment reflects the structural equivalence of the graphs:
\[
\textit{isomorphic graphs with equivalent structures}
\;\Longrightarrow\;
\textit{aligned embeddings}.
\]
Thus, the equations \( F = m \times a \) and \( V = I \times R \) share the same functional relationships (\(\text{output} = \text{parameter} \times \text{input}\)), leading to  aligned embeddings despite differing physical meanings. It confirms that after processing by the GIN, the two isomorphic graphs have the same node embeddings and graph-level embeddings. Figure~\ref{fig:fig_7000} depicts a visualization of the transformation, including the graphs and how their embeddings adapt after one layer of GIN. 

The argument with relation to Graph-PReFLexOR is that rather than developing a training objective to model isomorphic mappings explicitly, we model these transformations explicitly by mapping a task into a graph representation, then a symbolic abstraction, and then to the final answer, as shown in Figure~\ref{fig:fig_01}. In this visualization, the bottom part of the image particularly shows how problems are expressed in shared tokenization form to yield effectively a similar result as visualized in Figure~\ref{fig:fig_7000}. This reflects a mapping into abstraction for generalization and back, achieving mulitple goals simultaneously. 

\end{document}